\documentclass[twoside,11pt]{article}
\usepackage{jair, rawfonts}
\usepackage[backend=biber, style=apa]{biblatex}
\usepackage{graphicx}
\usepackage{hyperref}
\usepackage{color}
\usepackage{epigraph}
\usepackage{csquotes}
\usepackage{amssymb}
\usepackage{amsmath}
\usepackage{amsthm}
\usepackage{bbm}
\usepackage{tikz}
\usetikzlibrary{arrows}
\usetikzlibrary{shapes.geometric}
\usepackage{caption}
\usepackage{subcaption}
\usepackage[algo2e,ruled,linesnumbered,noend]{algorithm2e}
\usepackage{soul}
\usepackage{bbm}
\usepackage{eurosym}
\usepackage{booktabs}
\usepackage{float}
\theoremstyle{plain}

\theoremstyle{definition}
\newtheorem{definition}{Definition}[section]
\newtheorem{remark}{Remark}[section]
\newtheorem{proposition}{Proposition}[section]
\newtheorem{example}{Example}[section]

\jairheading{82}{2025}{2279-2323}{12/2024}{04/2025}
\ShortHeadings{Counterfactual Situation Testing}{Alvarez \& Ruggieri}
\firstpageno{2279}

\begin{document}

\title{Counterfactual Situation Testing: \\ From Single to Multidimensional Discrimination}

\author{\name Jose M. Alvarez \email josemanuel.alvarez@kuleuven.be \\
\addr Department of Computer Science, KU Leuven \\ 3001 Leuven, Belgium
\AND
\name Salvatore Ruggieri \email salvatore.ruggieri@unipi.it \\
\addr Department of Computer Science, University of Pisa \\ 56126 Pisa, Italy}

\maketitle

\begin{abstract}
    As machine learning models enable decisions once performed only by humans, it is central to develop tools that assess the fairness of such models. Notably, within high-stake settings like hiring and lending, these tools must be able to detect potentially discriminatory models. We present counterfactual situation testing (CST), a causal data mining framework for detecting individual discrimination in a dataset of classifier decisions. CST answers the question ``what would have been the model outcome had the individual, or complainant, been of a different protected status?'' It extends the legally-grounded situation testing (ST) of \textcite{Thanh_KnnSituationTesting2011} by operationalizing the notion of \textit{fairness given the difference} via counterfactual reasoning. ST finds for each complainant similar protected and non-protected instances in the dataset; constructs, respectively, a control and test group; and compares the groups such that a difference in model outcomes implies a potential case of individual discrimination. CST, instead, avoids this idealized comparison by establishing the test group on the complainant's generated counterfactual, which reflects how the protected attribute when changed influences other seemingly neutral attributes of the complainant. Under CST we test for discrimination for each complainant by comparing similar individuals within the control and test group but dissimilar individuals across these groups. We consider single (e.g.,~gender) and multidimensional (e.g.,~gender and race) discrimination testing. For multidimensional discrimination we study multiple and intersectional discrimination and, as feared by legal scholars, find evidence that the former fails to account for the latter kind. Using a k-nearest neighbor implementation, we showcase CST on synthetic and real data. Experimental results show that CST uncovers a higher number of cases than ST, even when the model is counterfactually fair. CST, in fact, extends counterfactual fairness (CF) of \textcite{Kusner2017CF} by equipping CF with confidence intervals, which we report for all experiments.
\end{abstract}

\section{Introduction}
\label{sec:Introduction}
Many decisions today are increasingly enabled by machine learning (ML) models.
Algorithmic decision-making (ADM) is becoming ubiquitous and its societal discontents clearer \parencite{Angwin2016MachineBias, DastinAmazonSexist, Heikkila2022_DutchScnadal}. 
There is a shared urgency by regulators and researchers alike to develop frameworks that asses these ML models for potential discrimination based on protected attributes such as gender, race, and religion \parencite{Kleinberg2019DiscAgeOfAlgo, DBLP:conf/aaai/Ruggieri0PST23, DBLP:journals/ethicsit/AlvarezCEFFFGMPLRSSZR24}. 
Discrimination is often conceived as a causal claim on the effect of the protected attribute over an individual decision outcome \parencite{Heckman1998_DetectingDiscrimination}. It is, in particular, a conception based on counterfactual reasoning---what would have been the ML model outcome if the individual, or \textit{complainant}, were of a different protected status?---where we ``manipulate'' the protected attribute of the individual. \textcite{Kohler2018CausalEddie} calls such conceptualization of discrimination the \textit{counterfactual causal model of discrimination} (CMD). 

Most frameworks for proving discrimination are based on the CMD. 
Central to these frameworks is defining similar instances to the complainant; arranging them based on their protected status into control and test groups; and comparing the decision outcomes of these two groups to detect the effects of the protected attribute. 
Among the available frameworks \parencite{Romei2014MultiSurveyDiscrimination,DBLP:journals/jlap/MakhloufZP24, DBLP:journals/fdata/CareyW22}, however, there is a need for one that is both \textit{actionable} and \textit{meaningful}. 
A framework is actionable if it rules out random circumstances from the discrimination claim as required by courts (e.g.,~\textcite{Foster2004, EU2018_NonDiscriminationLaw, Nachbar2020algorithmic}) and meaningful if it accounts for known links between the protected attribute and all other attributes 
as demanded by social scientists (e.g.,~\textcite{Bonilla1997_RethinkingRace, Sen2016_RaceABundle, Kasirzadeh2021UseMisuse}).
In practice, we view actionability as an inferential concern to be handled by comparing multiple control-test instances around a complainant, while meaningfulness as an ontological concern to be handled by requiring domain-knowledge on the protected attribute and its effect on the other attributes of the complainant.

In this work, we present \textit{counterfactual situation testing} (CST), a causal data mining framework for detecting individual cases of discrimination in a dataset of classifier decisions.
The dataset can be the one used to train a ML model or the one of actual decisions by a ML model. 
In the former case we want to prevent learning discriminatory patterns, while in the latter case we want to detect discriminatory decisions.
The goal of CST is to be both an actionable and meaningful framework.
It combines (structural)\footnote{Not to be confused with counterfactual explanations \parencite{Wachter2017Counterfactual}. \textcite{Karimi2021_AlgoRecourse}, e.g.,~use ``structural'' to differentiate counterfactuals from counterfactual explanations.} 
counterfactuals \parencite{PearlCausality2009} with situation testing \parencite{Thanh_KnnSituationTesting2011}.
\textit{Counterfactuals} answer to counterfactual queries, such as the one motivating the CMD, and are generated via structural causal models. 
Under the right causal knowledge, counterfactuals reflect at the individual level how changing the protected attribute affects other seemingly neutral attributes of a complainant.
\textit{Situation testing} is a data mining method, based on the homonymous legal tool \parencite{Bendick2007SituationTesting, Rorive2009_ProvingDiscrimination}. 
For each complainant, given a search algorithm and distance function for measuring similarity, situation testing finds and compares a control and test group of similar protected and non-protected instances in the dataset, where a difference between the decision outcomes of the groups implies potential discrimination.
Hence, \textit{CST follows the situation testing pipeline with the important exception that it constructs the test group around the complainant's counterfactual instead of the complainant.}
To illustrate CST and how it compares to standard situation testing, let us use Example~\ref{ex:IllustrativeExample} below.

\begin{example}(An illustrative example)
\label{ex:IllustrativeExample}
    Let us consider the scenario in Figure~\ref{fig:KarimiV2} in which a bank uses the classifier $b$ to accept or reject ($\hat{Y}$) individual loan applications based on annual salary ($X_1$) and account balance ($X_2$), such that $b(X_1, X_2) = \hat{Y}$. 
    Suppose a female applicant ($A=1$) with $x_1= 35000$ and $x_2=7048$ is rejected and files for gender discrimination. 
    According to Figure~\ref{fig:KarimiV2}, the bank uses non-sensitive information to calculate $\hat{Y}$, but there is also a ``known'' link between $A$ and $\{X_1, X_2\}$ that questions the neutrality of the information. 
    \textit{Under situation testing}, we find a number of females and males with similar characteristics in terms of $X_1$ and $X_2$ to the complainant and compare them.
    Comparing multiple instances allows to check whether the complainant's claim is representative of an unfavorable pattern toward female applicants by the model (i.e.,~actionability). 
    However, knowing what we know about $A$ and its influence on $X_1$ and $X_2$, would it be fair to compare these similar females and males? 
    As argued by works like \textcite{Kohler2018CausalEddie}, the answer is no as this \textit{idealized comparison} 
    takes for granted the effect of $A$ on $X_1$ and $X_2$ by allowing the former to change while expecting the latter two to remain the same despite the known links.
    \textit{Under counterfactual situation testing}, instead, we generate the complainant's counterfactual using the auxiliary causal knowledge, creating a male applicant with a higher $x_1=50796$ and $x_2=13852$, and use him 
    to find similar male instances for constructing the test group.
    The resulting control and test groups have similar $X_1$ and $X_2$ within them but different $X_1$ and $X_2$ between them. 
    This disparate comparison embodies \textit{fairness given the difference}, explicitly acknowledging the lack of neutrality when looking at $X_1$ and $X_2$ based on $A$ (i.e.,~meaningfulness).
    We come back to this example in Section~\ref{sec:Experiments}.
\end{example}
\begin{figure}[t]
\begin{minipage}{.45\linewidth}
\begin{figure}[H]
\centering
    \begin{tikzpicture}
        \node (A)  at (-1.75, 0) [circle, draw]{$A$};
        \node (X1) at (0, 0.85) [circle, draw]{$X_1$};
        \node (X2) at (0,-0.85) [circle,draw]{$X_2$};
        \node (Y)  at (1.75, 0) [circle, draw]{$\widehat{Y}$};
        \draw[->] (A) to (X1) {};
        \draw[->] (A) to (X2) {};
        \draw[->] (X1) to (Y) {};
        \draw[->] (X1) to (X2) {};
        \draw[->] (X2) to (Y) {};
    \end{tikzpicture}
\end{figure}
\end{minipage}
\begin{minipage}{.45\linewidth}
\begin{align*}
\mathcal{M} \, & 
\begin{cases}
    A & \leftarrow U_{A} \\
    X_1 & \leftarrow f_1(A)  + U_1 \\
    X_2 & \leftarrow f_2(X_1, A) + U_2
\end{cases}
\end{align*}
\begin{align*}
    \widehat{Y} & = b(X_1, X_2) 
\end{align*}
\end{minipage}
\caption{The auxiliary causal knowledge for Example~\ref{ex:IllustrativeExample} (and Section~\ref{sec:Experiments.IllustrativeExample}). Let $A$ denote gender, $X_1$ annual salary, $X_2$ account balance, and $\widehat{Y}$ the loan decision by $b()$. It consists of a causal graph (left) and a set of structural equations (right), both introduced in Section~\ref{sec:CausalKnowledge}.}
\label{fig:KarimiV2}
\end{figure}

We evaluate the CST framework on synthetic and real ADM datasets.
We use a k-nearest neighbor implementation of the framework, k-NN CST, to compare it to its situation testing counterpart, k-NN ST, by \textcite{Thanh_KnnSituationTesting2011}.
The $k$ denotes the number of instances for each control and test groups, determining the size of the many-to-many comparison of each complainant in the dataset.
Our experiments show that CST detects a higher number of individual discrimination cases across different $k$ sizes. 
The results illustrate the impact of moving from the idealized comparison of the k-NN ST to the fairness given the difference comparison of the k-NN CST.
This last point is important as legal scholars continue to call for an alternative to the idealized comparison \parencite{Kohler2018CausalEddie}.
Importantly, we consider single and multidimensional discrimination, meaning, respectively, claims based on one and many protected attributes.
While single discrimination testing is commonly studied, multidimensional discrimination testing is largely unexplored and often portrayed as a straightforward extension to single discrimination testing \parencite{Xenidis2020_TunningEULaw}.

Multidimensional discrimination covers two forms: multiple and intersectional. 
In multiple discrimination the complainant must be discriminated for each of the protected attributes, while in intersectional discrimination the complainant must be discriminated at the intersection of the protected attributes.
To illustrate this distinction, suppose the complainant in Example~\ref{ex:IllustrativeExample} is also non-white and makes a claim based on gender and race. 
Multiple discrimination occurs if the complainant is discriminated, separately, as a female and as a non-white individual.
Intersectional discrimination occurs if the complainant is discriminated, simultaneously, as a female-non-white individual.
Each form of multidimensional discrimination, in turn, poses different problem formulations for discrimination testing \parencite{DBLP:conf/fat/0001HN23, WangRR22}.
Beyond the distinct problem formulations, an open issue with these two forms of discrimination is that only multiple discrimination is recognized by non-discrimination law. 
Legal scholars have raised concerns on this lack of recognition for intersectional discrimination, arguing that multiple discrimination fails to account for it \parencite{Xenidis2020_TunningEULaw}.
We test for multiple and intersectional discrimination using CST, finding that the former does not capture the latter.
This work is the first to evaluate and provide evidence for this legal concern.

Additionally, CST provides an actionable extension to \textit{counterfactual fairness} by \textcite{Kusner2017CF}, which remains the leading causal fair ML definition \parencite{DBLP:journals/jlap/MakhloufZP24}.
A ML model is counterfactually fair when the complainant's and its counterfactual's decision outcomes are the same. 
These are the same instances used by CST to construct, respectively, the control and test groups, which allows to equip this popular fairness definition with measures for uncertainty due to the many-to-many comparison. 
CST links counterfactual fairness claims with statistical significance, and
positions it for discrimination testing as uncertainty measures are often required by courts \parencite{EU2018_NonDiscriminationLaw}.
By looking at the control and test groups rather than the literal comparison of the factual versus counterfactual instances, CST evaluates whether the counterfactual fairness claim itself is representative of similar instances.
Our results show that individual discrimination can occur even when the ML model is counterfactually fair, capturing the scenario where a model discriminates when evaluating borderline instances.

In summary,
with CST we present a meaningful and actionable framework for detecting individual discrimination.
Our main contributions are threefold.
First, we offer the first explicit operationalization of \textit{fairness given the difference} for discrimination testing and, in doing so, define a new view on similarity that is more flexible than the standard idealized comparison.
Second, we explore single and multidimensional discrimination testing, studying the latter's tension between multiple and intersectional discrimination.
Third, we equip counterfactual fairness with confidence intervals, introducing an actionable extension to the popular causal fairness definition.

The rest of the paper is organized as follows.
We present the related work in Section~\ref{sec:RelatedWork} and the role of auxiliary causal knowledge within CST for discrimination testing in Section~\ref{sec:CausalKnowledge}.
We introduce the CST framework in Section~\ref{sec:CST}, including its k-NN implementation.
We showcase CST using two classification scenarios in Section~\ref{sec:Experiments}.
We discuss the main limitations of this work in Section~\ref{sec:Discussion}.
We conclude in Section~\ref{sec:Conclusion}.

%
%

\section{Related Work}
\label{sec:RelatedWork}
We position CST with respect to current frameworks for discrimination testing along the goals of actionability and meaningfulness.
Later in Section~\ref{sec:CausalKnowledge} we discuss the role of causality for conceiving discrimination.
For a broader, multidisciplinary view on discrimination testing, we refer to the survey by~\textcite{Romei2014MultiSurveyDiscrimination}. 
For a recent survey of the fair ML testing literature, see \textcite{DBLP:journals/tosem/ChenZHHS24}.

Regarding actionability, it is important when proving discrimination to insure that the framework accounts for sources of randomness in the decision-making process. Popular non-algorithmic frameworks---such as natural \parencite{Godin2000Orchestra} and field \parencite{Bertrand2017_FieldExperimentDiscrimination} experiments, audit \parencite{Fix&Struyk1993_ClearConvincingEvidence} and correspondence \parencite{Bertrand2004_EmilyAndGreg, Rooth2021} studies---address this issue by using multiple observations to build inferential statistics. Similar statistics are asked in court for proving discrimination \parencite[Section 6.3]{EU2018_NonDiscriminationLaw}. 
Few algorithmic frameworks address this issue due to model complexity preventing formal inference \parencite{Athey2019MachineLearningForEconomists}. An exception are data mining frameworks for discrimination discovery \parencite{DBLP:conf/kdd/PedreschiRT08, DBLP:journals/tkdd/RuggieriPT10} that operationalize the non-algorithmic notions, including situation testing \parencite{Thanh_KnnSituationTesting2011, Zhang_CausalSituationTesting_2016}.
These frameworks \parencite{TR-DBLP:conf/sigsoft/GalhotraBM17, TR-DBLP:journals/corr/abs-1809-03260, DBLP:journals/jiis/QureshiKKRP20} keep the focus on comparing multiple control-test instances for making individual claims, providing evidence similar to that produced by the quantitative tools used in court.
It remains unclear if the same can be said about existing causal fair machine learning methods
as these have yet to be used beyond academic circles.
The suitability of algorithmic fairness methods for testing discrimination, be it or not ADM, remains an ongoing discussion \parencite{DBLP:conf/fat/WeertsXTOP23}.

Regarding meaningfulness, situation testing and the other methods previously mentioned have been criticized for their handling of the counterfactual question behind the causal model of discrimination \parencite{Kohler2018CausalEddie, Hu_facct_sex_20, Kasirzadeh2021UseMisuse}. In particular, these actionable methods take for granted the influence of the protected attribute on all other attributes. This point can be seen, e.g.,~in how situation testing constructs the test group, which is equivalent to changing the protected attribute while keeping everything else equal. Such an approach goes against how most social scientists interpret the protected attribute and its role as a social construct when proving discrimination \parencite{Bonilla1997_RethinkingRace, rose_constructivist_2022, Sen2016_RaceABundle, Hanna2020_CriticalRace}. 
It is in that regard where structural causal models \parencite{PearlCausality2009} and their ability to generate counterfactuals via the abduction, action, and prediction steps (e.g.,~\textcite{Chiappa2019_PathCF, Yang2021_CausalIntersectionality}), including counterfactual fairness \parencite{Kusner2017CF}, have an advantage.
This advantage is overlooked by critics of counterfactual reasoning \parencite{Kasirzadeh2021UseMisuse, Hu_facct_sex_20}: generating counterfactuals, as long as the structural causal model is properly specified, accounts for the effects of changing the protected attribute on all other attributes. 
Hence, a framework like counterfactual fairness, relative to situation testing and other discrimination discovery methods, is more meaningful in its handling of protected attributes. 

CST bridges these two lines of work, borrowing the actionability aspects from frameworks like situation testing and meaningful aspects from frameworks like counterfactual fairness. 
Counterfactual generation allows to create a comparator for the complainant that accounts for the influence of the protected attribute on the other attributes, departing from the idealized comparison.
It is not far, conceptually, from the broader ML problem of learning fair representations \parencite{Zemel2013LearningFairRepresentations} since we wish to learn (read, map) a new representation of the complainant that reflects where it would have been had it belonged to the non-protected group. 
It is a normative claim on what a non-protected instance similar to the complainant looks like.
Beyond counterfactual fairness and derivatives (e.g., \textcite{Chiappa2019_PathCF}), other works address this problem of deriving such a pair for the complainant.
For instance, \textcite{Plevcko2020FairDataAdaptation} use a quantile regression approach while \textcite{DBLP:journals/corr/abs-2307-12797} use a residual-based approach for generating the pair.
Both works rely on having access to a structural causal model, but do not exploit the abduction, action, and prediction steps for generating counterfactual distributions.
\textcite{BlackYF20_FlipTest}, instead, propose the FlipTest, a non-causal approach that uses an optimal transport mapping to derive the pair for the complainant.
These three works exemplify ML methods that use counterfactual reasoning to operationalize different interpretations of individual similarity. 
With CST we align with these and similar efforts to propose an alternative to the idealized comparison often used in discrimination testing.

%
%

\section{Causal Knowledge for Discrimination Testing}
\label{sec:CausalKnowledge}
In this section, we discuss the role of auxiliary causal knowledge within CST. 
We formulate causality using structural causal models (SCM). 
CST requires access to the dataset of decisions, $\mathcal{D}$, and the ML model that produced it, $b()$.
CST also requires a SCM describing the data generating model behind $\mathcal{D}$.
We view this last requirement as an input space for stakeholders and domain experts. 
SCM are a convenient way for organizing assumptions on the source of the discrimination, facilitating stakeholder participation and supporting collaborative reasoning about contested concepts \parencite{Mulligan2022_AFCP}. 
There is no ground-truth concerning the SCM for $\mathcal{D}$.
The SCM describes an agreed view on the discrimination problem, though, not necessarily the only nor correct view of it.

Let $\mathcal{D}$ contain the set of relevant attributes $X$, the set of protected attributes $A$, and the decision outcome $\hat{Y}$ such that $\hat{Y}=b(X)$. 
We describe $\mathcal{D}$ as a collection of $n$ tuples, each $(x_i, a_i, \widehat{y}_i)$ representing the $i^{th}$ individual profile, with $i \in [1, n]$. $\hat{Y}$ is binary with $\hat{Y} = 1$ denoting the positive outcome (e.g.,~loan granted). 
For illustrative purposes, we assume a single binary $A$ with $A=1$ denoting the protected status (e.g., female gender). 
We relax this assumption in Section~\ref{sec:CST.Multi} when formalizing multidimensional discrimination.
 
\subsection{Structural Causal Models and Counterfactuals}
\label{sec:CausalKnowledge.SCM}

A \textit{structural causal model} (SCM) \parencite{PearlCausality2009} $\mathcal{M}=\{ \mathcal{S}, \mathcal{P}_{\mathbf{U}} \}$ describes how the set of $p$ variables $W = X \cup A$ is determined based on corresponding sets of structural equations $\mathcal{S}$,
and $p$ latent variables $U$ with prior distribution $\mathcal{P}_{\mathbf{U}}$. Each $W_j \in W$ is assigned a value through a deterministic function $f_j \in \mathcal{S}$ of its causal parents $W_{pa(j)} \subseteq W \setminus \{ W_j \}$ and latent variable $U_j$ with distribution $P(U_j) \in \mathcal{P}_{\mathbf{U}}$. Formally, for $W_j \in W$ we have that: 
\begin{equation}
\label{eq:SCM}
    W_j \leftarrow f_j(W_{pa(j)}, U_j)
\end{equation}
indicating the flow of information in terms of child-parent or cause-effect pairs. We consider the associated \textit{causal graph} $\mathcal{G} = (\mathcal{V}, \mathcal{E})$, where a node $V_j \in \mathcal{V}$ represents a $W_j$ variable and a directed edge $E_{(j, j')} \in \mathcal{E}$ a causal relation.
We can use $\mathcal{M}$ to derive $\mathcal{G}$.\footnote{This is based on the global Markov and faithfulness properties, summarized in the notion of the d-separation. We skip d-separation as we do not use it in this paper. See \textcite{Peters2017_CausalInference}.}

We make three assumptions for the SCM $\mathcal{M}$ that are common within the causal fairness literature \parencite{DBLP:journals/jlap/MakhloufZP24}. 
First, we assume \textit{causal sufficiency}, meaning there are no hidden common causes in $\mathcal{M}$, or confounders. 
Second, we assume $\mathcal{G}$ to be \textit{acyclical}, which turns $\mathcal{G}$ into a directed acyclical graph (DAG), allowing for no feedback loops.
Third, we assume \textit{additive noise models} (ANM) to insure an invertible class of SCM \parencite{Hoyer2008_ANM}.
The ANM assumption implies $\mathcal{S} = \{W_j \leftarrow f_j(W_{pa(j)}) + U_j \}_{j=1}^p$ in \eqref{eq:SCM}.
These assumptions are not necessary for generating counterfactuals, but do simplify the process \parencite{Pearl2016_CausalInference}. 
The CST framework is not tied to any of these assumptions as long as the generated counterfactuals are reliable.

The causal sufficiency assumption is particularly deceitful as it is difficult to both test and account for a hidden confounder \parencite{DBLP:journals/corr/abs-1902-10286, mccandless2007bayesian, DBLP:conf/nips/LouizosSMSZW17}. 
The risk of a hidden confounder is a general modeling problem. Here, the dataset $\mathcal{D}$ delimits our context. By this we mean that we expect it to contain all relevant information used by $b()$.
Causal sufficiency implies independence among the random variables in $U$, which allows to factorize $P_{\mathbf{U}}$ into its individual components:
\begin{equation}
\label{eq:SCM.Causal.Sufficiency}
    P(U_1, \dots, U_j) = P(U_1) \times \dots \times P(U_j).
\end{equation}

For a given SCM $\mathcal{M}$ we want to run \textit{counterfactual queries} to build the test group for a complainant. Counterfactual queries answer to \textit{what would have been if} questions. 
In CST, we ask such questions around the protected attribute $A$. 
By setting $A$ to the non-protected status $\alpha$ using the \textit{do-operator} $do(A := \alpha)$ \parencite{PearlCausality2009}, we capture the individual-level effects $A$ has on $X$ according to the SCM $\mathcal{M}$. 
Let $X^{CF}$ denote \textit{the set of counterfactual variables} obtained via the three steps: abduction, action, and prediction \parencite{Pearl2016_CausalInference}. 
Further, 
let $P(X^{CF}_{A \leftarrow \alpha}(U) \; | \; X, A)$ denote \textit{counterfactual distribution}.
We now describe each step.
\textit{Abduction}: for each prior distribution $P(U_i)$ that describes $U_i$, we compute its posterior distribution given the evidence, or $P(U_i \;| \; X, A)$. \textit{Action}: we intervene $A$ by changing its structural equation to $A := \alpha$, which gives way to a new SCM $\mathcal{M}'$. \textit{Prediction}: we generate the \textit{counterfactual distribution} $P(X^{CF}_{A \leftarrow \alpha}(U) \; | \; X, A)$ by propagating the abducted $P(U_i \;| \; X, A)$ through the revised structural equations in $\mathcal{M}'$.
%
Unlike counterfactual explanations \parencite{Wachter2017Counterfactual}, generating counterfactuals and, thus, CST,  does not require a change in the individual decision outcome. 

\subsection{On Conceiving Discrimination}
\label{sec:CausalKnowledge.IndDisc}

Discrimination is a comparative process \parencite{Lippert2006BadnessOfDiscrimination}. 
Non-discrimination law is centered on Aristotle's maxim of treating similar (or similarly situated) individuals similarly \parencite{Westen1982EmptyEquality}.
Granted that we can agree on what similar (or similarly situated) individuals are,\footnote{We briefly discuss this issue in the next section, though similarity in itself is a complex, ongoing, legal discussion. We recommend \textcite{Westen1982EmptyEquality} for further reading.} in practice, testing for discrimination reduces to comparing similar protected and non-protected by non-discrimination law individuals to see if their outcomes differ within the context of interest \parencite{DBLP:conf/fat/WeertsXTOP23}.
Most, if not all, discrimination tools operationalize this comparative process \parencite{Kohler2018CausalEddie}.

The legal setting of interest in this work is indirect discrimination under EU non-discrimination law. 
Indirect discrimination occurs when an apparently neutral practice disadvantages individuals that belong to a protected group. Following \textcite{Hacker2018TeachingFairness}, we focus on indirect discrimination for three reasons. 
First, unlike disparate impact under US law \parencite{Barocas2016_BigDataImpact}, the decision-maker can still be liable for it despite lack of premeditation and, thus, all practices need to consider potential indirect discrimination implications. 
Second, many ML models are not allowed to use the protected attribute as input, making it difficult for regulators to use the direct discrimination setting.\footnote{This point, though, has been contested recently by \textcite{Adams2022DirectlyDiscriminatoryAl}.}
Third, we conceive discrimination as a product of a biased society where $b()$ continues to perpetuate the bias reflected in $\mathcal{D}$ because it cannot escape making deriving $\hat{Y}$ based on $X$.
The indirect setting best describes how biased information can still be an issue for a ML model that never uses the protected attribute.
That said, it does not mean CST cannot be implemented in other legal contexts. 
We simply acknowledge that it was developed with the EU non-discrimination legal framework in mind due to the previous reasons.

Causality is often used for formalizing the problem of discrimination testing.
This is because of the legal framing of discrimination in which we are interested in the protected attribute as a direct or indirect cause of the decision outcome \parencite{Heckman1998_DetectingDiscrimination, Kohler2018CausalEddie}.
Previous works \parencite{Kilbertus2017AvoidDiscCau, Chiappa2019_PathCF, Plecko2022_CFA, Tschantz2022_ProxyDisc} focus more on whether the paths between $A$ and $\hat{Y}$ are direct or indirect, leading to the two kinds of discrimination prescribed under EU non-discrimination law. 
The causal setting here is much simpler. 
We know that $b()$ only uses $X$, and are interested in how information from $A$ is carried by $X$ and how we account for these links when testing for discrimination by using the auxiliary causal knowledge. 

\subsection{Fairness Given the Difference}
\label{sec:CausalKnowledge.KHC}

The SCM required by CST allows to operationalize the notion of \textit{fairness given the difference}, FGD for short, and depart from the standard idealized comparison.
Access to a SCM enables the generation of a counterfactual instance for the complainant, allowing to represent how the protected attribute influences the non-protected attributes used by $b()$.
We come back to this point in the next section; here, we motivate FGD.
The reference work is \textcite{Kohler2018CausalEddie}.
FGD captures that work's overall criticism toward the counterfactual causal model of discrimination (CMD) introduced in Section~\ref{sec:Introduction}.\footnote{We attribute this phrase to Kohler-Hausmann as she used it during a panel discussion at the NeurIPS'21 Workshop on Algorithmic Fairness through the Lens of Causality and Robustness (\href{https://www.afciworkshop.org/afcr2021}{AFCR}). It is not, however, present in her paper. 
The phrase first appears in \textcite{DBLP:conf/eaamo/AlvarezR23}.}

As argued by \textcite{Kohler2018CausalEddie} and others before her \parencite{Bonilla1997_RethinkingRace, Sen2016_RaceABundle}, it is difficult to deny that most protected attributes, if not all of them, are \textit{social constructs}. 
These are attributes that were used to classify \textit{and} divide groups of people in a systematic way that conditioned the material opportunities of multiple generations \parencite{Mallon2007SocialConstruction, rose_constructivist_2022}. 
Recognizing $A$ as a social constructs means recognizing that its effects can be reflected in seemingly neutral variables in $X$. It is recognizing that $A$, the attribute, cannot capture alone the meaning of belonging to $A$ and that we might, as a minimum, have to link it with other attributes to better capture this, such as $A \rightarrow X$ where $A$ and $X$ change in unison. Protected attributes summarize the historical processes that fairness researchers are trying to address today and should not be treated lightly.\footnote{An example is the use of race by US policy makers after WWII. See, e.g.,~the historical evidence provided by \textcite{Rothstein2017Color} (for housing), \textcite{Schneider2008Smack} (for narcotics), and \textcite{Adler2019MurderNewOrleansJimCrow} (for policing).}

FGD centers on how $A$ is treated in the CMD. 
It goes beyond the standard manipulation concern in which $A$ is an inmutable attribute \parencite{Angrist2008MostlyHarmless}. 
Instead, granted that we \textit{can} or, more precisely, \textit{have to} manipulate $A$ for running a discrimination analysis, FGD puts into question how a testing framework operationalize such manipulation.
If $A$ is a social construct with clear influence on $X$, then \textit{when $A$ changes, $X$ should change as well}.
This is precisely what FGD entails.
As discussed in Section~\ref{sec:CST},
within CST it manifests by building the test group on the complainant's counterfactual, letting $X^{CF}$ reflect the effects of changing $A$ instead of assuming $X = X^{CF}$. 
This is because we view the test group as a representation of the hypothetical counterfactual world of the complainant.

Based on FGD we consider two types of manipulations that summarize existing discrimination testing frameworks. 
The \textit{ceteris paribus} (CP), or all else equal, manipulation in which $A$ changes but $X$ remains the same. 
Examples of it include situation testing \parencite{Thanh_KnnSituationTesting2011, Zhang_CausalSituationTesting_2016} and the famous correspondence study by \textcite{Bertrand2004_EmilyAndGreg}. 
The \textit{mutatis mutandis} (MM), or changing what needs to be changed, manipulation in which $X$ changes when we manipulate $A$ based on some additional knowledge, like a structural causal model, that explicitly links $A$ to $X$. Counterfactual fairness \parencite{Kusner2017CF} uses this manipulation. The MM is clearly preferred over the CP manipulation when we view $A$ as a social construct.
See \textcite{MutatisMutandis} for a detailed discussion on the CP and MM manipulations.

%
%

\section{Counterfactual Situation Testing}
\label{sec:CST}
The goal of CST is to construct and compare a control and test group for each protected individual (read, \textit{complainant}) $c$ in the dataset in a meaningful and actionable way. 
The focus is on the tuple $(x_c, a_c, \widehat{y}_c) \in \mathcal{D}$, with $c \in [1,n]$, that motivates the individual discrimination claim.
CST requires access to the ADM $b()$, the dataset $\mathcal{D}$, and the auxiliary causal knowledge SCM $\mathcal{M}$ and DAG $\mathcal{G}$.

Three additional inputs are central to CST:
the number of instances, $k$; 
the similarity function, $d$; 
and the strength of the evidence for the discrimination claim, $\alpha$.
Here, 
$k$ determines the size of the control and test groups for $c$; 
$d$ determines how much these two groups resemble $c$; 
and $\alpha$ determines the statistical significance required when comparing these two groups to trust the claim around $c$.
We must also define a search algorithm for implementing CST.
We use the k-nearest neighbors, or k-NN, algorithm \parencite{Hastie2009_ElementsSL}, resulting in the present k-NN CST.
The k-NN is intuitive, easy to implement, and commonly used by other frameworks.
The k-NN implementation is straightforward. 
We provide the relevant algorithms in Appendix~\ref{Appendix.Supplements}.
Other implementations are possible as long as the following definitions are adjusted.

\subsection{Measuring Individual Similarity}
\label{sec:CST.Distance}

We start by defining the \textit{similarity measure} $d$.
We use the same $d$ as the one used by \textcite{Thanh_KnnSituationTesting2011} to compare our implementation against its standard situation testing counterpart.
Let us define the \textit{between tuple distance} $d(x_1, x_2)$ as:
\begin{equation}
\label{eq:Distance}
    d(x_1, x_2) = \frac{\sum_{i=1}^{|X|} d_i(x_{1, i}, x_{2, i})}{|X|}
\end{equation}
such that $d(x_1, x_2)$ averages the sum of the \textit{per-attribute distances} $d_i(x_{1,i}, x_{2, i})$ across all attributes in $X$. 
It does not use the protected attribute(s) $A$.
A lower $d$ implies a higher similarity between the tuples $x$ and $x'$ and further implies two similar individuals.
The k-NN CST handles non-normalized attributes but, as default, we normalize them to insure comparable per-attribute distances.

The specific $d_i$ used depends on the type of the \textit{i-th} attribute.
It equals the \textit{overlap measurement} ($ol$) if the attribute $X_i$ is categorical; otherwise, it equals the \textit{normalized Manhattan distance} ($md$) if the attribute $X_i$ is continuous, ordinal, or interval.
Under this conception, $d$ amounts to Gower's distance \parencite{Gower1971}.
For illustrative purposes, we recall both $md$ and $ol$ distances below. We define $md$ as:
\begin{equation}
    md(x_{1,i}, x_{2, i}) = \frac{| x_{1,i} - x_{2, i} |}{(\max(X_i) - \min(X_i))}
\end{equation}
and we define $ol$ as:
\begin{equation}
    ol(x_{1,i}, x_{2, i}) = 
    \begin{cases}
    1 & \text{if } x_{1, i} \neq x_{2, i} \\
    0 & \text{otherwise}.
\end{cases}
\end{equation}
The choices of $d_i$ and, in turn, of $d$ are not restrictive.
We plan to explore other formulations in subsequent works, like heterogeneous distance functions \parencite{WilsonM97_HeteroDistanceFunctions} and propensity score matching \parencite{DBLP:journals/jiis/QureshiKKRP20}.
Hence, why we view $d$ as an input to rather than a characteristic of k-NN CST.

\subsection{Building the Control and Test Groups}
\label{sec:CST_ControlTest}
 
For a complainant $c$, the control and test groups are built on the search spaces and search centers for each group. 
The search spaces are derived from $\mathcal{D}$.
The search centers, however, are derived separately. The one for the control group comes from $\mathcal{D}$ while the one for the test group comes from its corresponding generated counterfactual dataset $\mathcal{D}^{CF}$. 

\begin{definition}[Search Spaces]
\label{def:SearchSpaces}
    Under a binary $A$, with $A=1$ denoting the protected status, we partition $\mathcal{D}$ into the \textit{control search space} $\mathcal{D}_c=\{(x_i, a_i, \widehat{y}_i) \in \mathcal{D}: a_i=1\}$ and the \textit{test search space} $\mathcal{D}_t=\{(x_i, a_i, \widehat{y}_i) \in \mathcal{D}: a_i=0\}$.
\end{definition}
\begin{definition}[Counterfactual Dataset]
\label{def:CounterDataset}
    Given the ADM $b()$ and SCM $\mathcal{M}$, the \textit{counterfactual dataset} $\mathcal{D}^{CF}$ includes the counterfactual mapping of each instance with a protected status in $\mathcal{D}$ via the abduction, action, and prediction steps when setting a binary $A$ to the non-protected status, or \textit{do}($A:=0$).
\end{definition}
\begin{definition}[Search Centers]
\label{def:SearchCenters}
    We use $x_c$ from the tuple of interest $(x_c, a_c, \hat{y}_c) \in \mathcal{D}$ as the \textit{control search center} for exploring $\mathcal{D}_c \subset \mathcal{D}$, and use $x_c^{CF}$ from the tuple of interest's counterfactual $(x_c^{CF}, a_c^{CF}, \hat{y}_c^{CF}) \in \mathcal{D}^{CF}$ as the \textit{test search center} for exploring $\mathcal{D}_t \subset \mathcal{D}$.
\end{definition}

We extend these definitions for $|A| > 1$ in Section~\ref{sec:CST.Multi}.
Importantly, to obtain $\mathcal{D}^{CF}$ we consider a SCM $\mathcal{M}$ in which $A$ has no causal parents, $A$ affects only elements of $X$, and $\hat{Y}=b(X)$.
See, e.g.,~Figures~\ref{fig:KarimiV2} and \ref{fig:LawSchool}.
Under such auxiliary causal knowledge, if $A$ changes then $X$ changes too. 
Here, $\mathcal{D}^{CF}$
represents the world that the complainants would have experienced had they belonged to the non-protected group. 
This is why we draw the test search center from $\mathcal{D}^{CF}$ and not from $\mathcal{D}$. 

Given the $\mathcal{D}$ and $\mathcal{D}^{CF}$, we construct the control and test groups for $c$ using the k-NN algorithm with distance function $d$ \eqref{eq:Distance}.
We want each group (read, neighborhood) to have size $k$. 
For \textbf{the control group} (\textit{k-ctr}) we use the factual tuple of interest $(x_c, a_c, \hat{y}_c) \in \mathcal{D}$ as the search center to explore $\mathcal{D}_c$:
\begin{equation}
\label{eq:kctr}
    \text{\textit{k-ctr}} =
    \{ (x_i, a_i, \widehat{y}_i) \in \mathcal{D}_c: rank_{d}( x_c, x_i) \leq k \}
\end{equation}
where $rank_{d}(x_c, x_i)$ is the rank position of $x_i$ among tuples in $\mathcal{D}_c$ with respect to the ascending distance $d$ from $x_c$. 
For \textbf{the test group} (\textit{k-tst}) we use the counterfactual tuple of interest $(x^{CF}_c, a^{CF}_c, \widehat{y}^{CF}_c) \in \mathcal{D}^{CF}$ as the search center to explore $\mathcal{D}_t$:
\begin{equation}
\label{eq:ktst}
    \text{\textit{k-tst}} = \{ (x_i, a_i, \widehat{y}_i) \in \mathcal{D}_t: rank_{d}( x^{CF}_c, x_i) \leq k \}
\end{equation}
where $rank_{d}(x_c^{CF}, x_i)$ is the rank position of $x_i$ among tuples in $\mathcal{D}_t$ with respect to the ascending distance $d$ from $x_c^{CF}$. 
We use the same $d$ for each group. 
Neither $A$ nor $\hat{Y}$ are used for constructing the groups.
Further, we can always expand \eqref{eq:kctr} and \eqref{eq:ktst} by adding constraints such as, for instance, a maximum distance $\epsilon > 0$: $\text{\textit{k-ctr}} = \{ x_i \in \mathcal{D}_c: rank_{d}( \mathbf{x}_c, x_i) \leq k \land d(x_c, x_i) \leq \epsilon \}$ and $\text{\textit{k-tst}} = \{ x_i \in \mathcal{D}_t: rank_{d}( x_c^{CF}, x_i) \leq k \land d(x_c^{CF}, x_i) \leq \epsilon \}$.

%
\begin{remark}(Meaningfulness)
\label{rem:meaningfulness}
    The choice of search centers for \eqref{eq:kctr} and \eqref{eq:ktst} operationalizes \textit{fairness given the difference} in CST, making it a meaningful framework for testing individual discrimination.
    Using $x_c$ and $x_c^{CF}$ to search for, respectively, protected and non-protected individuals in $\mathcal{D}_c$ and $\mathcal{D}_t$ is a statement on how we view the role of \textit{within group ordering} as imposed by the protected attribute $A$.
    Each search center reflects the $A$-specific ordering imposed on the search space it targets.
\end{remark}

To illustrate Remark~\ref{rem:meaningfulness}, let us consider Example~\ref{ex:IllustrativeExample}.
If being a female imposes certain systematic limitations that hinder $x_c$, then comparing $c$ to other females preserves the group ordering prescribed by $X|A=1$ as all instances involved experience $A$ in the same way.
Similarly, the generated counterfactual male instance for $c$ should reflect the group ordering prescribed by $X|A=0$. 
Here, in particular, we would expect $x_c \leq x_c^{CF}$ given what we know about the effects of $A$ on $X$.
A way to reason about this remark is through the notion of effort. 
If being female requires a higher individual effort than being male to achieve the same $x_c$, then it is fair to compare $c$ to other female instances. 
However, it is unfair to compare $c$ to other male instances without adjusting for the extra effort undertaken by $c$ to be comparable to these male instances. 
The counterfactual $x_c^{CF}$ reflects said adjustment for $c$.
For a formal discussion on effort and its role on individual fairness \parencite{DworkHPRZ12}, see \textcite{Chzhen2020WassersteinBarycenters, Chzhen2022MiniMax}.

\subsection{Detecting Discrimination}
\label{sec:CST_Disc}

For a complainant $c$, we compare the control and test groups by looking at the \textit{difference in proportion of negative decision outcomes}:
\begin{equation}
\label{eq:delta}
    \Delta p = p_c - p_t
\end{equation}
such that $p_c$ and $p_t$ represent the count of tuples with a negative decision outcome, respectively, in the control group \eqref{eq:kctr} and test group \eqref{eq:ktst}. Formally:
\begin{equation}
\label{eq:p1_and_p2}
\begin{aligned}
    p_c & = \frac{|\{ (x_i, a_i, \widehat{y}_i) \in \text{\textit{k-ctr}}: \hat{y}_i = 0 \}|}{k} \\
    p_t & = \frac{|\{ (x_i, a_i, \widehat{y}_i) \in \text{\textit{k-tst}}: \hat{y}_i = 0 \}|}{k}
\end{aligned}
\end{equation}
where only $\hat{Y}$ is used for deriving the proportions.
It follows that $p_c, p_t \in [0, 1]$ and $\Delta p \in [-1, 1]$.
We compute $\Delta p$ for all complainants in $\mathcal{D}$ regardless of their decision outcome.

CST uses $\Delta p$ to test for the complainant's individual discrimination claim. 
Implicit to this task is the \textit{accepted deviation} $\tau \in [-1, 1]$ for $\Delta p$.
It represents the maximum acceptable difference between $p_c$ and $p_t$, such that any deviation from it amounts to discrimination: i.e., $\Delta p > \tau$. 
The $\tau$ is often implied with the default choice of $\tau=0$, as we wish for protected and non-protected individuals to be rejected at equal rates.
As $\Delta p$ is a proportion comparison, $\Delta p$ is asymptotically normally distributed, which allows to build \textit{Wald confidence intervals} (CI) around it. For other confidence interval methods, such as exact methods for small samples, see \textcite{Newcombe1998}.
With the CI we equip the complainant's claim with a measure of certainty and answer whether the claim, meaning the deviation from $\tau$, is or not statistically significant.
If $\tau$ falls within the \textit{one-sided CI}, then we cannot say that the complainant's claim is statistically significant.
We write such CI for $\Delta p$ as:
\begin{align}
\label{eq:CIs}
    [\Delta p - w_{\alpha}, + \infty),
    & \; \; \; \text{with} \; \; \;
    w_{\alpha} =  z_{\alpha} \sqrt{\frac{p_c(1 - p_c) + p_t(1 - p_t)}{k}}.
\end{align}
where $z_{\alpha} = \Phi^{-1}(1-\alpha)$ is the $1-\alpha$ quantile of the standard normal distribution $\mathcal{N}$ under a \textit{significance level} of $\alpha$ or, equivalently, a \textit{confidence level} $(1 - \alpha) \cdot 100$\%.\footnote{\textcite{DBLP:conf/eaamo/AlvarezR23} contains a typo in the numerator of $w_{\alpha}$: we wrote a minus instead of a plus sign. In the code, however, it was implemented correctly. It also discusses a two-sided CI.}
The $+ \infty$ represents that there is no upper bound, as we are interested in values greater than $\tau$.
The choice of $\alpha$ and $\tau$, as with $k$, depends on the context of the discrimination claim. 
These parameters are motivated by legal requirements (set, e.g.,~by the court \parencite{Thanh_KnnSituationTesting2011}), or technical requirements (set, e.g.,~via power analysis \parencite{Cohen2013StatisticalPower}), or both.
A common choice for $\alpha$ is 0.05, though common alternatives are also 0.01 and 0.10. 

The CI represents a one-sided statistical test based on the hypothesis that there is individual discrimination, providing a measure of certainty on $\Delta p$ through a range of possible values.
Formally, 
let $\pi$ be the true difference in proportion of negative decision outcomes between the control and test groups. Then the \textit{null hypothesis} is $H_0: \pi = \tau$, while the \textit{alternative hypothesis} $H_1: \pi > \tau$.
When $\tau$ falls within the range of probable values in CI, we fail to reject $H_0$ with $\alpha$ significance level.
Given $\Delta p$ \eqref{eq:delta} and its CI \eqref{eq:CIs}, we can now proceed to define individual discrimination under CST.

\begin{remark}(Two Versions of CST)
\label{rem:SearchCenters}
    CST can include or exclude the search centers in \eqref{eq:delta} and \eqref{eq:CIs}. 
    If we exclude them, then both remain as is; 
    if we include them, then $\hat{y}_c$ and $\hat{y}_c^{CF}$ are counted in $p_c$ and $p_t$, leading to a denominator of $k + 1$ in both as well as in the $w_{\alpha}$ calculation.
    To distinguish between the two versions of CST, we will use CST w/o when excluding and CST w/ when including the search centers.
    We add this option for comparing CST against situation testing \parencite{Thanh_KnnSituationTesting2011}, which excludes the search centers, and counterfactual fairness \parencite{Kusner2017CF}, which only uses the search centers.
    We use this option extensively in Section~\ref{sec:Experiments}. 
\end{remark}
\begin{definition}[Individual Discrimination]
\label{def:IndDisc}
    There is (potential) individual discrimination toward the complainant $c$ if $\Delta p > \tau$, meaning the negative decision outcomes rate for the control group is greater than for the test group by some accepted deviation $\tau \in [-1, 1]$.
\end{definition}
\begin{definition}[Confidence on the Individual Discrimination Claim]
\label{def:CIs}
    A detected (potential) discrimination claim for the complainant $c$ by Definition~\ref{def:IndDisc}
    is statistically significant with significance level $\alpha$ if the CI
    excludes $\tau$.
\end{definition}

We highlight the use of the word \textit{potential} in both definitions. 
It implies, formally, that under CST, as with any individual or group discrimination testing framework \parencite{Romei2014MultiSurveyDiscrimination}, we test for \textit{prima facie} discrimination.
Uncovering discrimination amounts to a series of steps among which there is the need to provide evidence of the discrimination claim. 
Even if said evidence is found, it still needs to be argued for in court.
For a discussion on the EU discrimination testing pipeline, see \textcite{DBLP:conf/fat/WeertsXTOP23}.

\begin{remark}(Actionability)
\label{rem:actionability}
    The many-to-many comparison behind $\Delta p$ is what makes CST an actionable framework for testing individual discrimination. 
    The single comparison is not enough when proving \textit{prima facie} discrimination 
    as we want to ensure, one, that the individual claim is representative of the population, and two, be certain about the individual claim.
    Implicit to both concerns is finding a pattern of unfavorable decisions against the protected group of the complainant on which we are confident enough.
\end{remark}

The notion of repetition is important in Remark~\ref{rem:actionability}.
Similar to flipping a coin multiple times to uncover its (un)fairness, we expect a significant pattern of unfavorable decisions (read, discrimination) to emerge through ``repeating'' the decision-making process in question.
Such repetition is often not possible in practice. 
In a non-ADM setting we cannot ask the same female complainant in Example~\ref{ex:IllustrativeExample} to apply multiple times to the same bank;
we can, instead, look at other similar instances that underwent the same decision process.
Similarly, even when repetition is deterministic, such as entering the same input multiple times into the ADM, it is non-trivial to generalize that the individual case represents a group-wise pattern.
What rules out that the $\Delta p$ for complainant $c$ is an exception rather than a systematic effect?
We can, for instance, assume a theoretical distribution of comparisons with $\pi$ to account for potential randomness in what we detect from the single point estimate that is $\Delta p$.
The $p_c$ and $p_t$ help tackle these concerns.

\paragraph{On positive discrimination.}
Positive individual discrimination, or affirmative action, refers to the setting in which complainant $c$ is shown to be favored in the decision-making process \parencite{Romei2014MultiSurveyDiscrimination}.
Policies like diversity quotas are an example of positive discrimination.
We can operationalize positive discrimination easily under the current k-NN CST implementation: we would rewrite Definitions \ref{def:IndDisc} and \ref{def:CIs} by looking at the same complainant $c$ but focusing on the opposite effect.
Formally, we would consider $\Delta p < \tau$ (where now $\tau < 0$) and, in turn, the CI $(- \infty, \Delta p + w_\alpha]$ as we would test for the alternative hypothesis $H_1: \pi < \tau$. The rest of the k-NN CST pipeline would apply the same.

Positive discrimination remains understudied within algorithmic discrimination. 
We believe this is due to standard discrimination being more prevalent as a societal and research problem.
We also believe positive discrimination poses a different set of conceptual challenges over traditional discrimination, making it harder to justify by those in favor of it.\footnote{Take, e.g., the US Supreme Court's overturn of affirmative action \parencite{NPR2023AffirmativeAction}.}
This is, at least, our reading from the lack of discussion positive discrimination enjoys by the legal works we cite.
In short, although testing for positive discrimination under CST is straightforward, a clear legal narrative is lacking for us to comfortably operationalize it.
We formalize and test for positive discrimination in Appendices~\ref{Appendix.Supplements} and \ref{Appendix.AddExperiments}, respectively.
We do so mainly for illustrative purposes since we are focused on understanding traditional discrimination in its single and multidimensional forms.

\subsection{Connection to Counterfactual Fairness}
\label{sec:CST.OnCF}

There is a clear link between CST and counterfactual fairness (CF) of \textcite{Kusner2017CF}.
Recall that the ADM $b()$ is counterfactually fair if it outputs the same outcome for the factual tuple as for its counterfactual tuple, where the latter is generated through the abduction, action, and prediction steps when intervening the protected attribute $A$.\footnote{Formally, $P(\hat{Y}_{A \leftarrow a}(U)=y \, | \, X, A) = P(\hat{Y}_{A \leftarrow a'}(U)=y \, | \, X, A)$, where the left side is the factual $A=a$ and the right side the counterfactual $A=a'$. Similar with $\tau$ in $\Delta p$, the equality can be relaxed given some permissible difference threshold $\epsilon > 0$ between the factual and counterfactual quantities.} 
Hence, the factual $(x_c, a_c, \hat{y}_c)$ and counterfactual $(x_c^{CF}, a_c^{CF}, \hat{y}_c^{CF})$ tuples used in CST are also the ones used in CF for evaluating the counterfactual fairness for complainant $c$.

We view CST, when including the search centers, as an actionable extension of CF.
CST equips CF with CI \eqref{eq:CIs}, providing a certainty measure on the counterfactual fairness of $b()$. 
Previous works on CF have addressed uncertainty concerns \parencite{DBLP:conf/nips/RussellKLS17, DBLP:conf/uai/KilbertusBKWS19}, but with a focus on the structure of the SCM $\mathcal{M}$ and how that affects the measured unfairness of $b()$. 
We instead address certainty on the literal comparison that motivates the CF definition.
A key consequence of this link between CST and CF is that we can have an ADM $b()$ that is counterfactually fair but discriminatory. 
We summarize this point in Proposition~\ref{prop:ActioanbleCF}. 
We also present a sketch of proof for it.

\begin{proposition}[Actionable Counterfactual Fairness]
\label{prop:ActioanbleCF} 
    Counterfactual fairness does not imply nor it is implied by individual discrimination as conceived in Definition~\ref{def:IndDisc}.
\end{proposition}

We now present a sketch of proof to Proposition~\ref{prop:ActioanbleCF}.
Consider the factual tuple $(x_c, a_c=1, \widehat{y}_c=0)$ and assume the generated counterfactual is $(x_c^{CF}, a_c^{CF}=0, \widehat{y}_c^{CF}=0)$. 
Since $\widehat{y}_c = \widehat{y}_c^{CF}$, this is a case in which CF holds. However, the decision boundary of the ADM $b()$ can be purposely set such that the $k$-nearest neighbors of $x_c$ are all within the decision $\hat{Y}=0$, and less than $1-\tau$ fraction of the $k$-nearest neighbors of $x_c^{CF}$ are within the decision $\hat{Y}=0$. 
This leads to a $\Delta p > 1-(1-\tau) = \tau$, showing that there is individual discrimination. 
Similarly, the other way can be shown by assuming $\widehat{y}_c \neq \widehat{y}_c^{CF}$ but the sets of $k$-nearest neighbors have rates of negative decisions whose difference is smaller than $\tau$.

Proposition~\ref{prop:ActioanbleCF} alludes to the scenario in which $b()$ is counterfactually fair yet discriminatory. 
Intuitively, it is possible to handle \textit{borderline cases} where the tuple of interest and its counterfactual both get rejected by $b()$, though the latter is closer to the decision boundary than the former. 
The model $b()$ would be considered counterfactually fair, but would that disprove the individual discrimination claim? 
CST, by constructing the control and test groups around this single comparison, accounts for this actionability concern.

Importantly, for the purposes of Proposition~\ref{prop:ActioanbleCF} we consider the \textit{two-sided CI} over the previous one-sided CI \eqref{eq:CIs}.
We are interested in addressing the statistical significance of the ``counterfactual fairness claim'' using the neighborhoods built by the k-NN---note via the CST w/ version, cfr. Remark~\ref{rem:SearchCenters}---algorithm around the factual and counterfactual instances of CF.
We write such CI for $\Delta p$ as:
\begin{align}
\label{eq:CIsforCF}
    [\Delta p - w_{\alpha/2}, \Delta p +  w_{\alpha/2}],
    & \; \; \; \text{with} \; \; \;
    w_{\alpha / 2} =  z_{\alpha / 2} \sqrt{\frac{p_c(1 - p_c) + p_t(1 - p_t)}{k + 1}}
\end{align}
where $z_{\alpha / 2} = \Phi^{-1}(1-\alpha / 2)$ be the $1-\alpha / 2$ quantile of $\mathcal{N}$ under a \textit{significance level} of $\alpha$ or, equivalently, a \textit{confidence level} $(1 - \alpha) \cdot 100$\%.
We use \eqref{eq:CIsforCF} when addressing CF as a whole, and use \eqref{eq:CIs} when addressing the discrimination claim through CF.


\subsection{The Multidimensional Setting}
\label{sec:CST.Multi}

Let us revisit the previous definitions under multidimensional discrimination. 
It occurs when $|A| > 1$.
Following the legal literature \parencite{Xenidis2020_TunningEULaw}, we distinguish two forms of multidimensional discrimination: multiple and intersectional.

\begin{definition}[Multiple Discrimination]
\label{def:MultipleDisc}
    Under Definition~\ref{def:IndDisc}, there is (potential) multiple individual discrimination toward the complaint $c$ with the set of $|A| = q > 1$ protected attributes, if $\Delta p > \tau$ for each $\{A_i\}_{i=1}^{q}$ protected attribute.
\end{definition}
\begin{definition}[Intersectional Discrimination]
\label{def:IntersectionaleDisc}
    Under Definition~\ref{def:IndDisc}, there is (potential) intersectional individual discrimination toward the complaint $c$ with the set of $|A| = q > 1$ protected attributes, if $\Delta p > \tau$ for the attribute $A^* = \mathbbm{1}\{ A_1=1 \wedge A_2=1 \wedge \dots \wedge A_q=1 \}$ obtained by the intersection of the protected attributes.
\end{definition}

Regarding Definition~\ref{def:CIs}, meaning the confidence on the individual claim under multiple and intersectional discrimination, notice that both Definitions \ref{def:MultipleDisc} and \ref{def:IntersectionaleDisc} work with $\Delta p$ point estimates. Definition~\ref{def:MultipleDisc} looks at $q$ deltas for $c$ given the $q$ protected attributes, while Definition~\ref{def:IntersectionaleDisc} looks at a single delta for the attribute obtained by the intersection for $c$ of all the protected attributes. For intersectional discrimination, the single intersection delta must be statistically significant given $\alpha$. For multiple discrimination, instead, all the multiple deltas must be statistically significant given $\alpha/q$. The \textit{Bonferroni correction factor} $1/q$ counteracts the well-known multiple comparisons problem, allowing to test for a family-wise error rate of $\alpha$ in a set of $q$ (possibly, dependent) statistical tests.
We assume the same $\tau$ in both cases, in particular, with $\tau$ being the same irrespective of the protected attribute considered under multiple discrimination.

The difference between Definition~\ref{def:MultipleDisc} and Definition~\ref{def:IntersectionaleDisc} is subtle but central to detecting \textit{prima facie} individual discrimination.
In multiple discrimination, we require $c$ to be discriminated \textit{separately} $q$ times as a member of each protected attribute it belongs to: e.g.,~as a female and as a non-white individual.
In intersectional discrimination, instead, we require $c$ to be discriminated \textit{simultaneously} as a member of all the $q$ protected attributes it belongs to: e.g.,~as a female-non-white individual.
As we discuss below, this distinction has clear modeling implications.
The tension between these types of multidimensional discrimination occurs as it is possible for $c$ not to suffer multiple discrimination while suffering intersectional discrimination \parencite{Crenshaw1989_DemarginalizingTheIntersection}. 
This is, in particular, troubling as only the former is recognized under EU non-discrimination law \parencite{Xenidis2020_TunningEULaw}. 

For the present k-NN CST implementation we operationalize the two forms of multidimensional discrimination as follows:
\begin{itemize}
    \item For multiple discrimination we run CST separately for each $A_i$, including the generation of the corresponding counterfactual datasets via each $do(A_i := 0)$; and look for individual cases in which discrimination is detected across all runs.
    \item For intersectional discrimination we create the \textit{intersectional protected attribute} $A^*$ as in Definition~\ref{def:IntersectionaleDisc}; generate the corresponding counterfactual dataset via $do(A^* := 0)$; and run a single CST as we would for $|A|=1$.
\end{itemize}
Beyond using Definitions \ref{def:MultipleDisc} and \ref{def:IntersectionaleDisc}, respectively, both procedures have implications on Section~\ref{sec:CST_ControlTest}.
For multiple discrimination, we repeat Definitions~\ref{def:SearchSpaces}, \ref{def:CounterDataset}, and \ref{def:SearchCenters} for each of the $q$ protected attributes.
For intersectional discrimination, once we have generated $A^{*}$, we apply only once Definitions~\ref{def:SearchSpaces}, \ref{def:CounterDataset}, and \ref{def:SearchCenters} for this ``new'' protected attribute.
Section~\ref{sec:Experiments.Real} showcases both of these procedures for testing multidimensional discrimination.
Additionally, under these two producers we can also explore positive discrimination for multiple and intersectional discrimination using k-NN CST. We would simple revisit, respectively, Definitions \ref{def:MultipleDisc} and \ref{def:IntersectionaleDisc} by considering the opposite effect, meaning the relevant delta(s) being less than $\tau$.
We leave this for future work.

%
%

\section{Experiments}
\label{sec:Experiments}
We now showcase CST using its k-NN implementation (k-NN CST).
Throughout this section, we compare it to its standard counterpart (k-NN ST) of \textcite{Thanh_KnnSituationTesting2011} and counterfactual fairness (CF) of \textcite{Kusner2017CF}.
Henceforth, we drop the ``k-NN'' for CST and ST. 
As noted in Remark~\ref{rem:SearchCenters}, we exclude the search centers when comparing CST to ST (i.e., CST w/o) and include them (i.e., CST w/) when comparing CST to CF.\footnote{The code is available in this repository: \href{https://github.com/cc-jalvarez/counterfactual-situation-testing}{https://github.com/cc-jalvarez/counterfactual-situation-testing}.}

\subsection{Setup}
\label{sec:Experiments.SetUp}

We use a significance level of $\alpha=0.05$, an accepted deviation of $\tau=0.0$, and the neighborhood sizes of $k \in \{ 15, 30, 50, 100, 250 \}$. In practice, we expect these parameters to be provided by the legal context. For instance, $k=30$ is a common size used in non-algorithmic situation testing \parencite{Rorive2009_ProvingDiscrimination}. See Appendix~\ref{Appendix.AddExperiments} for additional experiments.

Section~\ref{sec:Experiments.IllustrativeExample} focuses on comparing CST to ST and CF, while Section~\ref{sec:Experiments.Real} focuses on illustrating CST for multidimensional discrimination testing.
We use synthetic and real data, each describing high-stakes decision-making scenarios involving an ADM $b()$.
We assume a corresponding SCM $\mathcal{M}$ and DAG $\mathcal{G}$ for each scenario.
$\mathcal{M}$ and $\mathcal{G}$ are only needed for CST and CF.
Similar to $\alpha$, $\tau$, and $k$, we expect $\mathcal{M}$ and $\mathcal{G}$ to be provided.
The assumptions on $\mathcal{M}$ and $\mathcal{G}$ in Section~\ref{sec:CausalKnowledge} simplify the abduction step when generating $\mathcal{D}^{CF}$. We stress that these assumptions are convenient but not necessary.

\subsection{An Illustrative Example}
\label{sec:Experiments.IllustrativeExample}

\begin{figure}[t]
    \centering
    \begin{subfigure}{.45\linewidth}
    \includegraphics[scale=0.45]{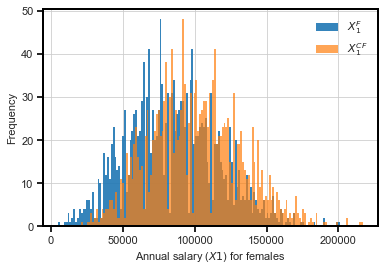}
    \end{subfigure}
    \hfill
    \begin{subfigure}{.45\linewidth}
    \includegraphics[scale=0.45]{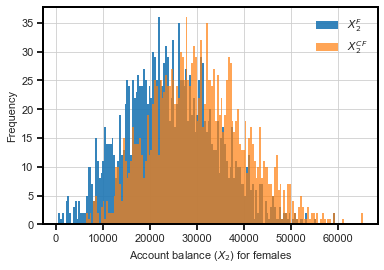}
    \end{subfigure}
\caption{Distribution of the attributes used by the ADM $b()$ in Section~\ref{sec:Experiments.IllustrativeExample}. The histograms compare the $\mathcal{D}$ and $\mathcal{D}^{CF}$ datasets for the female applicants. The shift to the right of both $X_1^{CF}$ and $X_2^{CF}$ shows the negative impact of $A$ on $X_1$ and $X_2$.}
\label{fig:LoanApp_Distributions}
\end{figure}

Let us consider the loan application scenario introduced in Example~\ref{ex:IllustrativeExample}.
We create the synthetic dataset $\mathcal{D}$ based on Figure~\ref{fig:KarimiV2}. 
It is a modified version of~\textcite[Figure 1]{Karimi2021_AlgoRecourse}. 
We add the protected attribute gender ($A$), which affects an applicant's annual salary ($X_1$) and bank balance ($X_2$) used by the bank's ADM $b()$ for approving ($\widehat{Y}=1$) or rejecting ($\widehat{Y}=0$) a loan application.
We generate $\mathcal{D}$ for $n=5000$ under $A \sim \text{Ber}(0.45)$ with $A=1$ if the individual is female and $A=0$ otherwise.
We define $f_1$ as $X_1 \leftarrow (-\$1500) \cdot \text{Poi}(10) \cdot A + U_1$ and $f_2$ as $X_2 \leftarrow (-\$300) \cdot \mathcal{X}^2(4) \cdot A + (3/10) \cdot X_1 + U_2$, assuming $U_1 \sim \$10000 \cdot \text{Poi}(10)$ and $U_2 \sim \$2500 \cdot \mathcal{N}(0, 1)$.
We define $b()$ as $\widehat{Y} = \mathbbm{1}\{ X_1 + 5 \cdot X_2 > \$225000\}$.
Importantly,
with Figure~\ref{fig:KarimiV2} we have access to the data generating model of $\mathcal{D}$.
It allows to study CST when we are able to control for potential model misspecifications in the SCM $\mathcal{M}$. 
$\mathcal{D}$ represents a biased scenario in which female applicants are penalized systematically in $X_1$ and $X_2$. Such penalties, e.g.,~capture the financial burdens female professionals face in the present after having been discouraged in the past from pursuing high-paying, male-oriented jobs \parencite{CriadoPerez2019InvisibleWomen}.
$\mathcal{D}$ represents a non-neutral status quo and, thus, a setting of interest to indirect non-discrimination law \parencite{Wachter2020BiasPreserving}.

We study the behavior of $b()$ toward $A$ in $\mathcal{D}$.
We use CST, ST, and CF to answer whether $b()$ discriminates against female loan applicants.
For CST and CF, we generate $\mathcal{D}^{CF}$ using Figure~\ref{fig:KarimiV2} based on the intervention $do(A:=0)$, or \textit{what would have happened had all loan applicants been male}? 
Comparing $\mathcal{D}$ to $\mathcal{D}^{CF}$ already highlights the unwanted systematic effects of $A$. 
These effects can be seen in Figure~\ref{fig:LoanApp_Distributions} by the rightward shift experienced in $X_1$ and $X_2$ for all female applicants when going from the factual to the counterfactual world.
$\mathcal{D}$ has a total of 1712 females (34.29\%) and, overall, $b()$ has an acceptance rate of 53.5\%.
Using \textit{demographic parity} (DP) \parencite{Kamiran2009_ClassifyingWihtoutDiscriminating}, we also observe the unfairness of $b()$ with 13.5\% acceptance probability for female and 40\% for male applicants.\footnote{\label{foot:dp}Formally, we define DP as $P(\hat{Y}|A=1) = P(\hat{Y}|A=0)$. We only use DP as we do not have access to the ground-truth $Y$ and cannot use, e.g.,~equalized odds \parencite{DBLP:conf/nips/HardtPNS16}.}

Table~\ref{table:k-results} reports the results for all methods based on Definitions~\ref{def:IndDisc} and \ref{def:CIs}.
Under Definition~\ref{def:IndDisc}, we detect individual discrimination when the complainant's $\Delta p > \tau$; and under Definition~\ref{def:CIs}, we detect individual discrimination that is statistically significant given $\alpha$ when $\Delta p > \tau$ and $\tau$ does not fall within $\Delta p$'s CI.
Figures~\ref{fig:CSTwoVsST_k_param}
and \ref{fig:TwoCSTs_k_param} further show how the results change under these definitions for all methods as $k$ increases, including cases beyond $k=250$.
Let us discuss these results in detail.

\begin{table}[t]
\caption{Number and (\% w.r.t. females) of individual discrimination cases based on $A$ in Figure~\ref{fig:KarimiV2}. Marked by * are the number of statistically significant cases.}
  \label{table:k-results}
  \centering
  \begin{tabular}{cccccc}
    \toprule
    Method & $k=15$ & $k=30$ & $k=50$ & $k=100$ & $k=250$\\
    \midrule
    CST w/o & 288 (16.8\%) & 313 (18.3\%) & 342 (20\%) & 395 (23.1\%)  & 534 (31.2\%) \\
     & 272* (15.9\%) & 306* (17.9\%) & 331* (19.3\%) & 383* (22.4\%)  & 519* (30.3\%) \\
     \midrule
    ST & 55 (3.2\%) & 65 (3.8\%) & 84 (5\%) & 107 (6.3\%) & 204 (11.9\%) \\
    & 44* (2.6\%) & 57* (3.3\%) & 65* (3.8\%) & 85* (5\%) & 148 (8.6\%) \\
    \midrule
    CST w/ & 420 (24.5\%) & 434 (25.4\%) & 453 (26.5\%) & 480 (28\%)  & 557 (32.5\%)\\
    & 272* (15.9\%) & 307* (17.9\%) & 334* (19.5\%) & 385* (22.5\%)  & 520 (30.4\%)\\
    \midrule
    CF &  376 (22\%) &  376 (22\%) &  376 (22\%) & 376 (22\%)  & 376 (22\%) \\
    & 241* (14.1\%) &  253* (14.8\%) & 265* (15.5\%) & 288* (16.8\%)  & 352* (20.6\%) \\
    \bottomrule
  \end{tabular}
\end{table}

\subsubsection{CST Relative to ST}
\label{sec:Experiments.IllustrativeExample.CSTvST}

We consider CST w/o as ST excludes the search centers when testing for individual discrimination. What is clear from Table~\ref{table:k-results} is that CST w/o detects more cases than ST across all $k$ sizes, even when accounting for statistical significance. The results highlight the impact of operationalizing \textit{fairness given the difference} since the only difference between the two methods is the choice of the test search center. 
ST performs an idealized comparison. If we consider the tuple $(x_1=35000, x_2=7948, a=1)$ as complainant $c$, then ST builds the test group by finding the $k$ most similar male profiles under $d$ \eqref{eq:Distance} using $c$ as the test search center. CST w/o, conversely, performs a more flexible comparison under \textit{fairness given the difference}. For the same $c$ and $d$ \eqref{eq:Distance}, it instead uses the counterfactual tuple $(x_1^{CF}=50796, x_2^{CF}=13852, a^{CF}=0)$ as the test search center. The control group is the same for both ST and CST as each uses $c$ as the control search center.

\begin{figure}[t]
    \centering
    \begin{subfigure}{.45\linewidth}
    \includegraphics[scale=0.45]{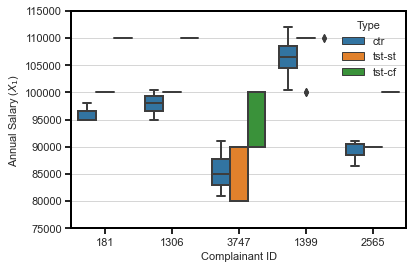}
    \end{subfigure}
    \hfill
    \begin{subfigure}{.45\linewidth}
    \includegraphics[scale=0.45]{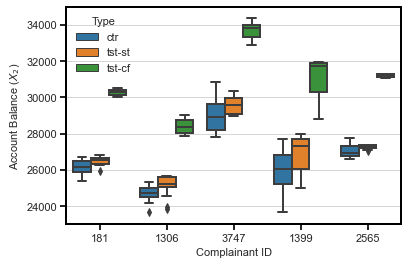}
    \end{subfigure}
\caption{The boxplots for a subset of cases detected by ST and CST w/o for $k=15$. We compare the ST and CST w/o control group (ctr) versus the ST (tst-st) and CST w/o (tst-cf) test groups. 
The tst-st are closer to ctr than tst-cf, illustrating the \textit{fairness given the difference} behind CST and the \textit{idealized comparison} behind ST.}
\label{fig:LoanApp_BoxPlots}
\end{figure}

Figure~\ref{fig:LoanApp_BoxPlots} shows the impact of \textit{fairness given the difference}.
We randomly choose five complainants that are discriminated by $b()$ according to ST and CST w/o for $k=15$, and plot the distributions of $X_1$ and $X_2$ as boxplots for the control group (ctr), ST test group (tst-st), and CST w/o test group (tst-cf).
The tst-cf are above the tst-st boxplots, indicating male profiles that are likelier to be accepted by $b()$. 
As all 55 ST cases for $k=15$ are included in the 288 CST w/o cases, the difference in test groups explains the difference in the number of cases between the methods.
Tables~\ref{table:k-results_for_STandCST} and \ref{table:k-results_for_CSTnotinST} illustrate this point.

\begin{table}[t]
\caption{Summary statistics for cases detected by ST and CST w/o for $k=15$. We present the averages for the control groups (ctr's) and the corresponding test groups (tst's).}
  \label{table:k-results_for_STandCST}
  \centering
  \begin{tabular}{cccc}
    \toprule
    Average of & ctr's for ST and CST w/o & tst's for ST & tst's for CST w/o \\
    \midrule
    Num. of neg. $\hat{Y}$ & 8.82 & 2.22 & 0.00 \\
    Prp. of neg. $\hat{Y}$ & 0.59 & 0.15 & 0.00 \\
    Avg. Annual salary & 94372.12 & 96181.82 & 106569.70 \\
    Std. Annual salary & 1646.29 & 611.57 & 286.34 \\
    Avg. Account balance & 26092.93 & 26347.46 & 30141.28 \\
    Std. Account balance & 558.30 & 352.78 & 273.45 \\
    \bottomrule
  \end{tabular}
\end{table}
\begin{table}[t]
\caption{Summary statistics, similar to Table~\ref{table:k-results_for_STandCST} but for all cases detected by CST w/o only. We include the corresponding ST test groups for comparison.}
  \label{table:k-results_for_CSTnotinST}
  \centering
  \begin{tabular}{cccc}
    \toprule
    Average of & ctr's for ST and CST w/o & tst's for ST & tst's for CST w/o \\
    \midrule
    Num. of neg. $\hat{Y}$ & 13.74 & 14.79 & 0.78 \\
    Prp. of neg. $\hat{Y}$ & 0.92 & 0.99 & 0.05 \\
    Avg. Annual salary & 86332.47 & 85911.30 & 96898.43 \\
    Std. Annual salary & 1524.33 & 906.11 & 391.49 \\
    Avg. Account balance & 24734.11 & 24790.05 & 28677.11 \\
    Std. Account balance & 482.15 & 152.07 & 171.61 \\
    \bottomrule
  \end{tabular}
\end{table}

Table~\ref{table:k-results_for_STandCST} reports the average summary statistics for the cases detected by ST and CST w/o for $k=15$.
Notice that (the average of) the average and standard deviation for $X_1$ and $X_2$ are similar between the ST control and test groups, while such similarity does not occur between the CST w/o groups.
The CST w/o test groups show a higher annual salary and account balance than their ST counterparts. 
It translates into a lower average number and proportion of negative $\hat{Y}$ as these male profiles are likelier to be accepted by $b()$.
There is still a clear difference in outcomes between the control groups and the ST and CST w/o test groups, which leads to both methods detecting these cases.

Table~\ref{table:k-results_for_CSTnotinST} reports the same summary statistics but for cases detected by CST w/o only. 
For comparison, we include the corresponding ST test groups. 
The groups for ST are again similar (and lower in average values) between them in terms of $X_1$ and $X_2$, but are also similar in terms of the number of negative $\hat{Y}$, explaining why ST does not detect these cases. 
The CST w/o test groups, instead, exhibit almost no cases of negative $\hat{Y}$. 
Intuitively, 
ST is comparing females and males with similar applicants not suitable for a loan while CST w/o is comparing non-suitable female to suitable male applicants.
This difference between ST and CST w/o is stark because $\mathcal{D}$ is purposely generated with a systematic bias against female applicants. 
We expect the CST w/o test groups to be more suitable than their ST counterparts as the former's test search centers account for the downstream effects of $A$ on $X_1$ and $X_2$ while the latter's test search centers ignore any effect at all.

\begin{figure}[t]
    \centering
    \begin{subfigure}{.32\linewidth}
    \includegraphics[scale=0.345]{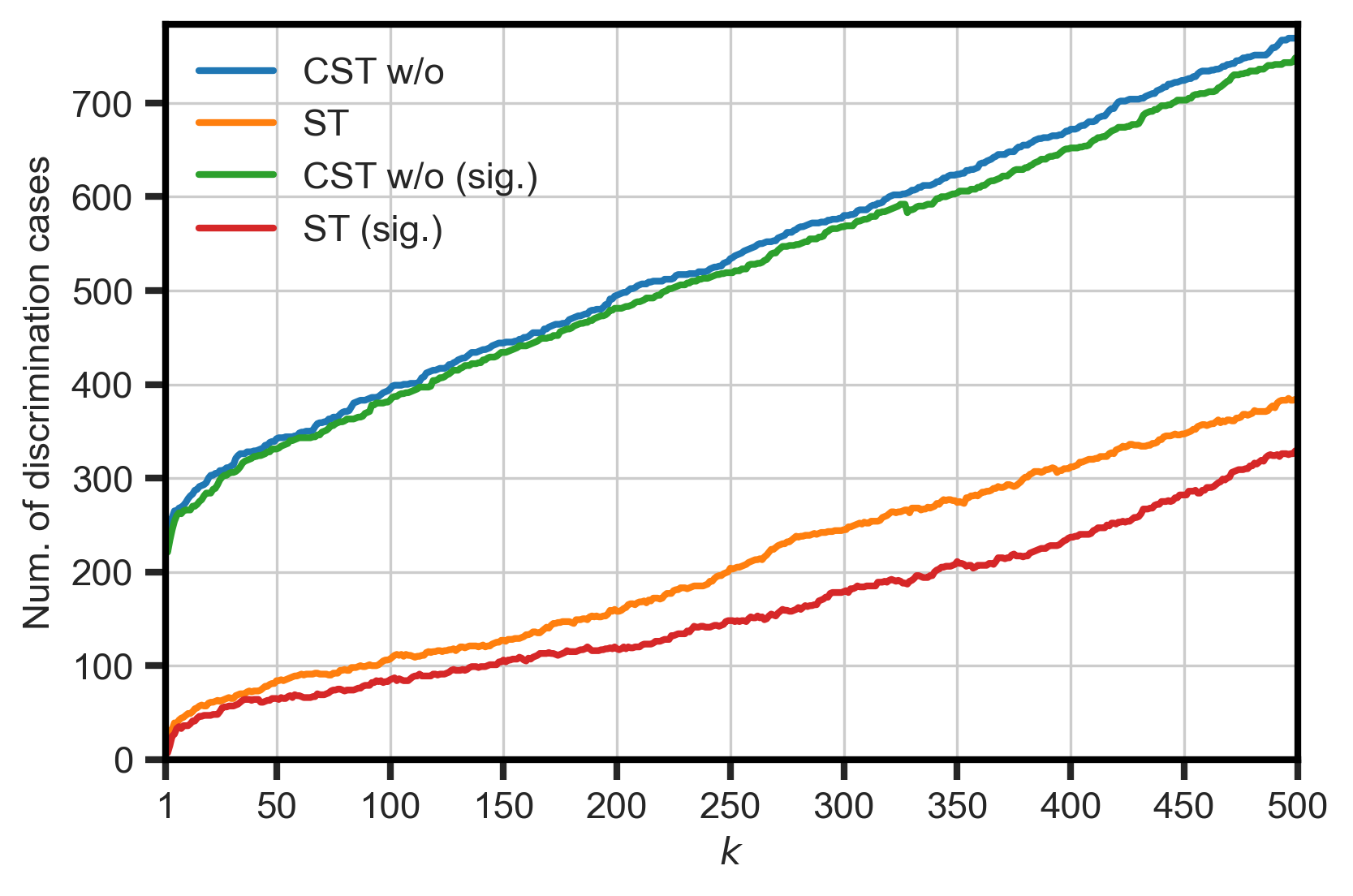}
    \end{subfigure}
    \hfill
    \begin{subfigure}{.32\linewidth}
    \includegraphics[scale=0.345]{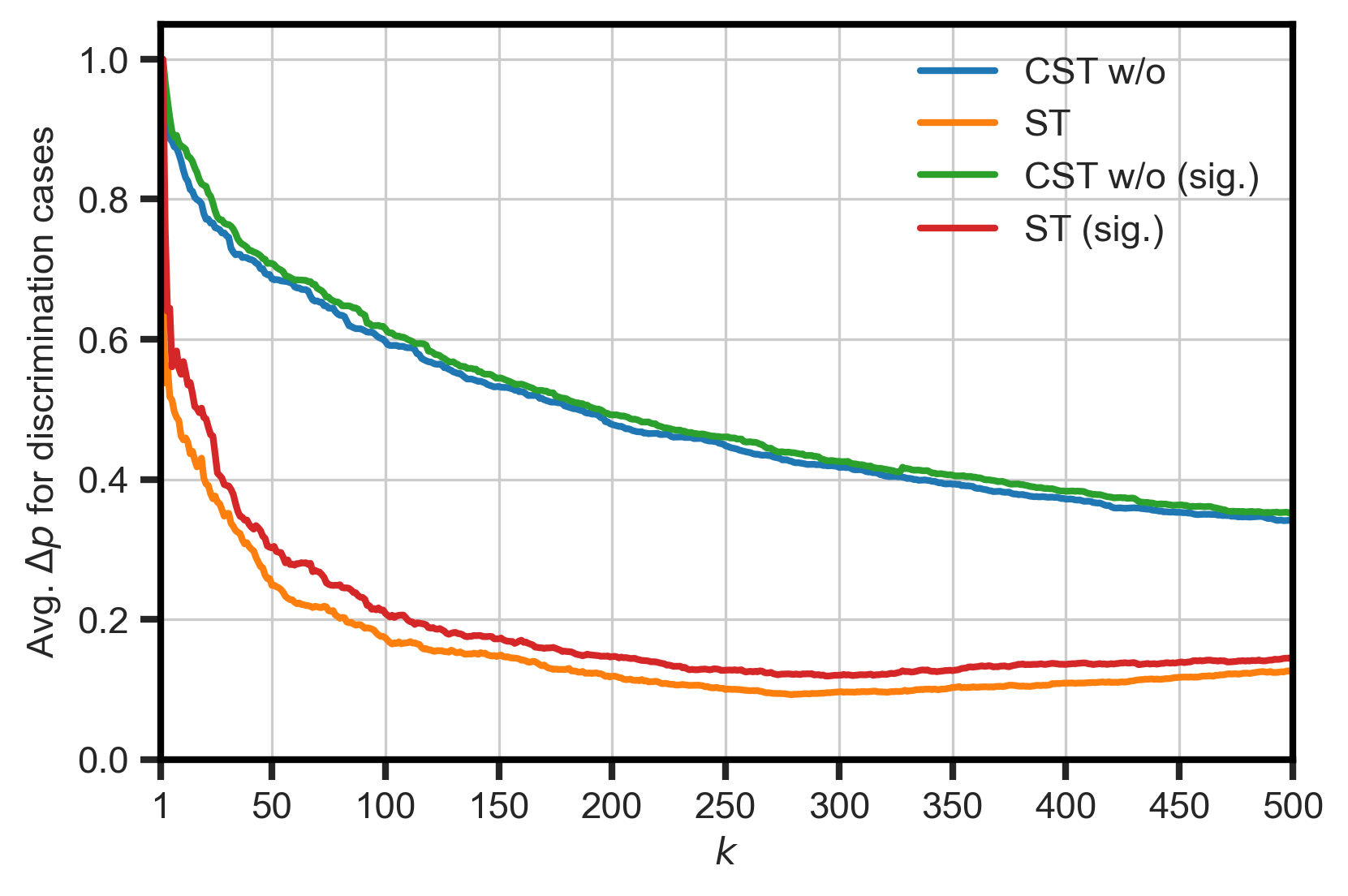}
    \end{subfigure}
    \hfill
    \begin{subfigure}{.32\linewidth}
    \includegraphics[scale=0.345]{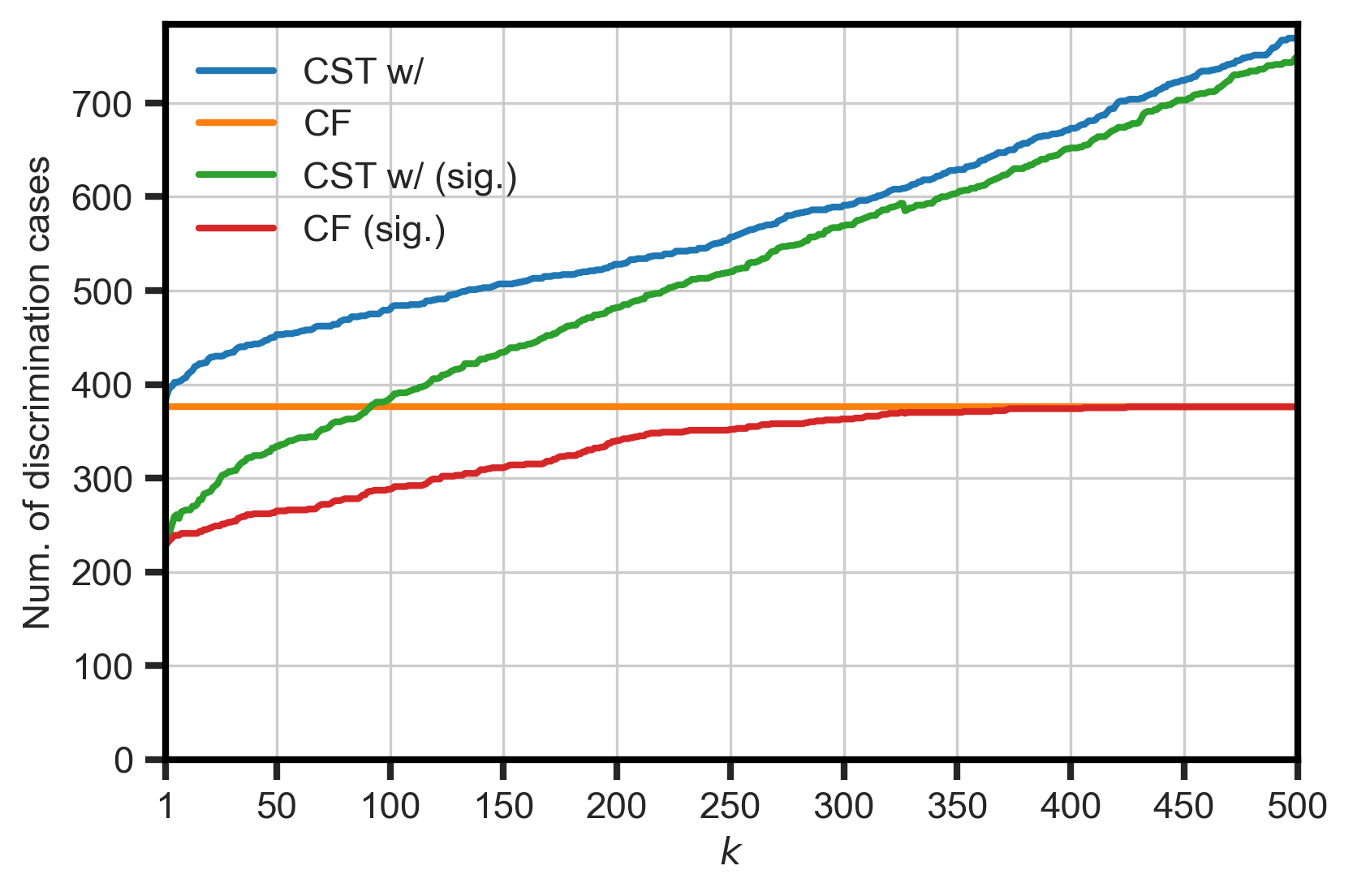}
    \end{subfigure}
\caption{Left and center: Number of cases by CST w/o and ST and their avg. $\Delta p$. Right: Number of cases by CST w/ and CF. We plot all and statistically significant (sig.) cases.}
\label{fig:CSTwoVsST_k_param}
\label{fig:CSTwiVsCF_k_param}
\end{figure}

Figure~\ref{fig:CSTwoVsST_k_param} illustrates how CST w/o and ST differ in terms of number of discrimination cases and $\Delta p$ up until $k=500$ for all and statistically significant cases.
In both plots, CST w/o and ST follow a similar trend and with signs of a slow convergence toward the other.
The impact of the \textit{mutatis mutandis} over the \textit{ceteris paribus} manipulation occurs on smaller $k$'s and persists over larger $k$'s, clearly differentiating CST w/o from ST and its operationalization of \textit{fairness given the difference} for sensible ranges of $k$.
The higher number of cases by CST w/o over ST (left) is driven by test groups that are, on average, dissimilar to the control groups, leading to a larger average $\Delta p$ by CST w/o over ST (center).
Both methods show similar trends when considering all and statistically significant cases, meaning the difference between CST w/o and ST is statistically significant.
The results validate the use of CST w/o over ST as well as the viability of the \textit{mutatis mutandis} over the \textit{ceteris paribus} manipulation for testing discrimination.


\subsubsection{CST Relative to CF}
\label{sec:Experiments.IllustrativeExample.CSTvCF}

We consider the CST w/ as CF uses the search centers for measuring the counterfactual fairness of $b()$.
Back to the illustrative factual and counterfactual tuples of $c$, we now include the control search center $(x_1=35000, x_2=7948, a=1)$ and the test search center $(x_1^{CF}=50796, x_2^{CF}=13852, a^{CF}=0)$ into the k-NN neighborhoods to be compared for $c$. 
These are the two instances we compare under CF to test for the counterfactual fairness of $b()$.
For comparison, we define CF discrimination as a case in which the factual $\widehat{y}_c=0 \in \mathcal{D}$ (from negative outcome) becomes $\widehat{y}_c^{CF}=1 \in \mathcal{D}^{CF}$ (to positive outcome) after the intervention on $A$.
This definition aligns with using $\tau=0.0$ for CST w/.


Table~\ref{table:k-results} shows how CST w/ detects for each $k$ a higher number of cases than CF, even when accounting for statistical significance.
In fact, all cases detected by CF are also detected by CST w/.
CF is independent from $k$ as it applies only to the individual comparison of the factual and counterfactual tuples for a given $c$.
However, the CI used for measuring the statistical significance of CF discrimination is a function $k$ and, thus, varies with it (cfr.,~\eqref{eq:CIs} and \eqref{eq:CIsforCF}).
As $k$ increases, so does the number of statistically significant CF and CST w/ individual discrimination cases.
Both CST w/ and CF use the one-sided CI \eqref{eq:CIs} as we test for $\Delta p > \tau$.
Figure~\ref{fig:CSTwiVsCF_k_param} (right) further shows how CST w/ and CF evolve as $k$ increases up to $k=500$ in terms of the number of cases detected.
It aligns with Table~\ref{table:k-results}. 
We do not plot the average $\Delta p$ as CF discrimination focuses on the literal comparison of the factual and counterfactual instances of $c$.
Further, the number of significant CF cases is bounded by all CF cases. 
As long as CST w/ detects more cases than CF, it appears that all CF cases become statistically significant under a large enough $k$. 

What sets CST w/ apart from CF is twofold.
First, CST w/ equips the CF comparison with certainty measures.
Table~\ref{table:CFwithCIs} shows individual cases of discrimination detected by both CF and CST w/ for $k=15$ along with the one-sided \eqref{eq:CIs} and two-sided \eqref{eq:CIsforCF} CI.
The latter CI is specific to CF and it is built through the CST w/ k-NN implementation. 
Beyond the CF discrimination claim, this CI provides a measure of certainty to the factual versus counterfactual comparison behind CF.
Hence, CST complements CF beyond detecting discrimination.
Second, CST w/ detects cases of individual discrimination that are counterfactually fair. 
In Table~\ref{table:NotinCFwithCIs} reports cases for $k=15$ that do not exhibit CF discrimination but exhibit a discriminatory pattern when looking at $\Delta p$.
The results highlight the importance of requiring multiple comparisons to insure that the complainant's experience is representative of the discrimination claim.

\begin{table}[t]
  \caption{Subset of cases detected by both CST w/ and CF for $k=15$. The * denotes statistical significance based on the one-sided CI.}
  \label{table:CFwithCIs}
  \centering
  \begin{tabular}{cccccc}
    \toprule
    Comp. (ID) & $p_c$ & $p_t$ & $\Delta p$ & CI \eqref{eq:CIs} & CI \eqref{eq:CIsforCF} \\
    \midrule
    44  & 1.00 & 0.00 & 1.00* & [1.00, +$\infty$) & [1.00, 1.00] \\
    55  & 0.81 & 0.00 & 0.81* & [0.65, +$\infty$) & [0.62, 1.00] \\
    150 & 1.00 & 0.94 & 0.06 & [-0.04, +$\infty$) & [-0.06, 0.18] \\
    203 & 1.00 & 0.87 & 0.13 & [-0.01, +$\infty$) & [-0.04, 0.29] \\
    218 & 0.56 & 0.00 & 0.56* & [0.36, +$\infty$) & [0.32, 0.81] \\
    \bottomrule
  \end{tabular}
\end{table}
\begin{table}[t]
  \caption{Subset of cases detected by CST w/ and not CF for $k=15$. The * denotes statistical significance based on the one-sided CI.}
  \label{table:NotinCFwithCIs}
  \centering
  \begin{tabular}{cccccc}
    \toprule
    Comp. (ID) & $p_c$ & $p_t$ & $\Delta p$ & CI \eqref{eq:CIs} & CI \eqref{eq:CIsforCF} \\
    \midrule
    5 & 0.06 & 0.00 & 0.06 & [-0.04, +$\infty$) & [-0.06, 0.18] \\
    147  & 0.50 & 0.00 & 0.5* & [0.29, +$\infty$) & [0.26, 0.75] \\
    435  & 0.38 & 0.00 & 0.38* & [0.18, +$\infty$) & [0.14, 0.61] \\
    1958 & 0.13 & 0.00 & 0.13 & [-0.01, +$\infty$) & [-0.04, 0.29] \\
    2926 & 0.75 & 0.00 & 0.75* & [0.57, +$\infty$) & [0.54, 0.96] \\
    \bottomrule
  \end{tabular}
\end{table}

Tables~\ref{table:k-results_for_CFvsCST_InBoth} and \ref{table:k-results_for_CFvsCST_InCSTOnly} report how a counterfactually fair $b()$ still discriminates according to CST w/.
As argued in Section~\ref{sec:CST.OnCF}, it occurs when considering a complainant and its counterfactual that are close to and on the same side of the decision boundary of $b()$.
When building the neighborhoods, CST w/ may include profiles from both sides of the decision boundary.
Table~\ref{table:k-results_for_CFvsCST_InBoth} summarizes the group characteristics of the cases detected by CF and CST w/.
If we consider (the average of) the average $X_1$ and $X_2$ for control and test groups, we obtain that $b()$ rejects the former and accept the latter, meaning it is not counterfactually fair.
Table~\ref{table:k-results_for_CFvsCST_InCSTOnly}, instead, summarizes cases detected by CST w/. 
For those cases, using again (the average of) the average $X_1$ and $X_2$ for control and test groups, $b()$ accepts both.

%
\begin{table}[t]
\caption{Summary statistics for cases detected by both CF and CST w/ under $k=15$. We present averages for the control groups (ctr’s) and the test groups (tst’s).}
  \label{table:k-results_for_CFvsCST_InBoth}
  \centering
  \begin{tabular}{ccc}
    \toprule
    Average of & ctr's CST w/ & tst's for CST w/ \\
    \midrule
    Num. of neg. $\hat{Y}$ & 15.46 & 5.75 \\
    Prp. of neg. $\hat{Y}$ & 1.03 & 0.38 \\
    Avg. Annual salary & 83281.25 & 94323.46 \\
    Std. Annual salary & 1410.97 & 1564.37 \\
    Avg. Account balance & 23752.64 & 27762.60 \\
    Std. Account balance & 449.85 & 550.18 \\
    \bottomrule
  \end{tabular}
\end{table}
\begin{table}[t]
\caption{Summary statistics for cases detected by CST w/ under $k=15$ and not CF. We present averages for the control groups (ctr’s) and the test groups (tst’s).}
  \label{table:k-results_for_CFvsCST_InCSTOnly}
  \centering
  \begin{tabular}{ccc}
    \toprule
    Average of & ctr's CST w/ & tst's for CST w/ \\
    \midrule
    Num. of neg. $\hat{Y}$ & 5.23 & 0.00 \\
    Prp. of neg. $\hat{Y}$ & 0.35 & 0.00 \\
    Avg. Annual salary & 92892.05 & 104541.25 \\
    Std. Annual salary & 1592.33 & 1434.18 \\
    Avg. Account balance & 26890.63 & 31161.38 \\
    Std. Account balance & 509.61 & 545.73 \\
    \bottomrule
  \end{tabular}
\end{table}

\subsubsection{CST w/o and CST w/}
\label{sec:Experiments.IllustrativeExample.BothCST}

Finally, we consider the two CST versions.
Three patterns are clear in Table~\ref{table:k-results}.
First, CST w/ detects a higher number of individual discrimination cases than CST w/o for all $k$.
Second, this difference decreases between the CST versions as $k$ increases.
Third, CST w/ and CST w/o detect roughly the same number of cases when accounting for statistical significance.
Let us explore these patterns.
Note that $\tau = 0.0$; any deviation of $p_c$ from $p_t$ (read, $\Delta p$) is detected by CST. 
Such deviation is statistically significant (read, representative) depending on the composition of the control and test groups behind $p_c$ and $p_t$. 

What differentiates CST w/o from w/ is the exclusion of the search centers (Remark~\ref{rem:SearchCenters}), making the size of the neighborhoods under CST w/o of size $k$ and under CST w/ of size $k+1$. 
Back to the illustrative factual and counterfactual tuples of $c$, the CST w/o control and test group are the same as the CST w/ ones with the key difference that the CST w/ control group has a \textit{(k+1)-th} that is $(x_1=35000, x_2=7948, a=1)$ and the test group has a \textit{(k+1)-th} that is $(x_1^{CF}=50796, x_2^{CF}=13852, a^{CF}=0)$.
The additional tuple pair compared in CST w/ is, notably, the most important pair
as the complainant and its counterfactual denote the closest possible comparison under CST.
Clearly, the inclusion (or exclusion) of the complainant-counterfactual pair into $\Delta p$ of CST w/ (or CST w/o) drives the statistically insignificant difference in the number of detected cases. 
Based on Table~\ref{table:k-results}, such pair is able to deviate $\Delta p$ from $\tau$ but is not representative of the corresponding control and test neighborhoods. 

\begin{table}[t]
\caption{Summary statistics for CST w/o cases for $k=15$ for the control (ctr’s) and test groups (tst’s) split by statical significance, which we denote with *.}
  \label{table:CST_without_k15}
  \centering
  \begin{tabular}{ccccc}
    \toprule
    Average of & ctr's & tst's & ctr's* & tst's* \\
    \midrule
    Num. of neg. $\hat{Y}$ & 3.94 & 2.62 & 13.32 & 0.51 \\
    Prp. of neg. $\hat{Y}$ & 0.26 & 0.17 & 0.89 & 0.03 \\
    Avg. Annual salary & 92808.34 & 104750.00 & 87577.21 & 98392.16 \\
    Std. Annual salary & 1919.18 & 1182.73 & 1525.77 & 323.69 \\
    Avg. Account balance & 25923.28 & 30121.55 & 24938.92 & 28888.21 \\
    Std. Account balance & 652.67 & 290.97 & 487.52 & 185.18 \\
    \bottomrule
  \end{tabular}
\end{table}
\begin{table}[t]
\caption{Summary statistics for CST w/ cases for $k=15$ for the control (ctr’s) and test groups (tst’s) split by statical significance, which we denote with *.}
  \label{table:CST_with_k15}
  \centering
  \begin{tabular}{ccccc}
    \toprule
    Average of & ctr's & tst's & ctr's* & tst's* \\
    \midrule
    Num. of neg. $\hat{Y}$ & 14.72 & 13.66 & 14.21 & 0.51 \\
    Prp. of neg. $\hat{Y}$ & 0.92 & 0.86 & 0.89 & 0.03 \\
    Avg. Annual salary & 78240.29 & 89308.43 & 87578.81 & 98705.10 \\
    Std. Annual salary & 1331.29 & 1672.54 & 1483.66 & 1484.45 \\
    Avg. Account balance & 22503.79 & 26478.22 & 24939.78 & 29011.26 \\
    Std. Account balance & 422.47 & 554.67 & 474.41 & 547.02 \\
    \bottomrule
  \end{tabular}
\end{table}

Let us consider the two CST versions for $k=15$.
Tables \ref{table:CST_without_k15} and \ref{table:CST_with_k15} summarize the control and test groups, respectively, of CST w/o and CST w/ split by statistical significance.
These tables are not mutually exclusive as CST w/o and CST w/ detect the same 272 statistically significant cases and the 288 CST w/o cases are included in the 420 CST w/ cases in Table~\ref{table:k-results}. 
The group pairs within each table are mutually exclusive as, different from previous tables, we split between statistically significant and non-significant cases. 
Based on the average characteristics in both tables, these groups are almost identical and enjoy, on average, a large $\Delta p$'s. 
Together, it explains why the inclusion (exclusion) of the the complainant and its counterfactual from, respectively, the control and test group in CST w/ (CST w/o) has little impact on these cases.
Comparing now the significant versus non-significant control and test groups in each table, we observe that the latter have a smaller difference between the average number of negative $\hat{Y}$ per group, which explains why these cases are detected by CST w/o and w/ but do not amount to statistically significant cases.

The non-statistically significant cases in Tables \ref{table:CST_without_k15} and \ref{table:CST_with_k15} vary more than the statistically significant cases in terms of their composition, with CST w/ having a considerably larger average number of negative decisions.
This difference illustrates the role of including/excluding the search centers and their representativeness relative to their corresponding neighborhoods.
Here, recall that CST w/ almost doubles CST w/o for all cases. 
CST w/ detects cases with more heterogeneous neighborhoods in terms of applicant suitability, which would explain the lower average proportion of negative outcomes.
CST w/, instead, detects more homogeneous neighborhoods marked by a lack of suitability due to the large average proportion of negative outcomes. What occurs for the non-statistically significant CST w/ groups is that the complainant and its counterfactual disagree, meaning the former is rejected and the latter is accepted, which would be enough to deviate $\Delta p$ from $\tau$ while both control and test neighborhoods appear to be, on average, clearly non-suitable. 

\begin{figure}[t]
    \centering
    \begin{subfigure}{.45\linewidth}
    \includegraphics[scale=0.45]{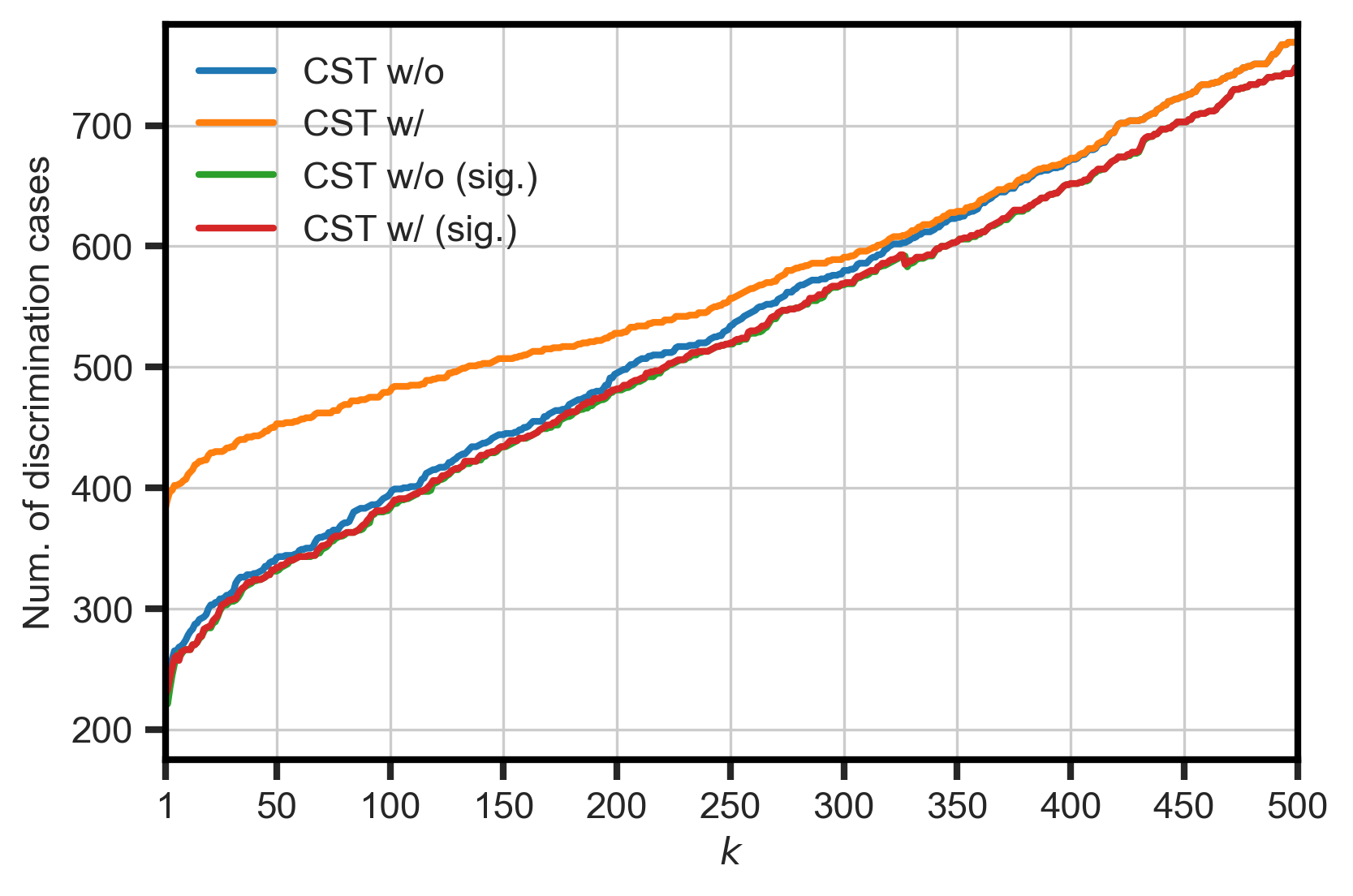}
    \end{subfigure}
    \hfill
    \begin{subfigure}{.45\linewidth}
    \includegraphics[scale=0.45]{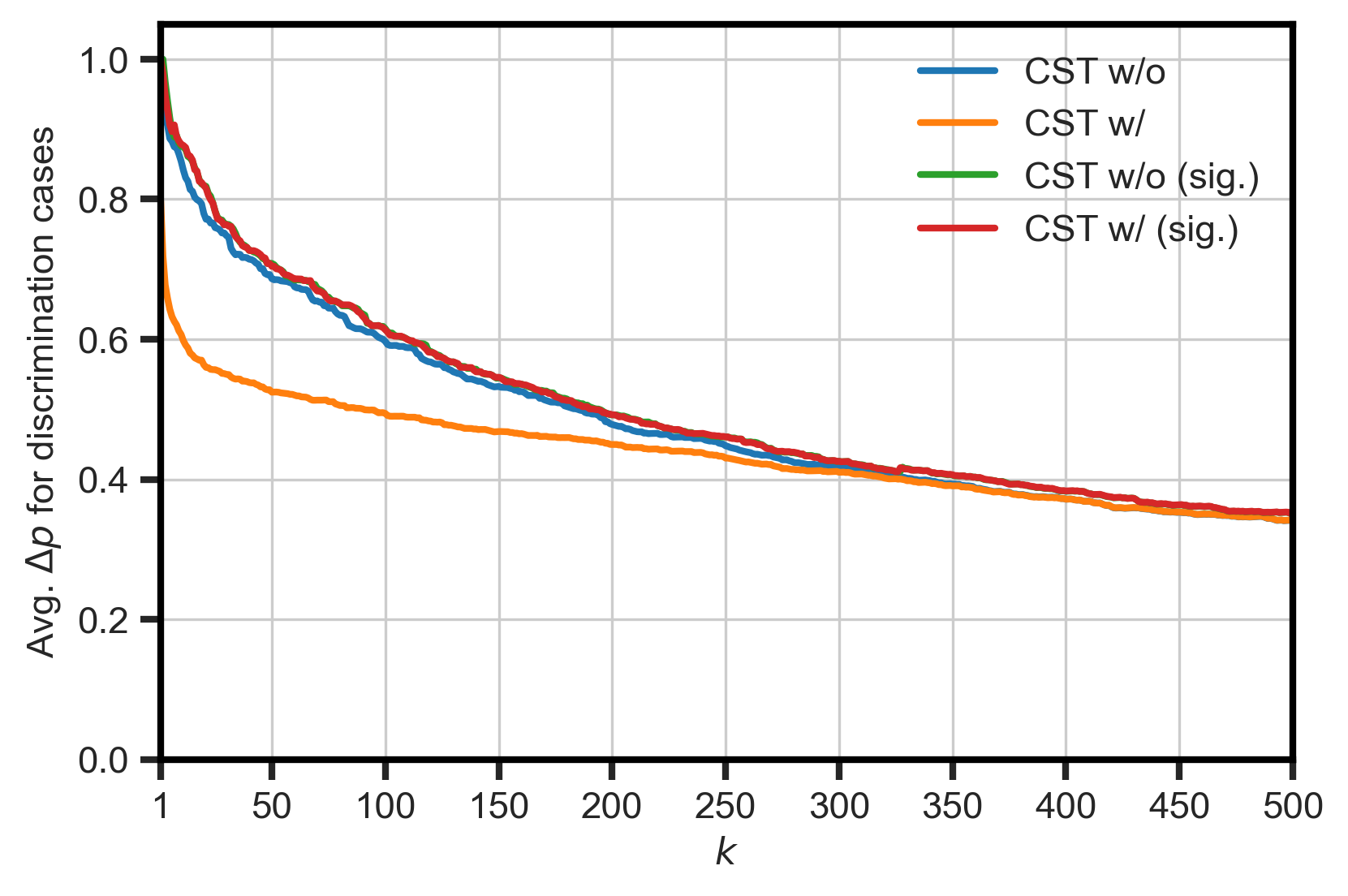}
    \end{subfigure}
\caption{Number of cases by CST w/o and w/ and their average $\Delta p$, respectively. We plot all and statistically significant (sig.) cases for both versions of CST.}
\label{fig:TwoCSTs_k_param}
\end{figure}

What about for larger neighborhood sizes? 
In Table~\ref{table:k-results}, CST w/o and CST w/ converge in number of cases as $k$ increases.
We argue that this occurs as larger neighborhoods diminish the impact of the \textit{(k+ 1)-th} pair, meaning the CST w/ is, in fact, converging toward CST w/o.
In Figure~\ref{fig:TwoCSTs_k_param} we plot the effect of increasing $k$ up to 500 on the number of discrimination cases and their average $\Delta p$. 
We do so for both CST versions, differentiating between all and statistically significant cases.
These plots align with Table~\ref{table:k-results}.
The plots support the previous analysis for Tables \ref{table:CST_without_k15} and \ref{table:CST_with_k15} regarding the impact of the additional pair in CST w/.
Consider the plots for the number of discrimination cases (left). As $k$ increases, the number of cases increases for all four. Notably, CST w/o's all and statistically significant cases and CST w/'s statistically significant cases align throughout $k$, while CST w/'s all cases converges to these three.
The same pattern occurs when we look at the plots for average $\Delta p$ (right), though the average $\Delta p$ decreases as $k$ increases.

Given Table~\ref{table:k-results} and Figure~\ref{fig:TwoCSTs_k_param}, it is worth asking whether having two versions of CST is useful?
Each version targets a different method, CST w/o relative to ST and CST w/ relative to CF; however, we might prefer CST w/o if expected to provide statistically significant results. 
What matters is that the difference between ST and both CST versions holds, highlighting the impact of using a \textit{mutatis mutandis} over a \textit{ceteris paribus} manipulation when testing for discrimination.
Also important is that we are able to provide certainty measures around CF under both CST versions as long as $k$ is large enough for $\Delta p$ under CST w/ to converge to $\Delta p$ under CST w/o.

Finally, the discrepancy between the CST versions raises questions on what it means for a $\Delta p$ to be representative.
Consider that we have not constrained $\mathcal{D}^{CF}$ w.r.t. $\mathcal{D}$, meaning it is possible for the counterfactual instances to deviate considerably from the factual instances. 
In that case, we would have ``unrepresentative'' counterfactual instances that motivate statistically insignificant comparisons, but that are still ``unavoidable'' in the sense that these same instances embody scenarios that should have happened for the complainants according to a SCM $\mathcal{M}$.
We believe that judging the representativeness of the counterfactual world by using the factual world as the underlying population might not properly capture the substantive equality goals behind non-discrimination law \parencite{Wachter2020BiasPreserving}. 
We come back to this last discussion in Section~\ref{sec:Discussion}.

\subsection{Law School Admissions}
\label{sec:Experiments.Real}

\begin{figure}[t]
\begin{minipage}{.35\linewidth}
\begin{figure}[H]
\centering
    \begin{tikzpicture}
        \node (A1)  at (-1.75, -0.55) [circle, draw]{R};
        \node (A2)  at (-1.75, 0.55) [circle, draw]{G};
        \node (X1) at (0, 0.85) [circle, draw]{UGPA};
        \node (X2) at (0,-0.85) [circle,draw]{LSAT};
        \node (Y)  at (1.25, 0) [circle, draw]{$\widehat{Y}$};
        \draw[->] (A1) to (X1) {};
        \draw[->] (A1) to (X2) {};
        \draw[->] (A2) to (X1) {};
        \draw[->] (A2) to (X2) {};
        \draw[->] (X1) to (Y) {};
        \draw[->] (X2) to (Y) {};
    \end{tikzpicture}
\end{figure}
\end{minipage}
\begin{minipage}{.55\linewidth}
\begin{align*}
\mathcal{M} \, & 
\begin{cases}
    R & \leftarrow U_{R}\\
    G & \leftarrow U_G \\
    UGPA & \leftarrow b_U + \beta_1 \cdot R + \lambda_1 \cdot G + U_1, \ \\
    LSAT & \leftarrow  \exp\{b_L + \beta_2 \cdot R + \lambda_2 \cdot G + U_2\}, \\
\end{cases}
\end{align*}
\begin{align*}
    \widehat{Y} & = b(UGPA, LSAT) 
\end{align*}
\end{minipage}
\caption{The auxiliary causal knowledge for Section~\ref{sec:Experiments.Real}. Let $R$ denote race, $G$ gender, \textit{LSAT} law school admissions test scores, \textit{UGPA} undergraduate grade-point average, and $\hat{Y}$ the admissions decision by $b()$.}
\label{fig:LawSchool}
\end{figure}

Let us now consider the law school admissions scenario popularized by \textcite[Figure 2]{Kusner2017CF}.
We use US data from the Law School Admission Council survey \parencite{Wightman1998_LawDataSource}, and recreate an admissions scenario for a top US law school. 
We consider as protected attributes an applicant's gender (\textit{G}, male/female), and race (\textit{R}, white/non-white). 
We add the ADM $b(UGPA, LSAT) = \hat{Y}$, which considers the applicant's undergraduate grade-point average (\textit{UGPA}) and law school admissions test scores (LSAT). 
If an applicant is successful, $\widehat{Y}=1$; otherwise $\widehat{Y}=0$. 
We summarize the scenario in Figure~\ref{fig:LawSchool}.
We define the ADM $b()$ using the median entry requirements for the top US law school to derive the cutoff $\psi$.\footnote{That being Yale University Law School; see \url{https://www.ilrg.com/rankings/law/index/1/asc/Accept}}
Formally, we define $b()$ as $\mathbbm{1}\{(0.6 \cdot UGPA + 0.4 \cdot LSAT) > \psi\}$.
The cutoff is the weighted sum of 60\%  in \textit{UGPA} (3.93 over 4.00), and 40\%  \textit{LSAT} (46.1 over 48), giving a total of 20.8; the maximum possible score given $b()$ is 22. 
The SCM $\mathcal{M}$ and DAG $\mathcal{G}$ follow \textcite{Kusner2017CF},
with $b_U$ and $b_L$ denoting the intercepts; $\beta_1$, $\beta_2$, $\lambda_1$, $\lambda_2$ the weights; and $U_1 \sim \mathcal{N}$ and $U_2 \sim \text{Poi}$ the probability distributions.

We study the behavior of $b()$ toward $G$ and $R$.
The dataset $\mathcal{D}$ contains $n=21790$ applicants, 43.8\% being female, 16.1\% being non-white, and 8.4\% being non-white-female.
Despite $b()$ being externally imposed by us for the purpose of illustrating the CST framework, under $b()$ only 0.8\% of the female applicants are successful compared to 1.5\% of the male applicants; similarly, only 0.2\% of the non-white applicants are successful compared to 2.2\% of the white applicants.
It is also the case when considering the intersectional group of non-white-females, with only 0.06\% of these applicants being admitted to law school based on $b()$ compared to the 2.25\% of white-female, non-white-male, and white-male successful applicants.
Notice that $b()$ is highly selective, with an acceptance rate of just 2.31\%, or 505 out of 21790 applicants considered.
Using DP as a fairness metric, $b()$ would still be considered unfair toward female, non-white, and non-white-female applicants.

\subsubsection{Single Discrimination}
\label{sec:Experiments.Real.Single}

\begin{table}[t]
\caption{Number (and \% w.r.t.~non-whites) of individual discrimination cases based on $R$ using Figure~\ref{fig:LawSchool}. Marked by * are the statistically significant cases.}
  \label{table:k-results_RACE}
  \centering
  \begin{tabular}{cccccc}
    \toprule
    Method & $k=15$ & $k=30$ & $k=50$ & $k=100$ & $k=250$\\
    \midrule
    CST w/o & 256 (7.3\%) & 309 (8.8\%) & 337 (9.6\%) & 400 (11.4\%)  & 503 (14.4\%) \\
     & 244* (6.9\%) & 301* (8.6\%) & 323* (9.2\%) & 391* (11.2\%)  & 494* (14.1\%) \\
     \midrule
    ST & 33 (0.9 \%) & 51 (1.5\%) & 61 (1.7\%) & 64 (1.8\%) & 78 (2.2\%) \\
    & 28* (0.8\%) & 28* (0.8\%) & 45* (1.3\%) & 47* (1.3\%) & 61 (1.7\%) \\
    \midrule
    CST w/ & 286 (8.2\%) & 309 (8.8\%) & 337 (9.6\%) & 400 (11.4\%)  & 503 (14.4\%)\\
    & 244* (6.9\%) & 301* (8.6\%) & 323* (9.2\%) & 391* (11.2\%)  & 494* (14.1\%) \\
    \midrule
    CF & 231 (6.6\%) & 231 (6.6\%) & 231 (6.6\%) &  231 (6.6\%) &  231 (6.6\%) \\
    & 190* (5.4\%) & 231* (6.6\%) & 231* (6.6\%) & 231* (6.6\%) & 231* (6.6\%) \\
    \bottomrule
  \end{tabular}
\end{table}
\begin{table}[t]
\caption{Number (and \% w.r.t.~females) of individual discrimination cases based on $G$ Figure~\ref{fig:LawSchool}. Marked by * are the statistically significant cases.}
  \label{table:k-results_GENDER}
  \centering
  \begin{tabular}{cccccc}
    \toprule
    Method & $k=15$ & $k=30$ & $k=50$ & $k=100$ & $k=250$\\
    \midrule
    CST w/o & 78 (0.8\%) & 120 (1.3\%) & 253 (2.7\%) & 296 (3.1\%)  & 493 (5.2\%) \\
     & 43* (0.5\%) & 88* (0.9\%) & 160* (1.7\%) & 221* (2.3\%)  & 341* (3.6\%) \\
     \midrule
    ST & 77 (0.8\%) & 101 (1.1\%) & 229 (2.4\%) & 258 (2.7\%) & 484 (5.1\%) \\
    & 57* (0.6\%) & 69* (0.7\%) & 111* (1.2\%) & 124* (1.3\%) & 366 (3.8\%) \\
    \midrule
    CST w/ & 99 (1.0\%) & 129 (1.4\%) & 267 (2.8\%) & 296 (3.1\%)  & 493 (5.2\%)\\
    & 54* (0.6\%) & 92* (0.9\%) & 160* (1.7\%) & 221* (2.3\%)  & 341* (3.6\%) \\
    \midrule
    CF & 56 (0.6\%) & 56 (0.6\%) & 56 (0.6\%) &  56 (0.6\%) &  56 (0.6\%) \\
    & 20* (0.2\%) & 15* (0.2\%) & 30* (0.3\%) & 21* (0.2\%) & 32* (0.3\%) \\
    \bottomrule
  \end{tabular}
\end{table}

Does $b()$ discriminate against non-white applicants? 
To answer this question using CST and CF, we generate the corresponding $\mathcal{D}^{CF}_R$ using Figure~\ref{fig:LawSchool} based in the intervention $do(R:=0)$, or \textit{what would have happened had all law school applicants been white?}
Similarly, does $b()$ discriminate against female applicants?
To answer this question using CST and CF, we generate the corresponding $\mathcal{D}^{CF}_G$ using Figure~\ref{fig:LawSchool} based in the intervention $do(G:=0)$, or \textit{what would have happened had all law school applicants been male?}
Both questions share the same $\mathcal{D}$.
Similar to Section~\ref{sec:Experiments.IllustrativeExample}, we use Definition~\ref{def:IndDisc} for detecting individual discrimination cases and Definition~\ref{def:CIs} for determining whether these cases are statistically significant.
Tables~\ref{table:k-results_RACE} and \ref{table:k-results_GENDER} show the results for all methods.
 
Tables~\ref{table:k-results_RACE} and \ref{table:k-results_GENDER} report similar patterns to those in Table~\ref{table:k-results}.
CST w/o detects a higher number of cases than ST; CST w/ detects a high number of cases than CF; and CST w/ detects a higher number than CST w/o. 
For all these patterns, though, the differences between the corresponding methods is smaller than those observed in Table~\ref{table:k-results}.
This is due to the composition of $\mathcal{D}$ and the nature of $b()$. 
Here, we work with a larger $\mathcal{D}$ (21790 versus 5000 applicants) and a more selective $b()$ (2.31\% versus 53.3\% acceptance rate).
CST w/o and CST w/ in Table~\ref{table:k-results_GENDER} converge already at larger neighborhood sizes, which does not occur in Table~\ref{table:k-results} (there is, though, a clear pattern that it occurs eventually as shown in Figure~\ref{fig:TwoCSTs_k_param}).
We also observe similar patterns once we account for statistical significance with CST w/o and CST w/ reaching the same number of cases in both tables for $k=250$.

\begin{figure}[t]
    \begin{subfigure}{.45\linewidth}
    \includegraphics[scale=0.45]{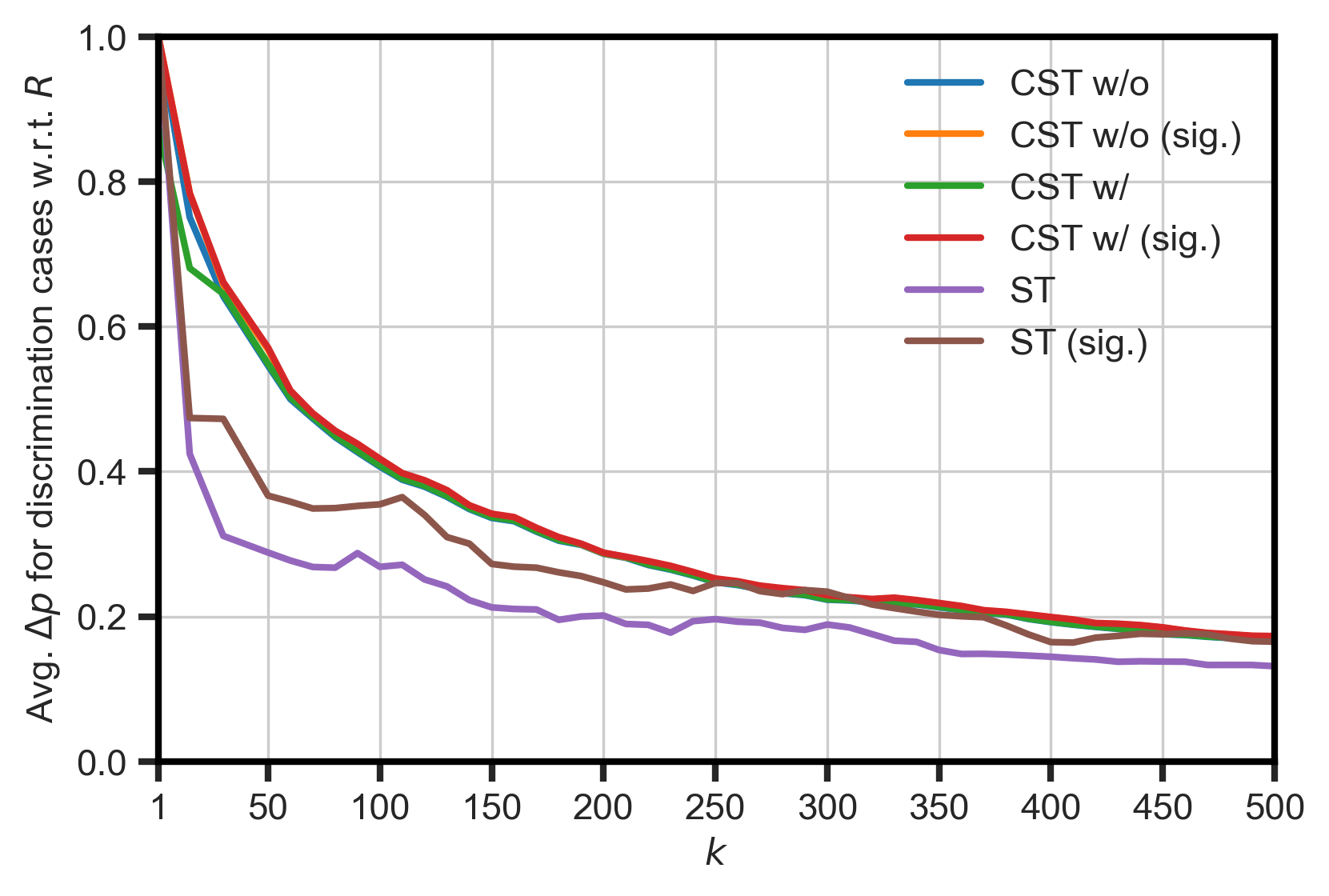}
    \caption{}
    \end{subfigure}
\hfill
    \begin{subfigure}{.45\linewidth}
    \includegraphics[scale=0.45]{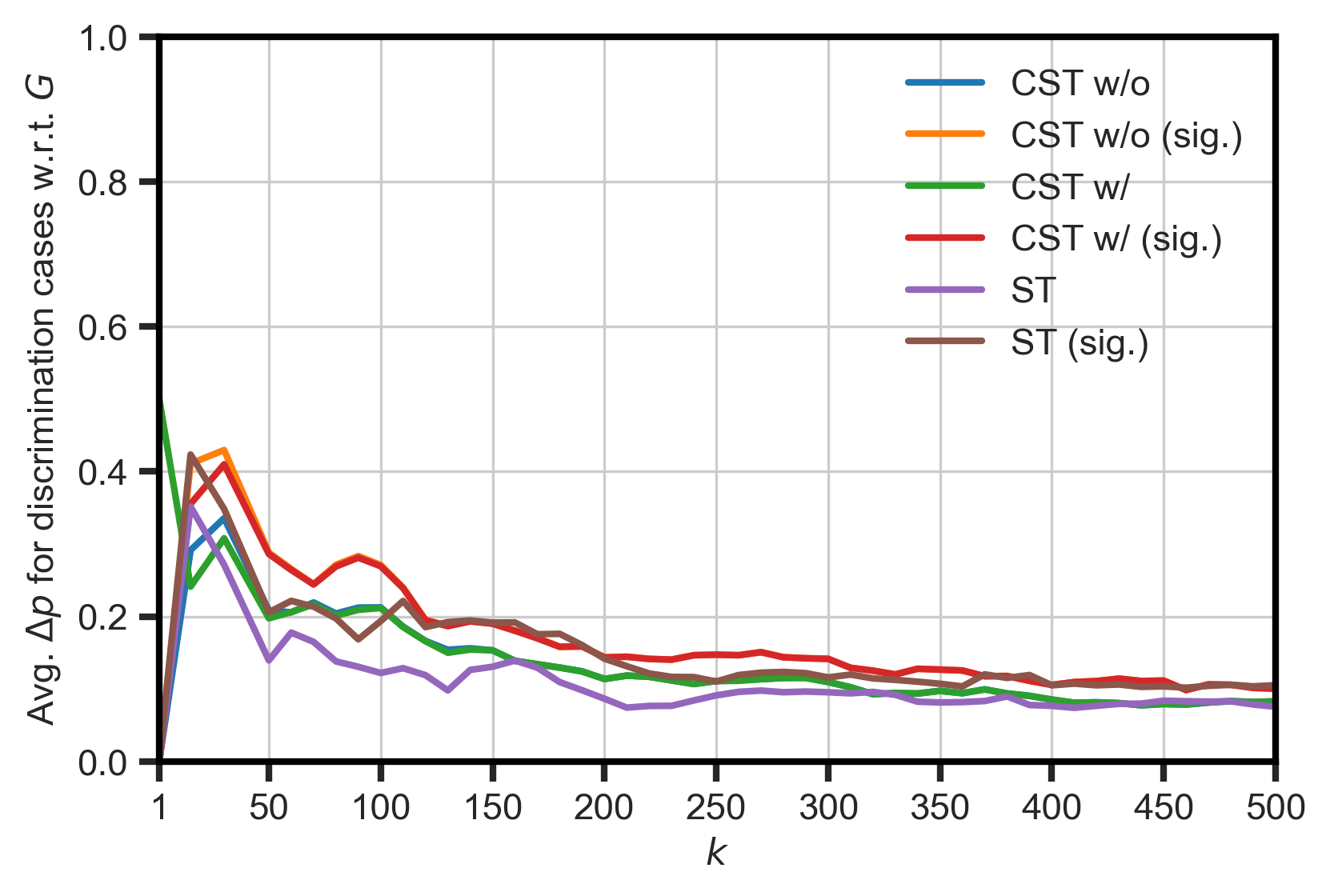}
    \caption{}
    \end{subfigure}
\medskip
   \begin{subfigure}{.45\linewidth}
    \includegraphics[scale=0.45]{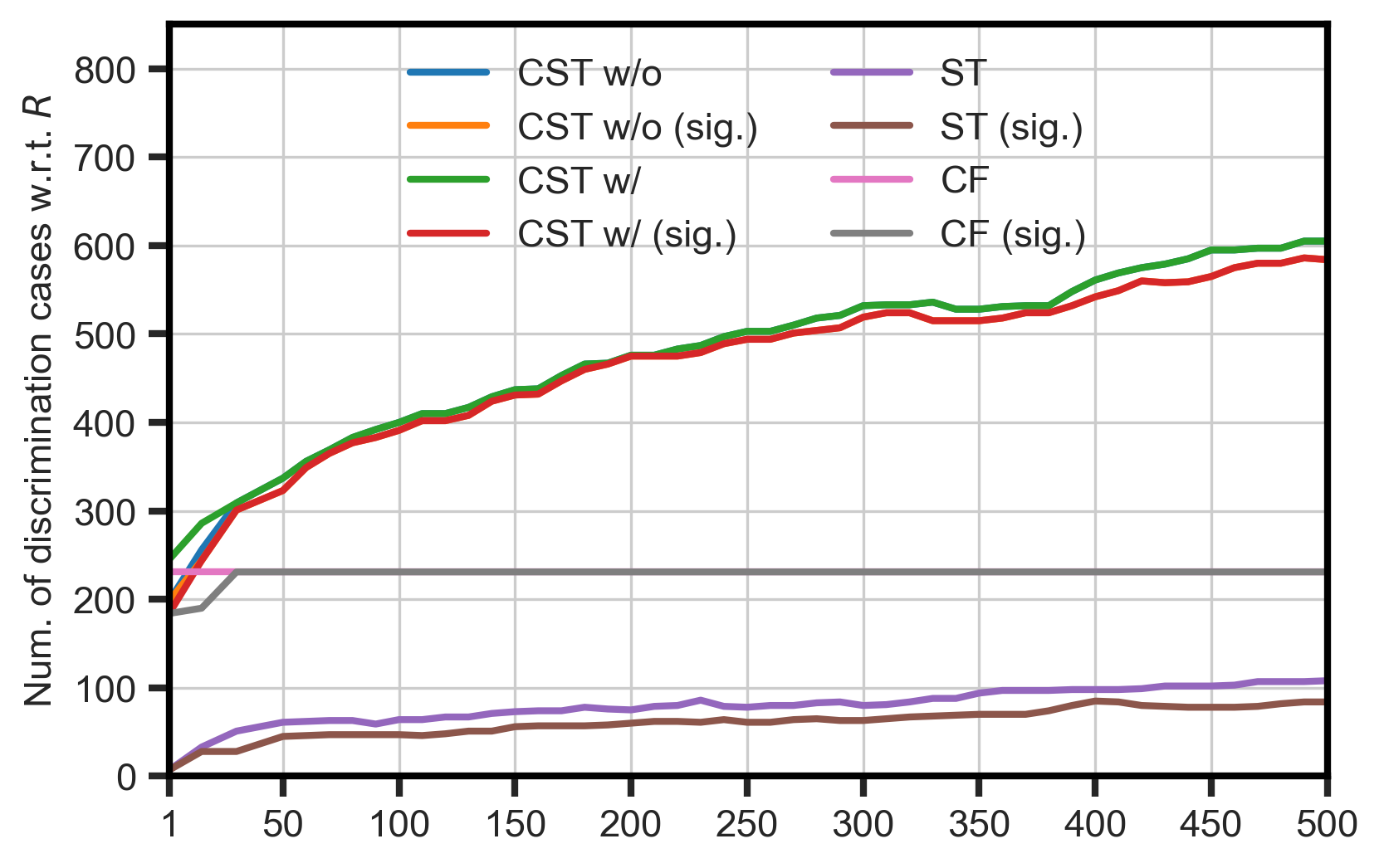}
    \caption{}
    \end{subfigure}
\hfill
    \begin{subfigure}{.45\linewidth}
    \includegraphics[scale=0.45]{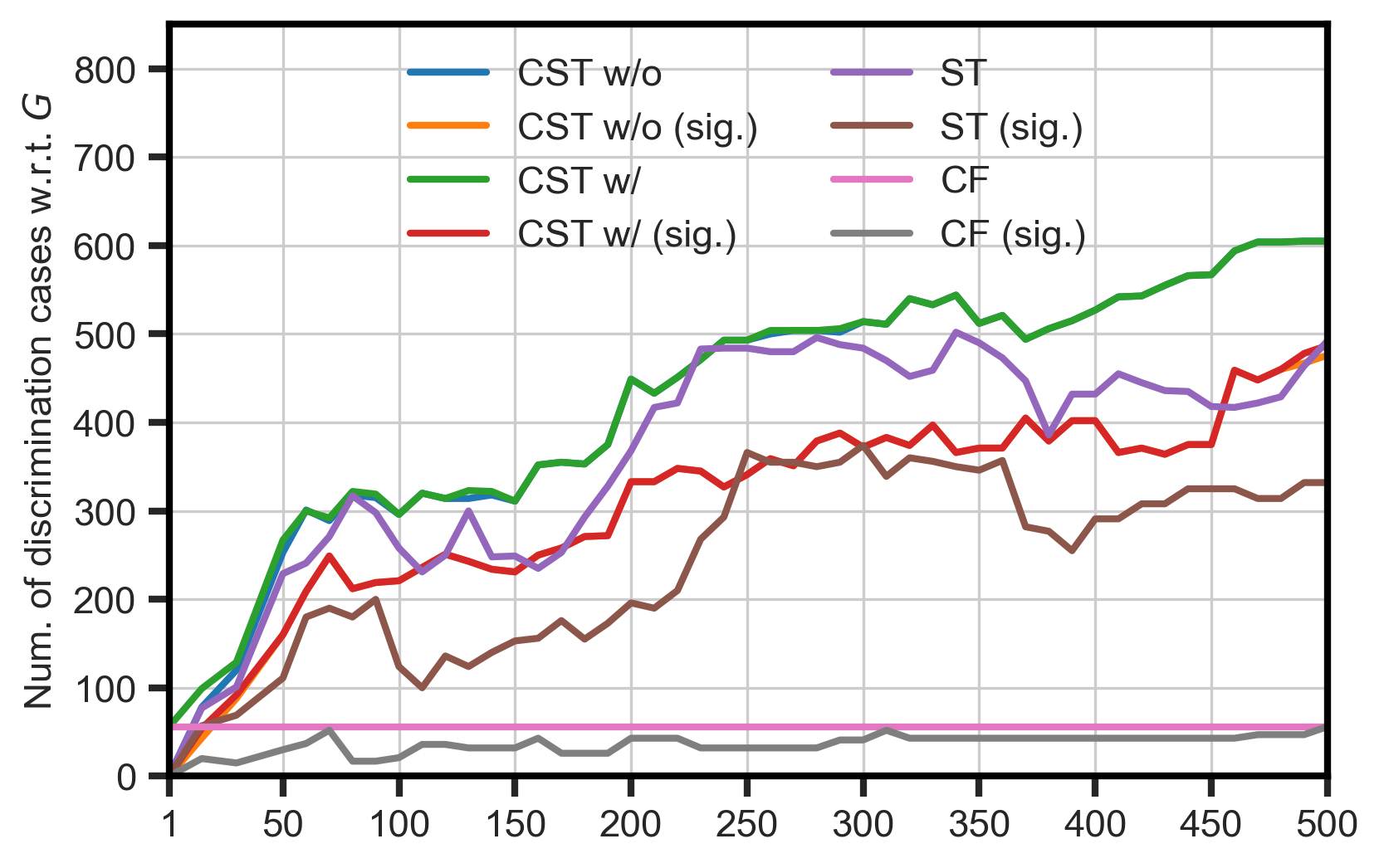}
    \caption{}
    \end{subfigure}
\caption{Average $\Delta p$ and number of cases for race ($R$) and gender ($G$), respectively. We plot all and statistically significant (sig.) cases for each method.}
\label{fig:LawSchoolSingleDiscrimiantion_allmethods_k_param}
\end{figure}

Figure~\ref{fig:LawSchoolSingleDiscrimiantion_allmethods_k_param} supports Tables~\ref{table:k-results_RACE} and \ref{table:k-results_GENDER}, showing the average $\Delta p$ and the number of cases up to $k=500$ for all methods.\footnote{Due the size of $\mathcal{D}$, we run the methods for $k$ equals 1, 15, 30 and between 50-500 in increments of 10.} 
For the average $\Delta p$, as shown in sub-figures (a) and (b), it decreases as $k$ increases, with all methods seemingly converging to a single value.
%
By observing that, as $k \rightarrow \infty$, the neighborhoods of the complainant and its counterfactual include all protected and unprotected instances, respectively, in $\mathcal{D}$, this value turns out to be the difference in DP: $P(\hat{Y}|A=1) - P(\hat{Y}|A=0)$ (cfr., footnote~\ref{foot:dp}) .
The average $\Delta p$ for significant cases is higher to, see ST in (a), or almost the same as, see CST w/ in (a), for all the cases.  
As $k$ increases, the control and tests groups constructed by ST, CST w/o, and CST w/ start considering new instances further away from the search centers and whatever detected initial deviation from $\tau$ dissipates slowly.
Similarly, as shown in sub-figures (c) and (d), the number of cases increases as $k$ increases except for CF which is independent of $k$ and its significant cases that are bounded by CF itself. 
We observe the CST versions converging, especially when accounting for statistical significance, while ST remains below both CST versions in (c) and mimics CST w/o in (d).
The number of significant cases is lower to ST in (c), or almost the same as both CST versions in (c) for all cases. 

The number of cases varies across the methods between Tables~\ref{table:k-results_RACE} and \ref{table:k-results_GENDER}. 
This is due to race and gender having different non-protected and protected search spaces.
The results are comparable, but represent separate tests for single discrimination.
Recall that non-whites represent 16.1\% while females represent 43.8\% of $\mathcal{D}$. 
It means that CST has access to a smaller search space when building the control groups for non-white complainants relative to female complainants.
Notably, in Table~\ref{table:k-results_RACE} statistically significant CF cases reach the 231 CF total cases within the $k$ values considered. 
This does not occur in Table~\ref{table:k-results_GENDER} in which the statistically significant CF cases slowly increase, though in a non-monotonically way, toward the 56 CF total cases.
Such oscillation, we believe, is due to the statistical estimator not yet reaching its asymptotic behavior.
These are, though, minor fluctuations as the number of statistically significant cases is around 0.2-0.3\%.\footnote{We use the CI \eqref{eq:CIs} from CST w/ though conditioned on CF discrimination occurring. We only detect 56 cases of CF discrimination (Table~\ref{table:k-results_GENDER}). As $k$ increases, we always look at these complainants.}
In Figure~\ref{fig:LawSchoolSingleDiscrimiantion_allmethods_k_param} we observe this non-monotonic increase for number of cases and decrease for the average $\Delta p$ of cases detected more clearly in the sub-figures (a) and (c) for race than for the sub-figures (b) and (d) for gender. 
The plots for gender are considerably less smooth than those for race. 
The composition of $\mathcal{D}$ clearly plays a role here as females represent 43.8\% and non-whites 16.1\% of the dataset, with the k-NN based methods varying more between iterations as they explore a much denser search space.

\subsubsection{Multidimensional Discrimination}
\label{sec:Experiments.Real.Multi}

We present the results for the forms of multidimensional discrimination, multiple and intersectional.
Given the focus on gender and race, the non-protected group amounts to the non-protected groups based on race and gender: i.e., white and male applicants.
As these two groups of applicants are not mutually exclusive, white-females and non-white-males are also part of the non-protected group.
This point is clearer when we consider the intersection of race and gender and focus on the protected group that is non-white-female applicants: the complementary of such group, meaning the non-protected group, includes white-female, non-white-male, and white-male applicants.

\begin{table}[t]
  \caption{Number (and \% w.r.t.~non-white-females) of multiple individual discrimination cases in Section~\ref{sec:Experiments.Real} for $R$ and $G$. Marked by * are the statistically significant cases.}
  \label{table:k-results_Multiple}
  \centering
  \begin{tabular}{cccccc}
    \toprule
    Method & $k=15$ & $k=30$ & $k=50$ & $k=100$ & $k=250$\\
    \midrule
    CST w/o & 8 (0.44\%) & 10 (0.55\%) & 20 (1.09\%) & 20 (1.09\%)  & 40 (2.18\%) \\
     & 4* (0.22\%) & 6* (0.33\%) & 11* (0.60\%) & 17* (0.93\%)  & 24* (1.31\%) \\
     \midrule
    ST & 5 (0.27\%) & 5 (0.27\%) & 12 (0.65\%) & 19 (1.04\%) & 24 (5.1\%) \\
    & 0* (0.0\%) & 0* (0.0\%) & 5* (0.27\%) & 5* (0.27\%)  & 15* (0.82\%) \\
    \midrule
    CST w/ & 9 (0.49\%) & 10 (0.55\%) & 21 (1.15\%) & 20 (1.09\%)  & 40 (2.18\%)\\
    & 4* (0.22\%) & 9* (0.49\%) & 11* (0.60\%) & 17* (0.93\%)  & 24* (1.31\%) \\
    \midrule
    CF & 5 (0.27\%) & 5 (0.27\%) & 5 (0.27\%) & 5 (0.27\%) & 5 (0.27\%) \\
    & 0* (0.0\%) & 3* (0.16\%) & 1* (0.05\%) & 1* (0.05\%) & 2* (0.11\%) \\
    \bottomrule
  \end{tabular}
\end{table}
\begin{table}[t]
  \caption{Number (and \% w.r.t.~non-white-females) of intersectional individual discrimination cases in Section~\ref{sec:Experiments.Real} for $R \times G$. Marked by * are the statistically significant cases.}
  \label{table:k-results_Intersectional}
  \centering
  \begin{tabular}{cccccc}
    \toprule
    Method & $k=15$ & $k=30$ & $k=50$ & $k=100$ & $k=250$\\
    \midrule
    CST w/o & 130 (7.1\%) & 138 (7.5\%) & 148 (8.1\%) & 160 (8.7\%)  & 199 (10.9\%) \\
    & 130* (7.1\%) & 138* (7.5\%) & 148* (8.1\%) & 160* (8.7\%)  & 199* (10.9\%) \\
     \midrule
    ST & 14 (0.8\%) & 14 (0.8\%) & 17 (0.9\%) & 24 (1.3\%) & 29 (1.6\%) \\
    & 14* (0.8\%) & 14* (0.8\%) & 13* (0.7\%) & 23* (1.3\%)  & 26* (1.4\%) \\
    \midrule
    CST w/ & 130 (7.1\%) & 138 (7.5\%) & 148 (8.1\%) & 160 (8.7\%)  & 199 (10.9\%) \\
    & 130* (7.1\%) & 138* (7.5\%) & 148* (8.1\%) & 160* (8.7\%)  & 199* (10.9\%) \\
    \midrule
    CF & 113 (6.2\%) & 113 (6.2\%) & 113 (6.2\%) & 113 (6.2\%) & 113 (6.2\%) \\
    & 113* (6.2\%) & 113* (6.2\%) & 113* (6.2\%) & 113* (6.2\%) & 113* (6.2\%) \\
    \bottomrule
  \end{tabular}
\end{table}

\begin{figure}[t]
    \begin{subfigure}{.45\linewidth}
    \includegraphics[scale=0.45]{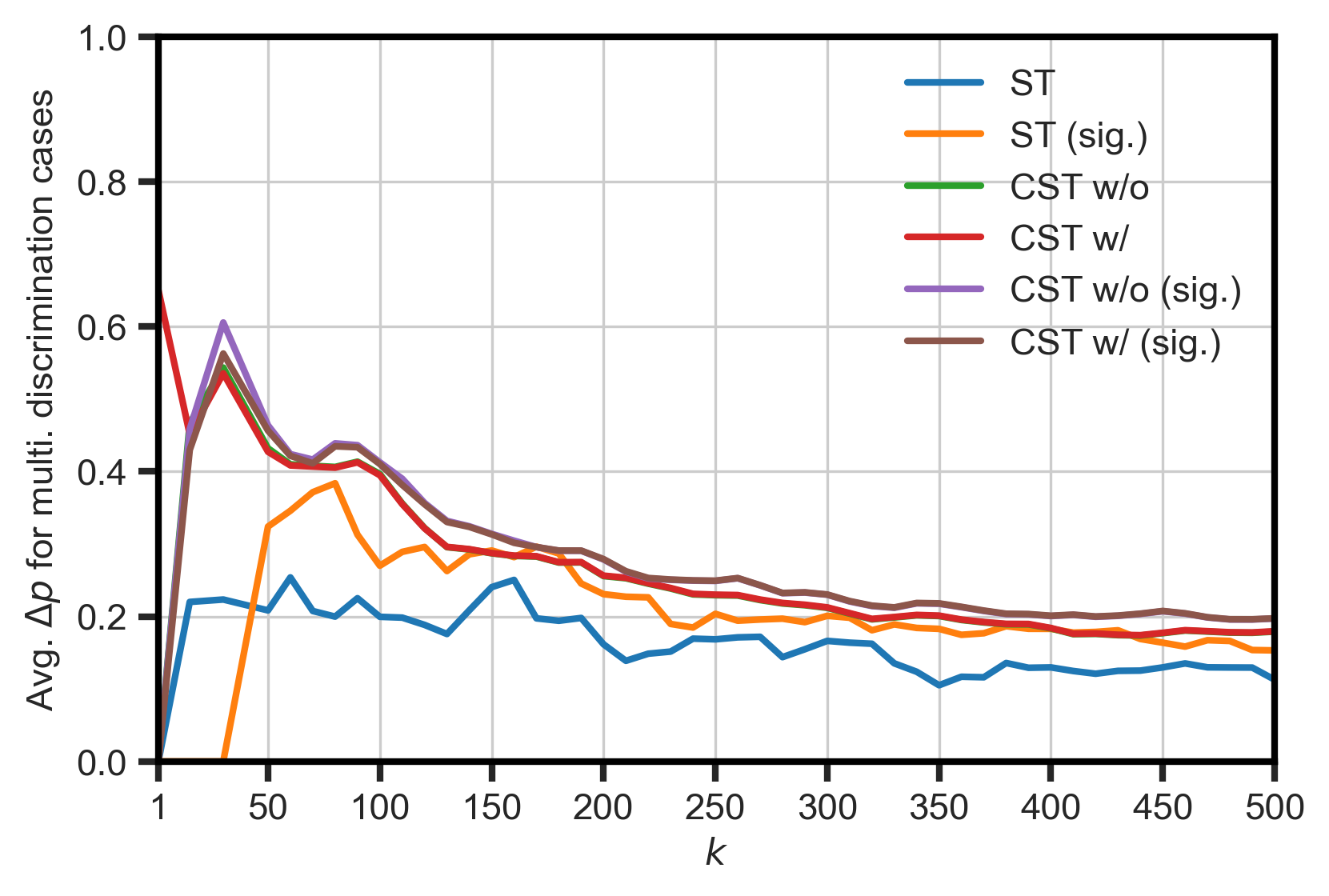}
    \caption{}
    \end{subfigure}
\hfill
    \begin{subfigure}{.45\linewidth}
    \includegraphics[scale=0.45]{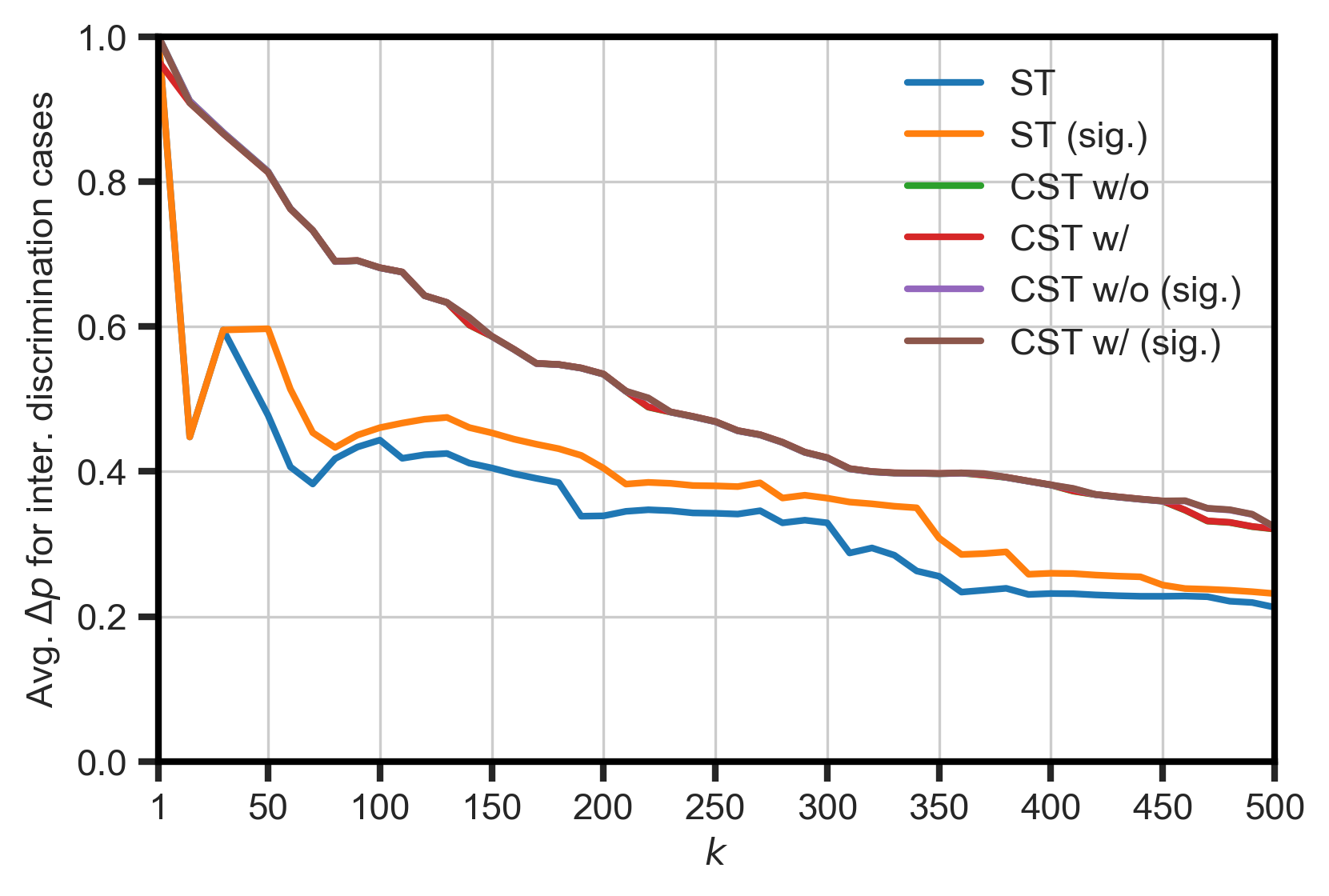}
    \caption{}
    \end{subfigure}
    \medskip
    \begin{subfigure}{.45\linewidth}
    \includegraphics[scale=0.45]{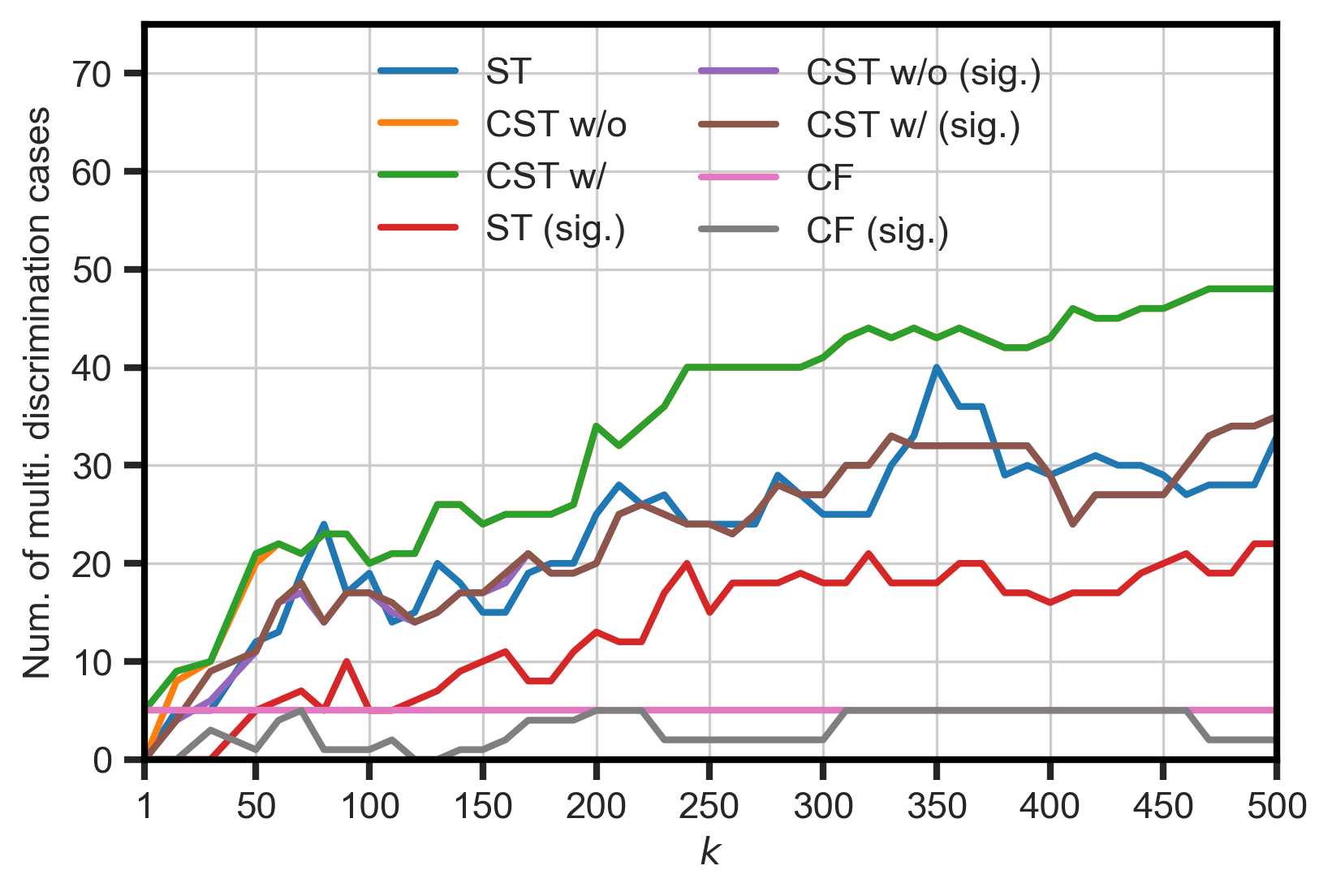}
    \caption{}
    \end{subfigure}
\hfill
    \begin{subfigure}{.45\linewidth}
    \includegraphics[scale=0.45]{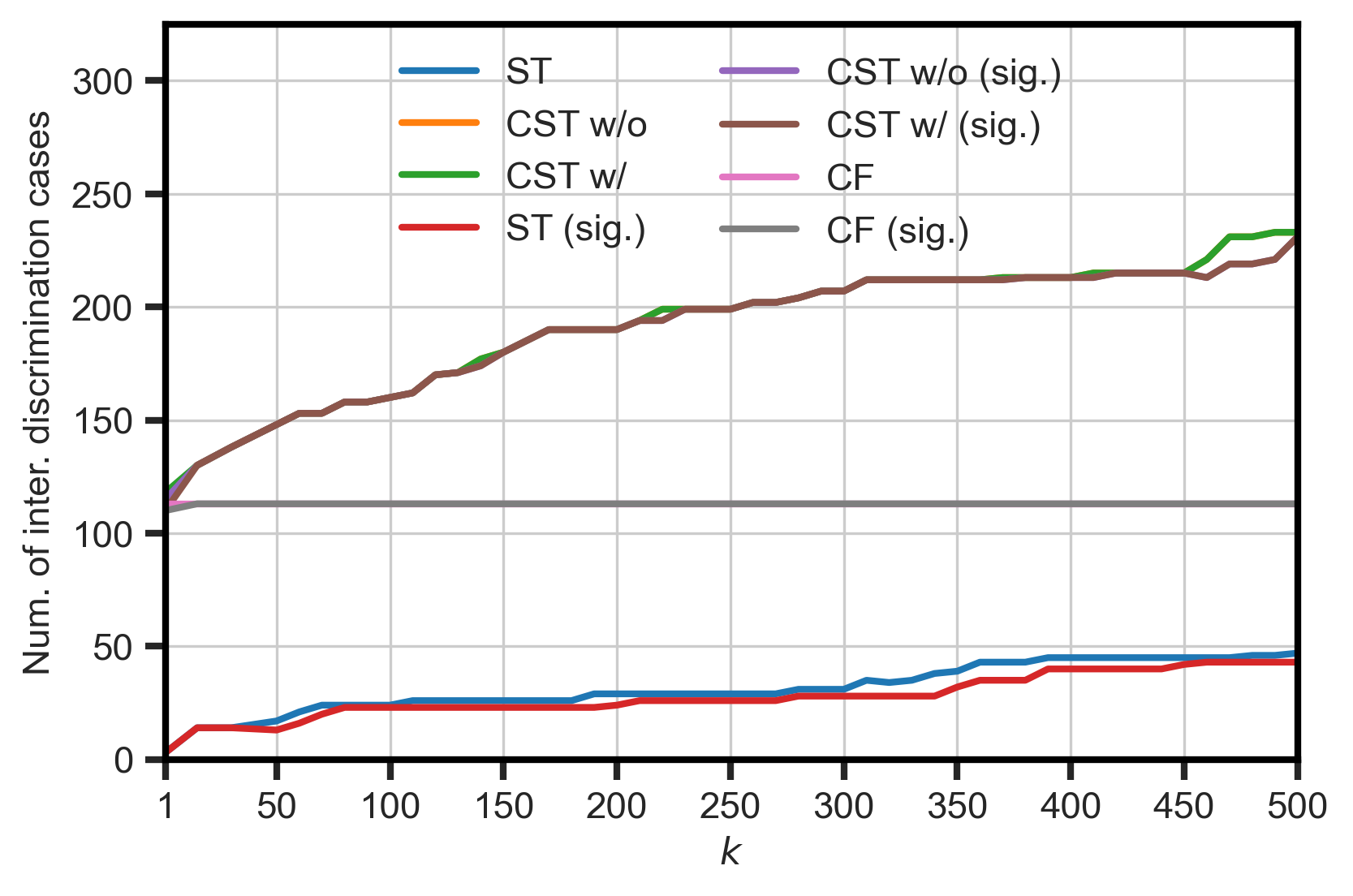}
    \caption{}
    \end{subfigure}
\caption{Avg. $\Delta p$ and number of cases for multiple (left) and intersectional (right) discrimination on $R$ and $G$. We plot all and statistically significant (sig.) cases for all methods.}
\label{fig:LawSchoolMultiDiscrimination_allmethods_k_param}
\end{figure}

Following Definition~\ref{def:MultipleDisc}, we count as multiple discrimination based on race and gender when $\Delta p > \tau$ occurs separately for each of these protected attributes. 
Only those cases that are statistically significant for each protected attribute under CI \eqref{eq:CIs}---given the Bonferroni corrected $\alpha/2$---amount to statistically significant cases. 
We still rely on the generated $\mathcal{D}_R^{CF}$ and $\mathcal{D}_G^{CF}$ for the construction of the test groups and the original $\mathcal{D}$ for the construction of the control groups.
Does $b()$ discriminate against the none-white-female applicants as non-white \textit{and} as female applicants?
We present the results in Table~\ref{table:k-results_Multiple}. 
Figure~\ref{fig:LawSchoolMultiDiscrimination_allmethods_k_param}, as shown in sub-figures (a) and (c), further illustrates the results up to $k=500$. 
For the average $\Delta p$, we average those from cases detected separately under race and gender.

Following Definition~\ref{def:IntersectionaleDisc}, instead, we count intersectional discrimination based on race and gender when $\Delta p > \tau$ occurs for the intersection of these protected attributes. 
In practice, it means constructing the new protected attribute $R \times G$; updating $\mathcal{D}$ into $\mathcal{D}'$, such that $R \times G \in \mathcal{D}'$; and generating $\mathcal{D}'^{CF}$ given Figure~\ref{fig:LawSchool} under $do(R \times G := 0)$.
It implies a single discrimination run but under the ``new'' single attribute $R \times G$, representing the intersection of $R$ and $G$. 
Cases are statistically significant under CI \eqref{eq:CIs} based on $R \times G$.
Formally, in Figure~\ref{fig:LawSchool} we merge the $R$ and $G$ nodes into the single $R \times G$ in $\mathcal{G}$ node and do the same for the corresponding equations in $\mathcal{M}$ by interacting the dummy variables for $R$ and $G$ and re-estimating the regression weights \parencite{Wooldridge2015IntroductoryEconometrics}.
Does $b()$ discriminate against the none-white-female applicants?
We present the results in Table~\ref{table:k-results_Intersectional}.
Figure~\ref{fig:LawSchoolMultiDiscrimination_allmethods_k_param}, as shown in sub-figures (b) and (d), further illustrates the results up to $k=500$.

In Tables~\ref{table:k-results_Multiple} and \ref{table:k-results_Intersectional}, the three methods show similar patterns between them.
CST w/o detects more cases relative to ST (including statistically significant cases); CST w/ detects more cases than CF (including statistically significant cases); and the two CST versions converge once statistical significance is considered.
The same line of reasoning used before still applies here for understanding how CST, ST, and CF relate to each other.
The difference is that the protected group and, in turn, the non-protected group are defined by more than one protected attribute.
What is interesting in Table~\ref{table:k-results_Multiple} is that CST w/o and CST w/ converge early on for all cases, not just for those cases that are statistically significant.
We suspect these are cases that are clearly discriminatory under both race and gender, making them likely to be detected by multiple and intersectional discrimination testing.
All cases in Table~\ref{table:k-results_Intersectional} are also statistically significant for all methods.
It is due to $R \times G=1$ representing the most un-favored protected group from the combination of $R$ and $G$, which results in larger $\Delta p$'s and a complete convergence of all and statistically significant cases for all methods relative to multiple discrimination. 
Compare, e.g.,~(a) versus (b) and (c) versus (d) in Figure~\ref{fig:LawSchoolMultiDiscrimination_allmethods_k_param}. 
We discuss the last point in the next section.  

\subsubsection{On Multiple and Intersectional Discrimination}
\label{sec:Experiments.Real.Multiple_vs_Intersectional}

The results from the previous section support claims by legal scholars on the risk of not recognizing intersectional discrimination under non-discrimination law.
These claims, to the best of our knowledge, date back to \textcite{Crenshaw1989_DemarginalizingTheIntersection} and have become prominent again with the ongoing discussion around algorithmic discrimination \parencite{Xenidis2020_TunningEULaw}.
We suspect that, since multiple and intersectional discrimination share the protected group and the non-protected groups, there is a tendency to dismiss the latter as a special case of the former.
An in-depth legal discussion on the tension between multiple and intersectional discrimination is beyond this paper, but Tables~\ref{table:k-results_Multiple} and \ref{table:k-results_Intersectional} support the calls by these researchers to treat intersectional discrimination separate from multiple discrimination.

We acknowledge that, from a modeling perspective, a difference between Tables~\ref{table:k-results_Multiple} and \ref{table:k-results_Intersectional} is expected since we implement different procedures.
Under multiple discrimination we look at the intersection of two separate single discrimination testing runs, while under intersectional discrimination we look at a single discrimination run representing the intersection. 
The claims by legal scholars like \textcite{Crenshaw1989_DemarginalizingTheIntersection, Xenidis2020_TunningEULaw}, however, were not as apparent to us, meaning, in principle, we had no reason to expect a higher number of cases for intersectional discrimination over multiple discrimination.
In fact, from a probability theory perspective (read, the conjunction rule), if we had to choose a difference between the Tables~\ref{table:k-results_Multiple} and \ref{table:k-results_Intersectional}, we would have guessed the opposite, with multiple discrimination acting as an upper limit to intersectional discrimination.
Given these results, we now would add that such expectation holds from a modeling perspective if we agree that the intersection of $G$ and $R$ is not its own category. Let us discuss further.

The two modeling procedures allow to represent the case in which $R \times G$ is its own category (intersectional) and in which is just the conjunction of $R$ and $G$ (multiple).
This distinction, with legal origins as already argued \parencite{Crenshaw1989_DemarginalizingTheIntersection}, in fact materializes through the counterfactual representation of each complainant under these two discrimination testing procedures.
Under multiple discrimination, we still rely on $\mathcal{D}_R^{CF}$ and $\mathcal{D}_G^{CF}$ for running the methods separately on $R$ and $G$. 
We look at male and non-white conterfactuals separately. 
Although together they cover all the non-protected groups, they do not do so simultaneously: the complainant under this procedure would have been male or non-white, but not female-non-white.
Under intersectional discrimination, instead, we rely on updated factual and counterfactual datasets based on a ``new'' protected attribute $R \times G$. 
The counterfactual for a given compliant implies simultaneously the possibilities of white-male, white-female, and non-white-male, which introduces more randomness.
 
In Figure~\ref{fig:LawSchool}, arguably, the worst-off sub-group between $R$ and $G$ is the group of non-white-female applicants.
The logic here is that non-white males, meaning $R=1$ and $G=0$, can always resort to their gender and white females, meaning $R=0$ and $G=1$, can always resort to their race.
Instead, non-white females have no single group within the space of $R \times G$ to resort to.
When we test for multiple discrimination we allow for these movements to occur by looking at $R$ and $G$ separately. 
This is because, by not intersecting $R$ and $G$, we do not consider the fact that the group at the intersection never has the choice to resort to a non-protected group. 
This lack of choice is what we represent when we test for intersectional discrimination by looking at $R \times G$ only.
The modeling problem for these forms of multidimensional discrimination requires further research with the goal of formalizing the role of the intersection and how it influences the control and test search spaces while taking into account the legal considerations discussed.

Back to Tables~\ref{table:k-results_Multiple} and \ref{table:k-results_Intersectional}, we argue that the observed difference comes from one protected attribute having a stronger influence than the other on the non-protected attributes. 
If that is the case, then testing separately for $R$ and $G$ should show individuals that are discriminated only by one of the protected attributes, which dismisses the multiple discrimination claim.
Table~\ref{table:MultivsInter} supports this argument. 
Notably, it is for this reason that lawyers discourage multiple discrimination claims and suggest that the complainant focuses on the most dominant protected attribute \parencite{Xenidis2020_TunningEULaw}.

In Table~\ref{table:MultivsInter}, given the results from the previous section, we focus on CST w/o for $k=15$ and look at individual discrimination cases detected as both multiple and intersectional discrimination.
We report the average $p_c$ and $p_t$ for $R$, $G$, and $R \times G$.
All multiple cases are included in the intersectional cases detected by CST w/o.
The first row shows these multiple discrimination cases. We observe that the average $p_c$ is greater than the average $p_t$, and thus the average $\Delta p > \tau$, for $R$, $G$, and $R \times G$. 
These are individual cases that suffer the negative effects of $R$ and $G$, separately and simultaneously, when applying to law school under $b()$.
The second row, instead, shows the intersectional cases only. 
For comparison, we provide the average $p_c$ and $p_t$ for these individuals' single discrimination tests for $R$ and $G$.
We observe that, on average, $R$ is the dominant protected attribute with a considerable difference in the proportion of negative outcomes between the control and test groups. 
This is not the case for $G$ where the average difference is negligible. 
Indeed, for these individuals, by looking at each protected attribute separately, we lose the multiple discrimination case.
In doing so, we also lose focus on what occurs at the intersection of $R \times G$.
The results in Table~\ref{table:MultivsInter} capture this lack of movement between protected and non-protected statuses experienced by those individuals at the bottom of the intersection of $R$ and $G$.

\begin{table}[t]
\caption{Avg. $p_c$ and $p_t$ for the ctr's and tst's groups of CST w/o for $k=15$.}
  \label{table:MultivsInter}
  \centering
  \begin{tabular}{ccccccc}
    \toprule
    & \multicolumn{6}{c}{Average}\\
    & $G$'s $p_c$ & $G$'s $p_t$ & $R$'s $p_c$ & $R$'s $p_t$ & $R \times G$'s $p_c$ & $R \times G$'s $p_t$\\
    \midrule
    Multi. and inter. & 0.59 & 0.33 & 0.51 & 0.00 & 0.66 & 0.00 \\
    Inter. only & 0.96 & 0.94 & 0.93 & 0.23 & 0.93 & 0.01 \\
    \bottomrule
  \end{tabular}
\end{table}
%

%
%

\section{Discussion}
\label{sec:Discussion}
With CST we move away from the standard, idealized comparison used in ST and other discrimination testing tools by operationalizing fairness given the difference. 
The results are promising but there are limitations to CST worth considering for future work.
In this section, we discuss the main ones.

\paragraph{The reliability of $\mathcal{D}^{CF}$.}
In this work, we do not constrain the generation of the counterfactual dataset. 
In principle, it is not an issue as the counterfactual distribution that results from intervening a given SCM $\mathcal{M}$ is unique and is the closest possible parallel world to the factual world \parencite{KarimiKSV2020_AlgoRecourseImperfectInfo, Woodward2005MakingThigsHappen}.
The resulting $\mathcal{D}^{CF}$ is reliable conditional on what is assumed by the SCM $\mathcal{M}$. 
Therefore, if the auxiliary causal knowledge is biased, then so will the counterfactual distribution and, in turn, $\mathcal{D}^{CF}$. 
What amounts to an unbiased SCM $\mathcal{M}$ is debatable when modeling humans behavior.
We come back to this point in the next theme.

Even if we do have access to an unbiased SCM $\mathcal{M}$, however, it is still difficult to judge what amounts to a reliable $\mathcal{D}^{CF}$ when our point of reference is the factual $\mathcal{D}$.
The issue here is accepting what $\mathcal{D}^{CF}$ represents based on our believes about $\mathcal{D}$. 
In Section~\ref{sec:Experiments.IllustrativeExample}, e.g.,~we find ourselves in a setting in which we have access to the data generating model of $\mathcal{D}$. 
The $\mathcal{D}^{CF}$, thus, represents the what would have been of the female applicants.
In Table~\ref{table:k-results} we observe the impact of considering the complainant and its counterfactual, observing how using CST increases the number of cases detected relative to ST and how such cases are penalized once we consider the statistical significance of $\Delta p$.
In Table~\ref{table:k-results}, one could argue that these counterfactuals are not reliable as they lead to non-statistically significant cases. 
One could also, though, argue that it is counterintuitive and even counterproductive to measure the reliability of the counterfactual world using the factual world.
In particular,
when it comes to indirect discrimination testing and its goal of achieving substantive equality \parencite{Wachter2020BiasPreserving}, 
we argue that it is conceivable to generate counterfactual distributions that cannot be other than non-representative of the current, non-neutral status quo behind the factual world. 
This is the case in Section~\ref{sec:Experiments.IllustrativeExample}.
For this reason alone, we expect \textit{ceteris paribus} over \textit{mutatis mutandis} to remain the preferred manipulation in discrimination testing as the idealized comparison is easier to motivate.
Further research is needed for defining what we mean, or want to mean, when speaking of a reliable $\mathcal{D}^{CF}$ when testing for discrimination, especially, if we wish to implement the \textit{mutatis mutandis} manipulation.

\paragraph{Auxiliary causal knowledge as groundtruth.}
Similarly, in this work we take as a given the derivation of the SCM $\mathcal{M}$, emphasizing it as a product of stakeholder engagement.
As with any model, the closer $\mathcal{M}$ is to the groundtruth, the better and more reliable is $\mathcal{D}^{CF}$ and, in turn, the results from CST.
This problem is often viewed in terms of missing confounders (see, e.g.,~\textcite{DBLP:conf/uai/KilbertusBKWS19}), such that what is missing are variables according to a theoretical model or domain expert. 
Given our focus on defining CST, we take a practical approach to the potential biasedness of the SCM $\mathcal{M}$, viewing it as evidence that is required from and that requires agreement among the stakeholders involved. 
$\mathcal{D}$ delimits the worldview of the discrimination context. Missing information, such as a confounder, should be addressed by the stakeholders when drawing $\mathcal{M}$ given $\mathcal{D}$.

Coming back to the first theme, the idea of groundtruth for describing human behavior is an open discussion within the fair ML community (see, e.g.,~\textcite{Hu_facct_sex_20, Kasirzadeh2021UseMisuse}). 
We agree that to speak of groundtruth in discrimination testing can be misleading, but that does not diminish the usefulness of using the SCM $\mathcal{M}$ for answering counterfactual questions.
As long as the SCM $\mathcal{M}$ is agreed upon by the stakeholders, we argue that the question on whether it amounts to groundtruth or not is secondary.\footnote{For instance, consider that there is already a similar and still unresolved discussion on what it means for individuals to be measurably similar between each other \parencite{Westen1982EmptyEquality}.}
That said, even if the stakeholders agree on a SCM $\mathcal{M}$, it is still possible for other SCMs to describe $\mathcal{D}$.
In this work, by always considering a single SCM $\mathcal{M}$, we implicitly work with faithful auxiliary causal knowledge \parencite{Peters2017_CausalInference}.
What happens when, e.g., the stakeholders agree on or are open to multiple SCMs?
We would find ourselves in a competing worlds setting for CST in which the sensitivity of the results depend on the faithfulness of the SCM $\mathcal{M}$ \parencite{DBLP:conf/nips/RussellKLS17}.
Future work should explore this line of work as a possible extension to CST.

\paragraph{A non-discriminatory ADM.}
This work positions CST within indirect discrimination testing for the ADM $b(X)=\hat{Y}$.
Recall that we express this type of discrimination using the DAG $A \rightarrow X \rightarrow Y$. 
Such DAG $\mathcal{G}$ implies the causal relation $X \leftarrow f(A, U)$ in the corresponding SCM $\mathcal{M}$.
We use this set up throughout Section~\ref{sec:Experiments}.
As emphasized in the previous two themes, $\mathcal{M}$ and $\mathcal{G}$ describe the dataset $\mathcal{D}$ on which $b()$ is used upon and condition the results of CST.
Given this setup, it is reasonable to ask: when does CST detect individual discrimination?
Broadly, the answer is when $A$ influences $X$ enough for $b()$ to make decisions that would have been different based on $X$ under a different $A$.
It follows, thus, that CST tests for whether the ADM relies or not on non-neutral information in $\mathcal{D}$. 
We stress, however, that detecting discrimination is not granted under CST only by having a biased $\mathcal{D}$.
It depends, for instance, on how much weight $b()$ gives to elements of $X$ and how these same elements relate to $A$.
Still,
since an unbiased $\mathcal{D}$ for high-stake decision-making context is rare to find \parencite{DBLP:journals/ethicsit/AlvarezCEFFFGMPLRSSZR24},
it is also reasonable to further ask: how often will CST detect individual discrimination or, put differently, what amounts to a non-discriminatory $b()$ given this setup? The answer to this follow up question is multidisciplinary and more complex.

When testing for indirect discrimination, we essentially suspect the presence of systematic biases---in the form of $A \rightarrow X$---that hinder $b()$ through $X$ and we test wether we are wrong given $\mathcal{D}$.
For high-stake settings, like those studied in Section~\ref{sec:Experiments}, we acknowledge that testing for indirect discrimination using CST sounds like a self-fulfilling prophecy.
However, two points are important here.
First, even if we do detect indirect discrimination in a context known to be biased, the evidence is one aspect of the discrimination testing pipeline \parencite{DBLP:conf/fat/WeertsXTOP23}. 
The decision-maker still can justify the use of $X$ in $b()$ as a business requirement.
We must keep in mind that CST detects \textit{prima facie} evidence.
Second, as argued by legal scholars \parencite{Hacker2018TeachingFairness, Xenidis2020_TunningEULaw, Wachter2020BiasPreserving}, the role of indirect non-discrimination law is to address this kind of setting. 
Methods like CST, at a minimum, raise awareness around the use of $X$ by $b()$ and motivate policies around the business requirements of the decision-maker. 

Raising awareness is an important byproduct of discrimination testing methods.
As shown in Section~\ref{sec:Experiments.IllustrativeExample}, CST detects the lack of neutrality of $X$ and its impact on $b()$ better than ST.
In that example, the bank could successfully justify to a court its use of annual income and account balance despite the results.
Further, beyond the business requirements, arguably, there could be a shared societal interest for the bank to make informed decisions, meaning we might want for the bank's $b()$ to be fair but not at the expense of the bank becoming insolvent or applicants going bankrupt \parencite{DBLP:journals/eor/KozodoiJL22, DBLP:conf/fat/DAmourSABSH20, DBLP:conf/fat/SchwobelR22}.
What matters, regardless, is showing through CST that the bank uses a non-neutral $X$, even if it does not translate into validating the complainant's discrimination claim.
It helps to acknowledge, through evidence, that we are not in the best possible circumstances as a society.
This ties back to the calls made by legal scholars on the role of indirect non-discrimination law and its aim for correcting the present non-neutral status quo. 
See, e.g., the discussion by \textcite{Wachter2020BiasPreserving} on substantive equality over formal equality in indirect non-discrimination law.
It also ties with calls within fair ML to acknowledge the present non-neutral status quo as a starting point and view the fair ML tools as corrective measures for achieving a better one.
See, e.g., work on revisiting the fairness-accuracy trade-off \parencite{DBLP:conf/nips/WickpT19} and on non-ideal over ideal fairness \parencite{Otto2024_WhatisImpossibleAboutAlgoFairness}.
Future work should formalize CST and its use of counterfactuals for envisioning and testing for a desired status quo. 

%
%

\section{Conclusion}
\label{sec:Conclusion}
In this work, we presented counterfactual situation testing (CST), a new actionable and meaningful framework for detecting individual discrimination in a dataset of classifier decisions.
We studied both single and multidimensional discrimination, focusing on the indirect setting.
For the latter kind, we compared its multiple and intersectional forms and provided the first evidence for the need to recognize intersectional discrimination as separate from multiple discrimination under non-discrimination law.
Compared to other methods, such as situation testing (ST) and counterfactual fairness (CF), CST uncovered more cases even when the classifier was counterfactually fair and after accounting for statistical significance.
For CF, in particular, we showed how CST equips it with confidence intervals, extending how we understand the robustness of this popular causal fairness definition. 

The decision-making settings tackled in this work are intended to showcase the CST framework and, importantly, to illustrate why it is necessary to draw a distinction between idealized and fairness given the difference comparisons when testing for individual discrimination. 
We hope the results motivate the adoption of the \textit{mutatis mutandis} manipulation over the \textit{ceteris paribus} manipulation.
We are aware that the experimental setting could be pushed further by considering higher dimensions or more complex causal structures. 
We leave this for future work.
Further,
extensions of CST should consider the impact of using different distance functions for measuring individual similarity \parencite{WilsonM97_HeteroDistanceFunctions}, and should explore a purely data-driven setup in which the running parameters and auxiliary causal knowledge are derived from the dataset \parencite{Cohen2013StatisticalPower, Peters2017_CausalInference}.
Furthermore,
extensions of CST should study settings in which the protected attribute goes beyond the binary, such as a high-cardinality categorical or an ordinal protected attribute \parencite{DBLP:journals/tkde/CerdaV22}. 
The setting in which the protected attribute is continuous is also of interest, though, in that case we could discretize it \parencite{DBLP:journals/tkde/GarciaLSLH13} and treat it as binary (the current setting) or as a high-cardinality categorical attribute.

Multidimensional discrimination testing is largely understudied \parencite{DBLP:conf/fat/0001HN23, WangRR22}. 
We have set a foundation for exploring the tension between multiple and intersectional discrimination, but future work should further study the problem of dealing with multiple protected attributes and their intersection.
It is of interest, for instance, formalizing the case in which one protected attribute dominates the others and the case in which the impact of each protected attribute varies based on individual characteristics.
While interaction terms and heterogeneous effects are understudied within SCM, both topics enjoy a well established literature in fields like economics \parencite{Wooldridge2015IntroductoryEconometrics}, which should enable future work.
We hope these extensions and, overall, the fairness given the difference powering the CST framework motivate new work on algorithmic discrimination testing.

%
%

\acks{A preliminary version of this work appeared in \textcite{DBLP:conf/eaamo/AlvarezR23}. This work has received funding from the European Union’s Horizon 2020 research and innovation program under Marie Sklodowska-Curie Actions (grant agreement number 860630) for the project ``NoBIAS - Artificial Intelligence without Bias''; and from M4C2 - Investimento 1.3, Partenariato Esteso PE00000013 - ``FAIR - Future Artificial Intelligence Research'' - Spoke~1 ``Human-centered AI''. This work reflects only the authors' views and the European Research Executive Agency (REA) is not responsible for any use that may be made of the information it contains. 
We thank the three anonymous reviewers for their comments, which helped us improve considerably this work.}

\appendix
\label{Appendix}

\section{Working Example for Generating Counterfactuals}
\label{Appendix.WorkingExampleAAP}

In this section, we present a working example to illustrate counterfactual generation. 
Given the assumptions we make for a SCM $\mathcal{M}$ \eqref{eq:SCM} in Section~\ref{sec:CausalKnowledge} and the additional assumption of an additive noise model (ANM) in Section~\ref{sec:Experiments}, such procedure is straightforward.
Suppose we have the following SCM $\mathcal{M}$ and corresponding DAG $\mathcal{G}$:

\begin{minipage}{.45\linewidth}
\begin{figure}[H]
\centering
    \begin{tikzpicture}
        \node (X1)  at (-1.65, 0) [circle, draw]{$X_1$};
        \node (X2) at (0, 0.85) [circle, draw]{$X_2$};
        \node (X3) at (1.65,0) [circle,draw]{$X_3$};
        \draw[->] (X1) to (X2) {};
        \draw[->] (X1) to (X3) {};
        \draw[->] (X2) to (X3) {};
    \end{tikzpicture}
\end{figure}
\end{minipage}
\begin{minipage}{.45\linewidth}
\begin{align*}
\mathcal{M} \, & 
\begin{cases}
    X_1 & \leftarrow U_1 \\
    X_2 & \leftarrow \alpha \cdot X_1  + U_2 \\
    X_3 & \leftarrow \beta_1 \cdot X_1 + \beta_2 \cdot X_2 + U_3
\end{cases}
\end{align*}
\end{minipage}

\medskip

\noindent
where $U_1, U_2, U_3$ represent the latent variables, $X_1, X_2, X_3$ the observed variables, and $\alpha, \beta_1, \beta_2$ the coefficient for the causal effect of, respectively, $X_1 \rightarrow X_2$,  $X_1 \rightarrow X_3$, and $X_2 \rightarrow X_3$. 
Suppose we want to generate the counterfactual for $X_3$, i.e., $X_3^{CF}$, had $X_1$ been equal to $x_1$. 
In the \textbf{abduction step}, we estimate $U_1$, $U_2$, and $U_3$ given the evidence, or what is observed, under the specified structural equations:
\begin{align*}
    \hat{U}_1 & = X_1 \\
    \hat{U}_2 & = X_2 - \alpha \cdot X_1 \\
    \hat{U}_3 & = X_3 - \beta_1 \cdot X_1 + \beta_2 \cdot X_2
\end{align*}
We generalize this step for \eqref{eq:SCM} as $U_j = X_j - f_j(X_{pa(j)})$ $\forall X_j \in X$. 
This step is an individual-level statement on the residual variation under SCM $\mathcal{M}$. 
It accounts for all that our assignment functions $f_j$, which are at the population level, cannot explain:~i.e., the \textit{error terms}. 
In the \textbf{action step}, we intervene $X_1$ and set all of its instances equal to $x_1$ via $do(X_1:=x_1)$ and obtain the intervened DAG $\mathcal{G}'$ and SCM $\mathcal{M}'$:

\begin{minipage}{.45\linewidth}
\begin{figure}[H]
\centering
    \begin{tikzpicture}
        \node (X1)  at (-1.65, 0) [circle, draw]{$do(x_1)$};
        \node (X2) at (0, 0.85) [circle, draw]{$X_2$};
        \node (X3) at (1.65,0) [circle,draw]{$X_3$};
        \draw[->] (X2) to (X3) {};
    \end{tikzpicture}
\end{figure}
\end{minipage}
\begin{minipage}{.45\linewidth}
\begin{align*}
\mathcal{M}' \, & 
\begin{cases}
    X_1 & = x_1 \\
    X_2 & \leftarrow \alpha \cdot x_1  + U_2 \\
    X_3 & \leftarrow \beta_1 \cdot x_1 + \beta_2 \cdot X_2 + U_3
\end{cases}
\end{align*}
\end{minipage}
\medskip

\noindent
where no edges come out from $X_1$ as it has been fixed to $x_1$. 
Finally, in the \textbf{prediction step}, we combine these two steps to calculate $X_3^{CF}$ under $\hat{U}$ and $\mathcal{M}'$:
\begin{align*}
    X_3^{CF} & \leftarrow \beta_1 \cdot x_1 + \beta_2 \cdot X_2 + \hat{U}_3 \\
             & \leftarrow \beta_1 \cdot x_1 + \beta_2 \cdot (\alpha \cdot x_1 + \hat{U}_2) + \hat{U}_3
\end{align*}
which is done for all instances in $X_3$. 
The same three steps apply to $X_2$ and $X_1$.

We view the above approach as \textit{frequentist}, in particular, with regard to the abduction step.
A more \textit{Bayesian} approach is what is done by \textcite{Kusner2017CF} in which they use a Monte Carlo Markov Chain (MCMC) to draw $\hat{U}$ by updating its prior distribution with the evidence $X$. 
In Section~\ref{sec:Experiments}, we used both approaches and found no difference in the results. 
In this work, we only present the results for the ``frequentist approach'' as it is less computationally expensive than running a MCMC. 

\section{Supplementary Material}
\label{Appendix.Supplements}

In this section, we present the supplementary material.

\subsection{Algorithms}

\begin{algorithm2e}[t]
\small 
    \caption{CST w/o $(\mathcal{D}, k, \tau, d)$}
    \label{alg:run_cfST}
	\SetKwInOut{Input}{Input}
	\SetKwInOut{Output}{Output}
	\Input{$\mathcal{D}$ - dataset, $k$ - neighborhood size, $\tau$ accepted deviation, $d$ distance function}
	\Output{$\mathcal{R}$ - set of pairs $(c, \Delta)$ of protected instance indexes and their $\Delta > \tau$}
	\BlankLine
	$\mathcal{D}_c=\{(x_i, a_i, \widehat{y}_i) \in \mathcal{D}: a_i=1\}$; \hfill\texttt{\scriptsize// control search space}\\
    $\mathcal{D}_t=\{(x_i, a_i, \widehat{y}_i) \in \mathcal{D}: a_i=0\}$\hfill\texttt{\scriptsize// test search space}\\
    $\mathcal{R} = \emptyset$\\
	\For{$(x_c, a_c, \widehat{y}_c) \in \mathcal{D}_c$}{
    $\text{\textit{k-ctr}} =
    \{ (x_i, a_i, \widehat{y}_i) \in \mathcal{D}_c: rank_{d}( x_c, x_i) \leq k \}$\hfill\texttt{\scriptsize// control group}\\
    $\text{\textit{k-tst}} = \{ (x_i, a_i, \widehat{y}_i) \in \mathcal{D}_t: rank_{d}( x^{CF}_c, x_i) \leq k \}$\hfill\texttt{\scriptsize// test group}\\
    $p_c = |\{ (x_i, a_i, \widehat{y}_i) \in \text{\textit{k-ctr}}: \hat{y}_i = 0 \}|/{k}$\hfill\texttt{\scriptsize// fraction of negative decisions for control}\\
    $p_t = |\{ (x_i, a_i, \widehat{y}_i) \in \text{\textit{k-tst}}: \hat{y}_i = 0 \}|/{k}$\hfill\texttt{\scriptsize// fraction of negative decisions for test}\\
    $\Delta = p_c - p_t$\hfill\texttt{\scriptsize// delta}\\
    \If{$\Delta > \tau$}{
    $\mathcal{R} = \mathcal{R} \cup \{(c, \Delta)\}$\hfill\texttt{\scriptsize// add pair to the result}\\
    }
	}
    \Return{$\mathcal{R}$}
\end{algorithm2e}

Algorithm~\ref{alg:run_cfST} reports the pseudo-code of the k-NN CST w/o algorithm. The pseudo-code is self-explanatory. 
After selecting the control and test search space (lines 1--2) as stated in Definition~\ref{def:SearchSpaces}, the algorithm iterates over the protected instances. For each of such instances, i.e., the complainant $c$, it builds (lines 5--6) the control and test groups as the $k$-nearest neighborhood instances relative to the distance $d$ for $x_c$ and for its counterfactual~$x^{CF}_c$ respectively, as stated in \eqref{eq:kctr} and \eqref{eq:ktst}. Then, the fractions $p_c$ and $p_t$ of negative decisions for the two groups are computed (lines 7--8) as stated in \eqref{eq:p1_and_p2}, as well as their difference $\Delta$ (line 9). If such a $\Delta$ is larger than the accepted deviation $\tau$, then the complainant $c$ and its $\Delta$ are added (lines 10--11) to the result $\mathcal{R}$.
The pseudo-code of k-NN CST w/ is a simple variant of Algorithm~\ref{alg:run_cfST}, which adds the search centers $x_c$ and $x^{CF}_c$ into the control and test groups, respectively, and divides by $k+1$ instead of $k$ at lines 7--8.

\subsection{Positive Discrimination}

Based on the discussion in Section~\ref{sec:CST_Disc}, we revisit Definitions \ref{def:IndDisc} and \ref{def:CIs} for testing individual positive discrimination under the k-NN CST. We still consider $\Delta p$ \eqref{eq:delta} and the converse of the one-sided CI $\eqref{eq:CIs}$. 

\begin{definition}[Positive Individual Discrimination]
\label{def:PostIndDisc}
    There is (potential) positive individual discrimination in favor of the complainant $c$ if $\Delta p < \tau$, meaning the negative decision outcomes rate for the control group is smaller than for the test group given some accepted deviation $\tau \in  [-1, 1]$.
\end{definition}
\begin{definition}[Confidence on the Positive Individual Discrimination Claim]
\label{def:PostCIs}
    A detected (potential) positive discrimination claim for the complainant $c$ by Definition~\ref{def:IndDisc}
    is statistically significant with significance level $\alpha$ if the CI $(- \infty, \Delta p + w_\alpha]$ excludes $\tau$.
\end{definition}

The concept of positive discrimination also applies to the case of multidimensional discrimination in Section~\ref{sec:CST.Multi}.
In that case, we would re-visit Definitions \ref{def:MultipleDisc} and \ref{def:IntersectionaleDisc} by looking at the opposite effect for $\Delta p$ across the protected attributes, be it each one of them or their intersection. 
We do not proceed with redefining these two definitions as we do not showcase them.
Again, especially for multidimensional discrimination, our focus is on discrimination \textit{against} protected groups.
Further, it is unclear what the legal scholarship views as positive multidimensional discrimination: from \textcite{Crenshaw1989_DemarginalizingTheIntersection} to \textcite{Xenidis2020_TunningEULaw}, the focus has been always on traditional discrimination.

\section{Additional Experiments}
\label{Appendix.AddExperiments}

In this section, we present additional experiments relative to the setup of Section~\ref{sec:Experiments}.

\subsection{Single Positive Discrimination}

For the same setup as in Section~\ref{sec:Experiments.SetUp} and the same data (factual and counterfactual) as in Section~\ref{sec:Experiments.IllustrativeExample}, we test for positive discrimination using Definitions \ref{def:PostIndDisc} and \ref{def:PostCIs}.
Regarding counterfactual fairness (CF), we define as positive CF discrimination when the factual has a positive decision outcome, $\hat{y}_c=1$ but its counterfactual a negative one, $\hat{y}_c^{CF}=0$.
Table~\ref{table:k-results_pos} summarizes the results for the protected attribute gender.

\begin{table}[t]
  \caption{Number (and \% females) of positive individual discrimination cases in Section~\ref{sec:Experiments.IllustrativeExample} based on gender. Marked by * are the statistically significant cases}
  \label{table:k-results_pos}
  \centering
  \begin{tabular}{clllll}
    \toprule
    Method & $k=15$ & $k=30$ & $k=50$ & $k=100$ & $k=250$\\
    \midrule
    CST w/o & 0 (0.0\%) & 0 (0.0\%) & 0 (0.0\%) & 0 (0.0\%)  & 0 (0.0\%) \\
     & 0* (0.0\%) & 0* (0.0\%) & 0* (0.0\%) & 0* (0.0\%)  & 0* (0.0\%) \\
     \midrule
    ST & 45 (2.6\%) & 50 (2.9\%) & 77 (4.5\%) & 118 (6.9\%) & 159 (9.3\%) \\
    & 41* (2.4\%) & 48* (2.8\%) & 55* (3.2\%) & 93* (5.4\%) & 120 (7.0\%) \\
    \midrule
    CST w/ & 0 (0.0\%) & 0 (0.0\%) & 0 (0.0\%) & 0 (0.0\%)  & 0 (0.0\%)\\
    & 0* (0.0\%) & 0* (0.0\%) & 0* (0.0\%) & 0* (0.0\%)  & 0* (0.0\%) \\
    \midrule
    CF & 0 (0.0\%) & 0 (0.0\%) & 0 (0.0\%) & 0 (0.0\%)  & 0 (0.0\%) \\
    & 0* (0.0\%) & 0* (0.0\%) & 0* (0.0\%) & 0* (0.0\%)  & 0* (0.0\%) \\
    \bottomrule
  \end{tabular}
\end{table}

Unlike all other single discrimination testing results in Section~\ref{sec:Experiments}, Table~\ref{table:k-results_pos} shows a ST that detects more cases than CST and a CST and CF that detect no cases at all. 
Both patterns hold when considering statistical significance.
These results are to be expected given how we generated the synthetic data for the loan application scenario. 
As described in Figure~\ref{fig:KarimiV2}, we introduced a negative systematic bias against female applicants in $\mathcal{D}$. 
This would explain why all the methods based on counterfactual generation---CF, CST w/, and CST w/o---detect zero cases: the generated male counterfactuals in $\mathcal{D}^{CF}$ can only improve over their female factual complainants in $\mathcal{D}$.
Hence, $\Delta p < \tau$ is very unlikely to occur when using CST or CF.
In other words, in a setting in which the protected individuals are always negatively affected in a systematic way, we should not expect to detect positive discrimination under methods that operationalize fairness given the difference.
Such results, for the purpose of this work, support our choice to consider only traditional discrimination as it is the most prevalent and important kind of discrimination when we suspect a negative systematic bias against the complainants. 

Table~\ref{table:k-results_pos} raises questions on what kind of comparison is better suited for testing positive discrimination. 
The fact that ST, which uses an idealized comparison by implementing the CP manipulation, detects discrimination cases in a known biased setting for female applicants puts further into question the role of standard methods like it.
Is the idealized comparison suitable for positive discrimination or does \textcite{Kohler2018CausalEddie}'s criticism also apply to this setting?
Based on these preliminary results, we would argue that the tension between \textit{ceteris paribus} and \textit{mutatis mutandis} manipulations applies also to testing positive discrimination.
The stark difference between ST and CST in Table~\ref{table:k-results_pos} reinforces our view as, essentially, once we account for fairness given the difference in a known biased setting, it is difficult to argue that such a thing as positive discrimination occurs at all. 
Perhaps this is why this kind of discrimination is not discussed as much by legal scholars looking at indirect discrimination.
We plan to revisit these results in future work.

\subsection{Single Discrimination Testing}

We re-run Section~\ref{sec:Experiments.IllustrativeExample} and \ref{sec:Experiments.Real} for $\tau=0.05$, keeping all other parameters equal.
The results align with the ones we present in the main body. 
We focus on individual discrimination for all cases, not distinguishing between statistically and non-statistically significant cases.

Table~\ref{table:k-results_tau005} shows the same pattern between the CST versions relative to ST and CF as in Table~\ref{table:k-results}.
It illustrates the robustness of our framework. 
Two points we want to raise regarding Table~\ref{table:k-results_tau005}. 
First, CF, as expected, detects the same number of cases as it always looks for the strict equality between the factual and counterfactual quantities.
Second, under $\tau=0.05$, CST w/ and CST w/o align in the number of cases for larger $k$ sizes. This shows how influential $\tau$ can be for detecting discrimination, but also shows that either CST version can tackle the discrimination problem.
Tables \ref{table:k-results_RACE_tau005} and \ref{table:k-results_GENDER_tau005} show similar results as in, which are the $\tau=0.05$ counterparts of Tables \ref{table:k-results_RACE} and \ref{table:k-results_GENDER}. 
The results are expected given the setup. 
For both experiments the number of cases drops under $\tau=0.05$ as we have increased the difficulty of proving the individual discrimination claims.

\begin{table}[H]
  \caption{Number and (\% w.r.t. females) of cases based on $A$ in Figure~\ref{fig:KarimiV2}.}
  \label{table:k-results_tau005}
  \centering
  \begin{tabular}{lcccc}
    \toprule
    Method & $k=15$ & $k=30$ & $k=50$ & $k=100$ \\
    \midrule
    CST w/o & 288 (16.8\%) & 307 (17.9\%) & 331 (19.3\%) & 360 (21.0\%)  \\
    ST & 55 (3.2\%) & 60 (3.5\%) & 75 (4.4\%) & 79 (4.6\%) \\
    CST w/ & 420 (24.5\%) & 309 (18.1\%) & 334 (19.5\%) & 363 (21.2\%)  \\
    CF &  376 (22\%) &  376 (22\%) &  376 (22\%) & 376 (22\%)  \\
    \bottomrule
  \end{tabular}
\end{table}
\begin{table}[H]
  \caption{Number (and \% w.r.t.~non-whites) of cases based on $R$ using Figure~\ref{fig:LawSchool}.}
  \label{table:k-results_RACE_tau005}
  \centering
  \begin{tabular}{lccccc}
    \toprule
    Method & $k=15$ & $k=30$ & $k=50$ & $k=100$ \\
    \midrule
    CST w/o & 256 (7.30\%) & 301 (8.59\%) & 323 (9.21\%) & 376 (10.72\%)  \\
    ST & 33 (0.94\%) & 48 (1.37\%) & 57 (1.63\%) & 46 (1.31\%) \\
    CST w/ & 286 (8.16\%) & 301 (8.59\%) & 323 (9.21\%) & 376 (10.72\%)  \\
    CF &  231 (6.59\%) &  231 (6.59\%) &  231 (6.59\%) & 231 (6.59\%)  \\
    \bottomrule
  \end{tabular}
\end{table}
\begin{table}[H]
  \caption{Number (and \% w.r.t.~females) of cases based on $G$ Figure~\ref{fig:LawSchool}.}
  \label{table:k-results_GENDER_tau005}
  \centering
  \begin{tabular}{lcccc}
    \toprule
    Method & $k=15$ & $k=30$ & $k=50$ & $k=100$ \\
    \midrule
    CST w/o & 78 (0.82\%) & 105 (1.10\%) & 224 (2.35\%) & 231 (2.42\%)  \\
    ST & 77 (0.81\%) & 92 (0.96\%) & 181 (1.90\%) & 185 (1.94\%) \\
    CST w/ & 99 (1.04\%) & 105 (1.10\%) & 224 (2.35\%) & 231 (2.42\%)  \\
    CF &  56 (0.59\%) &  56 (0.59\%) &  56 (0.59\%) & 56 (0.59\%)  \\
    \bottomrule
  \end{tabular}
\end{table}
\vskip 0.2in
\printbibliography

@inproceedings{DBLP:conf/eaamo/AlvarezR23,
  author       = {Jos{\'{e}} M. {\'{A}}lvarez and
                  Salvatore Ruggieri},
  title        = {Counterfactual Situation Testing: Uncovering Discrimination under
                  Fairness given the Difference},
  booktitle    = {{EAAMO}},
  pages        = {2:1--2:11},
  publisher    = {{ACM}},
  year         = {2023}
}

@inproceedings{DBLP:conf/nips/HardtPNS16,
  author       = {Moritz Hardt and
                  Eric Price and
                  Nati Srebro},
  title        = {Equality of Opportunity in Supervised Learning},
  booktitle    = {{NIPS}},
  pages        = {3315--3323},
  year         = {2016}
}

@article{DBLP:journals/tkde/CerdaV22,
  author       = {Patricio Cerda and
                  Ga{\"{e}}l Varoquaux},
  title        = {Encoding High-Cardinality String Categorical Variables},
  journal      = {{IEEE} Trans. Knowl. Data Eng.},
  volume       = {34},
  number       = {3},
  pages        = {1164--1176},
  year         = {2022}
}

@article{DBLP:journals/tkde/GarciaLSLH13,
  author       = {Salvador Garc{\'{\i}}a and
                  Juli{\'{a}}n Luengo and
                  Jos{\'{e}} Antonio S{\'{a}}ez and
                  Victoria L{\'{o}}pez and
                  Francisco Herrera},
  title        = {A Survey of Discretization Techniques: Taxonomy and Empirical Analysis
                  in Supervised Learning},
  journal      = {{IEEE} Trans. Knowl. Data Eng.},
  volume       = {25},
  number       = {4},
  pages        = {734--750},
  year         = {2013}
}

@article{Newcombe1998,
  author       = {Robert G. Newcombe},
  title        = {Interval estimation for the difference between independent proportions: comparison of eleven methods},
  journal      = {Statistics in Medicine},
  volume       = {17},
  number       = {8},
  pages        = {873--890},
  year         = {1998}
}

@inproceedings{Thanh_KnnSituationTesting2011,
  author    = {Binh Luong Thanh and
               Salvatore Ruggieri and
               Franco Turini},
  title     = {k-{NN} as an implementation of situation testing for discrimination
               discovery and prevention},
  booktitle = {{KDD}},
  pages     = {502--510},
  publisher = {{ACM}},
  year      = {2011}
}

@article{DBLP:journals/jiis/QureshiKKRP20,
  author    = {Bilal Qureshi and
               Faisal Kamiran and
               Asim Karim and
               Salvatore Ruggieri and
               Dino Pedreschi},
  title     = {Causal inference for social discrimination reasoning},
  journal   = {J. Intell. Inf. Syst.},
  volume    = {54},
  number    = {2},
  pages     = {425--437},
  year      = {2020}
}

@inproceedings{Zhang_CausalSituationTesting_2016,
  author    = {Lu Zhang and
               Yongkai Wu and
               Xintao Wu},
  title     = {Situation Testing-Based Discrimination Discovery: {A} Causal Inference
               Approach},
  booktitle = {{IJCAI}},
  pages     = {2718--2724},
  publisher = {{IJCAI/AAAI} Press},
  year      = {2016}
}

@article{Heckman1998_DetectingDiscrimination,
  title={Detecting Discrimination},
  author={Heckman, James J},
  journal={Journal of Economic Perspectives},
  volume={12},
  number={2},
  pages={101--116},
  year={1998}
}

@article{Bertrand2017_FieldExperimentDiscrimination,
  title={Field Experiments on Discrimination},
  author={Bertrand, Marianne and Duflo, Esther},
  journal={Handbook of Economic Field Experiments},
  volume={1},
  pages={309--393},
  year={2017},
  publisher={Elsevier}
}

@article{Hacker2018TeachingFairness,
  title={Teaching fairness to artificial intelligence: Existing and novel strategies against algorithmic discrimination under {EU} law},
  author={Hacker, Philipp},
  journal={Common Market Law Review},
  volume={55},
  number={4},
  year={2018}
}

@inproceedings{Hu_facct_sex_20,
  author    = {Lily Hu and
               Issa Kohler{-}Hausmann},
  title     = {What's sex got to do with machine learning?},
  booktitle = {FAT*},
  pages     = {513},
  publisher = {{ACM}},
  year      = {2020}
}

@inproceedings{Kasirzadeh2021UseMisuse,
  author    = {Atoosa Kasirzadeh and
               Andrew Smart},
  title     = {The Use and Misuse of Counterfactuals in Ethical Machine Learning},
  booktitle = {FAccT},
  pages     = {228--236},
  publisher = {{ACM}},
  year      = {2021}
}

@article{Kohler2018CausalEddie,
    title={Eddie {M}urphy and the Dangers of Counterfactual Causal Thinking about Detecting Racial Discrimination},
    author={Kohler-Hausmann, Issa},
    journal={Nw. UL Rev.},
    volume={113},
    pages={1163},
    year={2018},
    publisher={HeinOnline}
}

@article{Romei2014MultiSurveyDiscrimination,
  author    = {Andrea Romei and
               Salvatore Ruggieri},
  title     = {A multidisciplinary survey on discrimination analysis},
  journal   = {Knowl. Eng. Rev.},
  volume    = {29},
  number    = {5},
  pages     = {582--638},
  year      = {2014}
}

@article{Rorive2009_ProvingDiscrimination,
  title={Proving Discrimination Cases: The Role of Situation Testing},
  author={Rorive, Isabelle},
  year={2009},
  journal={Centre for Equal Rights and MPG},
  url = {https://ec.europa.eu/migrant-integration/library-document/proving-discrimination-cases-role-situation-testing_en}
}

@article{Bendick2007SituationTesting,
  title={Situation Testing for Employment Discrimination in the {U}nited {S}tates of {A}merica},
  author={Bendick, Marc},
  journal={Horizons strat{\'e}giques},
  volume={3},
  number={5},
  pages={17--39},
  year={2007}
}

@book{PearlCausality2009,
    author = {Pearl, Judea},
    edition = {2nd},
    publisher = {Cambridge University Press},
    title = {Causality: Models, Reasoning, and Inference},
    year = 2009
}

@book{Pearl2016_CausalInference,
  title={Causal Inference in Statistics: {A} Primer},
  author={Judea Pearl and Madelyn Glymour and Noicholas P. Jewell},
  year={2016},
  publisher={John Wiley \& Sons}
}

@article{Barocas2016_BigDataImpact,
  title     = {Big data's disparate impact},
  author    = {Barocas, Solon and Selbst, Andrew D},
  journal   = {California Law Review},
  volume    = {104},
  number    = {3},
  pages     = {671-732},
  year      = {2016},
  publisher = {HeinOnline}
}

@inproceedings{BlackYF20_FlipTest,
  author    = {Emily Black and
               Samuel Yeom and
               Matt Fredrikson},
  title     = {FlipTest: fairness testing via optimal transport},
  booktitle = {FAT*},
  pages     = {111--121},
  publisher = {{ACM}},
  year      = {2020}
}

@inproceedings{Karimi2021_AlgoRecourse,
  author    = {Amir{-}Hossein Karimi and
               Bernhard Sch{\"{o}}lkopf and
               Isabel Valera},
  title     = {Algorithmic Recourse: from Counterfactual Explanations to Interventions},
  booktitle = {FAccT},
  pages     = {353--362},
  publisher = {{ACM}},
  year      = {2021}
}

@article{Wachter2017Counterfactual,
  title={Counterfactual explanations without opening the black box: Automated decisions and the GDPR},
  author={Wachter, Sandra and Mittelstadt, Brent and Russell, Chris},
  journal={Harv. JL \& Tech.},
  volume={31},
  pages={841},
  year={2017},
  publisher={HeinOnline}
}

@inproceedings{Kilbertus2017AvoidDiscCau,
  author    = {Niki Kilbertus and
               Mateo Rojas{-}Carulla and
               Giambattista Parascandolo and
               Moritz Hardt and
               Dominik Janzing and
               Bernhard Sch{\"{o}}lkopf},
  title     = {Avoiding Discrimination through Causal Reasoning},
  booktitle = {{NIPS}},
  pages     = {656--666},
  year      = {2017}
}

@inproceedings{Kusner2017CF,
  author    = {Matt J. Kusner and
               Joshua R. Loftus and
               Chris Russell and
               Ricardo Silva},
  title     = {Counterfactual Fairness},
  booktitle = {{NIPS}},
  pages     = {4066--4076},
  year      = {2017}
}

@article{Godin2000Orchestra,
    Author = {Goldin, Claudia and Rouse, Cecilia},
    Title = {Orchestrating Impartiality: The Impact of ``Blind''' Auditions on Female Musicians},
    Journal = {American Economic Review},
    Volume = {90},
    Number = {4},
    Year = {2000},
    Pages = {715-741}
}

@article{Bertrand2004_EmilyAndGreg,
 author = {Marianne Bertrand and Sendhil Mullainathan},
 journal = {The American Economic Review},
 number = {4},
 pages = {991--1013},
 publisher = {American Economic Association},
 title = {Are {E}mily and {G}reg {M}ore {E}mployable than {L}akisha and {J}amal? {A} {F}ield {E}xperiment on {L}abor {M}arket {D}iscrimination},
 urldate = {2022-07-06},
 volume = {94},
 year = {2004}
}

@book{Fix&Struyk1993_ClearConvincingEvidence,
    title = {Clear and Convincing Evidence: Measurement of Discrimination in America},
    author = {Michael Fix and Raymond J. Struyk},
    year = {1993},
    publisher = {Urban Institute Press}
}

@inproceedings{Hoyer2008_ANM,
  author    = {Patrik O. Hoyer and
               Dominik Janzing and
               Joris M. Mooij and
               Jonas Peters and
               Bernhard Sch{\"{o}}lkopf},
  title     = {Nonlinear causal discovery with additive noise models},
  booktitle = {{NIPS}},
  pages     = {689--696},
  publisher = {Curran Associates, Inc.},
  year      = {2008}
}

@book{Peters2017_CausalInference,
    author = {Jonas Peters and Dominik Janzing and Bernhard Sch{\"{o}}lkopf},
    publisher = {The MIT Press},
    title = {Elements of Causal Inference: Foundations and Learning Algorithms},
    year = 2017
}

@book{Angrist2008MostlyHarmless,
  title={Mostly Harmless Econometrics},
  author={Angrist, Joshua D and Pischke, J{\"o}rn-Steffen},
  year={2008},
  publisher={Princeton University Press}
}

@article{Mallon2007SocialConstruction,
  title={A field guide to social construction},
  author={Mallon, Ron},
  journal={Philosophy Compass},
  volume={2},
  number={1},
  pages={93--108},
  year={2007},
  publisher={Wiley Online Library}
}

@book{Rothstein2017Color,
  title={The Color of Law: {A} Forgotten History of How our Government Segregated America},
  author={Rothstein, Richard},
  year={2017},
  publisher={Liveright Publishing}
}

@book{Schneider2008Smack,
  title={Smack: {H}eroin and the {A}merican city},
  author={Schneider, Eric C},
  year={2008},
  publisher={University of Pennsylvania Press}
}

@book{Adler2019MurderNewOrleansJimCrow,
  title={Murder in {N}ew {O}rleans: the creation of {J}im {C}row policing},
  author={Adler, Jeffrey S},
  year={2019},
  publisher={University of Chicago Press}
}

@book{CriadoPerez2019InvisibleWomen,
    title       = {Invisible Women},
    author      = {Caroline Criado-Perez},
    year        = {2019},
    publisher   = {Vintage},
}

@book{Wightman1998_LawDataSource,
 author = {Wightman, Linda F},
 publisher = {Law School Admission Council},
 title = {{LSAC} national longitudinal bar passage study},
 year = {1998}
}

@article{WilsonM97_HeteroDistanceFunctions,
  author    = {D. Randall Wilson and
               Tony R. Martinez},
  title     = {Improved Heterogeneous Distance Functions},
  journal   = {J. Artif. Intell. Res.},
  volume    = {6},
  pages     = {1--34},
  year      = {1997}
}

@book{Cohen2013StatisticalPower,
  title={Statistical power analysis for the behavioral sciences},
  author={Cohen, Jacob},
  year={2013},
  publisher={Routledge}
}

@article{rose_constructivist_2022,
	title = {A {Constructivist} {Perspective} on {Empirical} {Discrimination} {Research}},
	journal = {Journal of Economic Literature},
       volume={61},
       issue={3},
        pages={906–-923},
	author = {Rose, Evan K.},
	year = {2022},
}

@inproceedings{DworkHPRZ12,
  author    = {Cynthia Dwork and
               Moritz Hardt and
               Toniann Pitassi and
               Omer Reingold and
               Richard S. Zemel},
  title     = {Fairness through awareness},
  booktitle = {{ITCS}},
  pages     = {214--226},
  publisher = {{ACM}},
  year      = {2012}
}

@article{Xenidis2020_TunningEULaw,
author = {Raphaële Xenidis},
title ={Tuning {EU} equality law to algorithmic discrimination: Three pathways to resilience},
journal = {Maastricht Journal of European and Comparative Law},
volume = {27},
number = {6},
pages = {736-758},
year = {2020},    
}

@inproceedings{Chzhen2020WassersteinBarycenters,
  author    = {Evgenii Chzhen and
               Christophe Denis and
               Mohamed Hebiri and
               Luca Oneto and
               Massimiliano Pontil},
  title     = {Fair regression with {W}asserstein barycenters},
  booktitle = {NeurIPS},
  year      = {2020}
}

@article{Chzhen2022MiniMax,
  title={A minimax framework for quantifying risk-fairness trade-off in regression},
  author={Chzhen, Evgenii and Schreuder, Nicolas},
  journal={The Annals of Statistics},
  volume={50},
  number={4},
  pages={2416--2442},
  year={2022},
  publisher={Institute of Mathematical Statistics}
}

@article{Adams2022DirectlyDiscriminatoryAl,
  title={Directly Discriminatory Algorithms},
  author={Adams-Prassl, Jeremias and Binns, Reuben and Kelly-Lyth, Aislinn},
  journal={The Modern Law Review},
  volume={86},
  number={1},
  pages={144--175},
  year={2023},
  publisher={Wiley Online Library}
}

@inproceedings{WangRR22,
  author    = {Angelina Wang and
               Vikram V. Ramaswamy and
               Olga Russakovsky},
  title     = {Towards Intersectionality in Machine Learning: Including More Identities,
               Handling Underrepresentation, and Performing Evaluation},
  booktitle = {FAccT},
  pages     = {336--349},
  publisher = {{ACM}},
  year      = {2022}
}

@article{Angwin2016MachineBias,
    title = {Machine Bias},
    author = {Angwin, Julia and Larson, Jef and Mattu, Surya and Kirchner, Lauren},
    journal={ProPublica},
    year={2016}
}

@article{DastinAmazonSexist,
    author = {Jeffrey Dastin},
    year = {2018},
    title = {Amazon scraps secret {AI} recruiting tool that showed bias against women},
    journal = {Reuters}
}

@article{Heikkila2022_DutchScnadal,
    author = {Melissa Heikkila},
    year = {2022},
    title = {Dutch scandal serves as a warning for Europe over risks of using algorithms},
    journal = {POLITICO}
}

@article{Nachbar2020algorithmic,
  title={Algorithmic fairness, algorithmic discrimination},
  author={Nachbar, Thomas B},
  journal={Florida State University Law Review},
  volume={48},
  pages={50},
  year={2021},
  publisher={HeinOnline}
}

@inproceedings{TR-DBLP:conf/sigsoft/GalhotraBM17,
  author    = {Sainyam Galhotra and
               Yuriy Brun and
               Alexandra Meliou},
  title     = {Fairness testing: testing software for discrimination},
  booktitle = {{ESEC/SIGSOFT} {FSE}},
  pages     = {498--510},
  publisher = {{ACM}},
  year      = {2017}
}

@article{TR-DBLP:journals/corr/abs-1809-03260,
  author    = {Aniya Aggarwal and
               Pranay Lohia and
               Seema Nagar and
               Kuntal Dey and
               Diptikalyan Saha},
  title     = {Automated Test Generation to Detect Individual Discrimination in {AI}
               Models},
  journal   = {CoRR},
  volume    = {abs/1809.03260},
  year      = {2018}
}

@article{Sen2016_RaceABundle,
  title={Race as a bundle of sticks: Designs that estimate effects of seemingly immutable characteristics},
  author={Sen, Maya and Wasow, Omar},
  journal={Annual Review of Political Science},
  volume={19},
  number={1},
  pages={499--522},
  year={2016}
}

@article{Bonilla1997_RethinkingRace,
  title={Rethinking Racism: {T}oward a structural interpretation},
  author={Bonilla-Silva, Eduardo},
  journal={American Sociological Review},
  pages={465--480},
  year={1997}
}

@article{DBLP:journals/tosem/ChenZHHS24,
  author       = {Zhenpeng Chen and
                  Jie M. Zhang and
                  Max Hort and
                  Mark Harman and
                  Federica Sarro},
  title        = {Fairness Testing: {A} Comprehensive Survey and Analysis of Trends},
  journal      = {{ACM} Trans. Softw. Eng. Methodol.},
  volume       = {33},
  number       = {5},
  pages        = {137:1--137:59},
  year         = {2024}
}

@article{DBLP:journals/jlap/MakhloufZP24,
  author       = {Karima Makhlouf and
                  Sami Zhioua and
                  Catuscia Palamidessi},
  title        = {When causality meets fairness: {A} survey},
  journal      = {J. Log. Algebraic Methods Program.},
  volume       = {141},
  pages        = {101000},
  year         = {2024}
}

@article {Foster2004,
  title="Causation in antidiscrimination law: {B}eyond intent versus impact",
  author="S. R. Foster",
  Journal= {Houston Law Review},
  pages="1469--1548",
  volume=41,
  number=5,
  year=2004
}

@article{DBLP:journals/fdata/CareyW22,
  author    = {Alycia N. Carey and
               Xintao Wu},
  title     = {The Causal Fairness Field Guide: {P}erspectives From Social and Formal
               Sciences},
  journal   = {Frontiers Big Data},
  volume    = {5},
  pages     = {892837},
  year      = {2022}
}

@inproceedings{Yang2021_CausalIntersectionality,
  author    = {Ke Yang and
               Joshua R. Loftus and
               Julia Stoyanovich},
  title     = {Causal Intersectionality and Fair Ranking},
  booktitle = {{FORC}},
  series    = {LIPIcs},
  volume    = {192},
  pages     = {7:1--7:20},
  publisher = {Schloss Dagstuhl - Leibniz-Zentrum f{\"{u}}r Informatik},
  year      = {2021}
}

@misc{Mulligan2022_AFCP,
    author = {Deirdre Mulligan},
    title = {Invited Talk: Fairness and Privacy},
    year = {2022},
    howpublished = {https://www.afciworkshop.org/afcp2022},
    note = {At the NeurIPS 2022 Workshop on Algorithmic Fairness through the Lens of Causality and Privacy.}
}

@article{Rooth2021,
  author    = {Dan-Olof Rooth},
  title     = {Correspondence testing studies},
  journal   = {{IZA} World of Labor},
  volume    = {58},
  year      = {2021}
}

@article{Plecko2022_CFA,
    author       = {Drago Plecko and Elias Bareinboim},
    title        = {Causal Fairness Analysis: {A} Causal Toolkit for Fair Machine Learning},
    journal      = {Found. Trends Mach. Learn.},
    volume       = {17},
    number       = {3},
    pages        = {304--589},
    year         = {2024}
}

@inproceedings{Chiappa2019_PathCF,
  author    = {Silvia Chiappa},
  title     = {Path-Specific Counterfactual Fairness},
  booktitle = {{AAAI}},
  pages     = {7801--7808},
  publisher = {{AAAI} Press},
  year      = {2019}
}

@book{Hastie2009_ElementsSL,
  title={The elements of statistical learning: data mining, inference, and prediction},
  author={Hastie, Trevor and Tibshirani, Robert and Friedman, Jerome H and Friedman, Jerome H},
  volume={2},
  year={2009},
  publisher={Springer}
}

@article{Athey2019MachineLearningForEconomists,
  title={Machine learning methods that economists should know about},
  author={Athey, Susan and Imbens, Guido W},
  journal={Annual Review of Economics},
  volume={11},
  pages={685--725},
  year={2019},
  publisher={Annual Reviews}
}

@inproceedings{Hanna2020_CriticalRace,
  author    = {Alex Hanna and
               Emily Denton and
               Andrew Smart and
               Jamila Smith{-}Loud},
  title     = {Towards a critical race methodology in algorithmic fairness},
  booktitle = {FAT*},
  pages     = {501--512},
  publisher = {{ACM}},
  year      = {2020}
}

@inproceedings{Tschantz2022_ProxyDisc,
  author    = {Michael Carl Tschantz},
  title     = {What is Proxy Discrimination?},
  booktitle = {FAccT},
  pages     = {1993--2003},
  publisher = {{ACM}},
  year      = {2022}
}

@misc{EU2018_NonDiscriminationLaw,
    author = {EU-FRA},
    title = {Handbook on {E}uropean non-discrimination law},
    year = {2018},
    howpublished = {\url{https://fra.europa.eu}},
    note = {Downloaded in 2023.}
}

@inproceedings{DBLP:conf/uai/KilbertusBKWS19,
  author    = {Niki Kilbertus and
               Philip J. Ball and
               Matt J. Kusner and
               Adrian Weller and
               Ricardo Silva},
  title     = {The Sensitivity of Counterfactual Fairness to Unmeasured Confounding},
  booktitle = {{UAI}},
  series    = {Proceedings of Machine Learning Research},
  volume    = {115},
  pages     = {616--626},
  publisher = {{AUAI} Press},
  year      = {2019}
}

@inproceedings{DBLP:conf/nips/LouizosSMSZW17,
  author    = {Christos Louizos and
               Uri Shalit and
               Joris M. Mooij and
               David A. Sontag and
               Richard S. Zemel and
               Max Welling},
  title     = {Causal Effect Inference with Deep Latent-Variable Models},
  booktitle = {{NIPS}},
  pages     = {6446--6456},
  year      = {2017}
}

@article{mccandless2007bayesian,
  title={Bayesian sensitivity analysis for unmeasured confounding in observational studies},
  author={McCandless, Lawrence C and Gustafson, Paul and Levy, Adrian},
  journal={Statistics in Medicine},
  volume={26},
  number={11},
  pages={2331--2347},
  year={2007},
  publisher={Wiley Online Library}
}

@article{DBLP:journals/corr/abs-1902-10286,
  author    = {Alexander D'Amour},
  title     = {On Multi-Cause Causal Inference with Unobserved Confounding: Counterexamples,
               Impossibility, and Alternatives},
  journal   = {CoRR},
  volume    = {abs/1902.10286},
  year      = {2019}
}

@article{Kleinberg2019DiscAgeOfAlgo,
  author       = {Jon M. Kleinberg and
                  Jens Ludwig and
                  Sendhil Mullainathan and
                  Cass R. Sunstein},
  title        = {Discrimination in the Age of Algorithms},
  journal      = {CoRR},
  volume       = {abs/1902.03731},
  year         = {2019}
}

@inproceedings{DBLP:conf/aaai/Ruggieri0PST23,
  author       = {Salvatore Ruggieri and
                  Jos{\'{e}} M. {\'{A}}lvarez and
                  Andrea Pugnana and
                  Laura State and
                  Franco Turini},
  title        = {Can We Trust Fair-{AI}?},
  booktitle    = {{AAAI}},
  pages        = {15421--15430},
  publisher    = {{AAAI} Press},
  year         = {2023}
}

@inproceedings{DBLP:conf/kdd/PedreschiRT08,
  author       = {Dino Pedreschi and
                  Salvatore Ruggieri and
                  Franco Turini},
  title        = {Discrimination-aware data mining},
  booktitle    = {{KDD}},
  pages        = {560--568},
  publisher    = {{ACM}},
  year         = {2008}
}

@article{DBLP:journals/tkdd/RuggieriPT10,
  author       = {Salvatore Ruggieri and
                  Dino Pedreschi and
                  Franco Turini},
  title        = {Data mining for discrimination discovery},
  journal      = {{ACM} Trans. Knowl. Discov. Data},
  volume       = {4},
  number       = {2},
  pages        = {9:1--9:40},
  year         = {2010}
}

@inproceedings{DBLP:conf/nips/RussellKLS17,
  author       = {Chris Russell and
                  Matt J. Kusner and
                  Joshua R. Loftus and
                  Ricardo Silva},
  title        = {When Worlds Collide: Integrating Different Counterfactual Assumptions
                  in Fairness},
  booktitle    = {{NIPS}},
  pages        = {6414--6423},
  year         = {2017}
}

@inproceedings{DBLP:journals/corr/abs-2307-12797,
  author       = {Ludwig Bothmann and
                  Susanne Dandl and
                  Michael Schomaker},
  title        = {Causal Fair Machine Learning via Rank-Preserving Interventional Distributions},
  booktitle    = {AEQUITAS@ECAI},
  series       = {{CEUR} Workshop Proceedings},
  volume       = {3523},
  publisher    = {CEUR-WS.org},
  year         = {2023}
}

@article{Wachter2020BiasPreserving,
  title={Bias preservation in machine learning: the legality of fairness metrics under EU non-discrimination law},
  author={Wachter, Sandra and Mittelstadt, Brent and Russell, Chris},
  journal={W. Va. L. Rev.},
  volume={123},
  pages={735},
  year={2020},
  publisher={HeinOnline}
}

@inproceedings{DBLP:conf/fat/WeertsXTOP23,
  author       = {Hilde J. P. Weerts and
                  Rapha{\"{e}}le Xenidis and
                  Fabien Tarissan and
                  Henrik Palmer Olsen and
                  Mykola Pechenizkiy},
  title        = {Algorithmic Unfairness through the Lens of {EU} Non-Discrimination
                  Law: Or Why the Law is not a Decision Tree},
  booktitle    = {FAccT},
  pages        = {805--816},
  publisher    = {{ACM}},
  year         = {2023}
}

@article{Lippert2006BadnessOfDiscrimination,
  title={The badness of discrimination},
  author={Lippert-Rasmussen, Kasper},
  journal={Ethical Theory and Moral Practice},
  volume={9},
  pages={167--185},
  year={2006},
  publisher={Springer}
}

@article{Westen1982EmptyEquality,
  title={The empty idea of equality},
  author={Westen, Peter},
  journal={Harvard Law Review},
  pages={537--596},
  year={1982}
}

@article{DBLP:journals/ethicsit/AlvarezCEFFFGMPLRSSZR24,
  author       = {Jos{\'{e}} M. {\'{A}}lvarez and
                  Alejandra Bringas Colmenarejo and
                  Alaa Elobaid and
                  Simone Fabbrizzi and
                  Miriam Fahimi and
                  Antonio Ferrara and
                  Siamak Ghodsi and
                  Carlos Mougan and
                  Ioanna Papageorgiou and
                  Paula Reyero Lobo and
                  Mayra Russo and
                  Kristen M. Scott and
                  Laura State and
                  Xuan Zhao and
                  Salvatore Ruggieri},
  title        = {Policy advice and best practices on bias and fairness in {AI}},
  journal      = {Ethics Inf. Technol.},
  volume       = {26},
  number       = {2},
  pages        = {31},
  year         = {2024}
}

@inproceedings{DBLP:conf/fat/0001HN23,
  author       = {Arjun Roy and
                  Jan Horstmann and
                  Eirini Ntoutsi},
  title        = {Multi-dimensional Discrimination in Law and Machine Learning - {A}
                  Comparative Overview},
  booktitle    = {FAccT},
  pages        = {89--100},
  publisher    = {{ACM}},
  year         = {2023}
}

@inproceedings{Zemel2013LearningFairRepresentations,
  author       = {Richard S. Zemel and
                  Yu Wu and
                  Kevin Swersky and
                  Toniann Pitassi and
                  Cynthia Dwork},
  title        = {Learning Fair Representations},
  booktitle    = {{ICML} {(3)}},
  series       = {{JMLR} Workshop and Conference Proceedings},
  volume       = {28},
  pages        = {325--333},
  publisher    = {JMLR.org},
  year         = {2013}
}

@article{Plevcko2020FairDataAdaptation,
  title={Fair data adaptation with quantile preservation},
  author={Ple{\v{c}}ko, Drago and Meinshausen, Nicolai},
  journal={Journal of Machine Learning Research},
  volume={21},
  number={242},
  pages={1--44},
  year={2020}
}

@article{Gower1971,
 author = {J. C. Gower},
 journal = {Biometrics},
 number = {4},
 pages = {857--871},
 publisher = {International Biometric Society},
 title = {A General Coefficient of Similarity and Some of Its Properties},
 urldate = {2024-10-10},
 volume = {27},
 year = {1971}
}

@inproceedings{Kamiran2009_ClassifyingWihtoutDiscriminating,
  title={Classifying without discriminating},
  author={Kamiran, Faisal and Calders, Toon},
  booktitle={Int. Conf. on Computer, Control and Communication},
  pages={1--6},
  year={2009},
  organization={IEEE}
}

@article{Crenshaw1989_DemarginalizingTheIntersection,
  title={Demarginalizing the intersection of race and sex: A black feminist critique of antidiscrimination doctrine, feminist theory and antiracist politics},
  author={Crenshaw, Kimberl{\'e}},
  pages={139--167},
  year={1989},
  journal={University of Chicago Legal Forum},
  volume={1989},
  issue={1},
  publisher={University of Chicago}
}

@book{Wooldridge2015IntroductoryEconometrics,
  title={Introductory Econometrics: A Modern Approach},
  author={Wooldridge, Jeffrey M},
  year={2015},
  publisher={Cengage Learning}
}

@inproceedings{KarimiKSV2020_AlgoRecourseImperfectInfo,
  author    = {Amir{-}Hossein Karimi and
               Bodo Julius von K{\"{u}}gelgen and
               Bernhard Sch{\"{o}}lkopf and
               Isabel Valera},
  title     = {Algorithmic recourse under imperfect causal knowledge: a probabilistic
               approach},
  booktitle = {NeurIPS},
  year      = {2020}
}

@book{Woodward2005MakingThigsHappen,
  title={Making Things Happen: A Theory of Causal Explanation},
  author={Woodward, James},
  year={2005},
  publisher={Oxford university press}
}

@inproceedings{DBLP:conf/nips/WickpT19,
  author       = {Michael L. Wick and
                  Swetasudha Panda and
                  Jean{-}Baptiste Tristan},
  title        = {Unlocking Fairness: a Trade-off Revisited},
  booktitle    = {NeurIPS},
  pages        = {8780--8789},
  year         = {2019}
}

@article{Otto2024_WhatisImpossibleAboutAlgoFairness,
    author  = {Sahlgren, Otto},
    title   = {What’s Impossible about Algorithmic Fairness?},
    journal = {Philosophy and Technology},
    volume  = {37},
    number  = {4},
    paper = {124},
    year  = {2024},
}

@inproceedings{DBLP:conf/fat/DAmourSABSH20,
  author       = {Alexander D'Amour and
                  Hansa Srinivasan and
                  James Atwood and
                  Pallavi Baljekar and
                  D. Sculley and
                  Yoni Halpern},
  title        = {Fairness is not static: deeper understanding of long term fairness
                  via simulation studies},
  booktitle    = {FAT*},
  pages        = {525--534},
  publisher    = {{ACM}},
  year         = {2020}
}

@inproceedings{DBLP:conf/fat/SchwobelR22,
  author       = {Pola Schw{\"{o}}bel and
                  Peter Remmers},
  title        = {The Long Arc of Fairness: Formalisations and Ethical Discourse},
  booktitle    = {FAccT},
  pages        = {2179--2188},
  publisher    = {{ACM}},
  year         = {2022}
}

@article{NPR2023AffirmativeAction,
  author       = {Nina Totenberg},
  title        = {Supreme Court guts affirmative action, effectively ending race conscious admissions},
  year         = {2023},
  month        = {6},
  day          = {29},
  journal      = {NPR},
  url          = {https://www.npr.org/2023/06/29/1181138066/affirmative-action-supreme-court-decision}
}

@article{MutatisMutandis,
  author       = {Jos{\'{e}} M. {\'{A}}lvarez and
                  Salvatore Ruggieri},
  title        = {Mutatis Mutandis: Revisiting the Comparator in Discrimination Testing},
  journal      = {CoRR},
  volume       = {abs/2405.13693},
  year         = {2024}
}

@article{DBLP:journals/eor/KozodoiJL22,
  author       = {Nikita Kozodoi and
                  Johannes Jacob and
                  Stefan Lessmann},
  title        = {Fairness in credit scoring: Assessment, implementation and profit
                  implications},
  journal      = {Eur. J. Oper. Res.},
  volume       = {297},
  number       = {3},
  pages        = {1083--1094},
  year         = {2022}
}

\end{document}